\documentclass[pdflatex,sn-mathphys-num]{sn-jnl}

\usepackage{graphicx}%
\usepackage{multirow}%
\usepackage{amsmath,amssymb,amsfonts}%
\usepackage{amsthm}%
\usepackage{mathrsfs}%
\usepackage[title]{appendix}%
\usepackage{xcolor}%
\usepackage{textcomp}%
\usepackage{manyfoot}%
\usepackage{booktabs}%
\usepackage{algorithm}%
\usepackage{algorithmicx}%
\usepackage{algpseudocode}%
\usepackage{listings}%
\usepackage{rotating}

\usepackage{xcolor}
\usepackage{makecell}
\usepackage{multicol}
\usepackage{natbib}

\usepackage[normalem]{ulem}

\theoremstyle{thmstyleone}%
\theoremstyle{thmstyletwo}%

\theoremstyle{thmstylethree}%

\begin{document}
\begin{NoHyper}
\title[Data-free neural operator for NSE]{A data-free neural operator enabling fast inference of 2D and 3D Navier–Stokes equations}

\author[1]{\fnm{Junho} \sur{Choi}}\email{junho\_choi@kaist.ac.kr}

\author[2]{\fnm{Teng-Yuan} \sur{Chang}}\email{tony890048@snu.ac.kr}

\author*[3]{\fnm{Namjung} \sur{Kim}}\email{namjungk@gachon.ac.kr}

\author*[2]{\fnm{Youngjoon} \sur{Hong}}\email{hongyj@snu.ac.kr}

\affil[1]{\orgdiv{Department of Mathematical Sciences}, \orgname{Korea Advanced Institute of Science and Technology}, \orgaddress{\city{Daejeon}, \postcode{34141}, \country{Republic of Korea}}}

\affil[2]{\orgdiv{Department of Mathematical Sciences}, \orgname{Seoul National University}, \orgaddress{\city{Seoul}, \postcode{08826}, \country{Republic of Korea}}}

\affil[3]{\orgdiv{Department of Mechanical Engineering}, \orgname{Gachon University}, \orgaddress{\postcode{610101},  \country{Republic of Korea}}}


\abstract{
Ensemble simulations of high-dimensional flow models (e.g., Navier–Stokes–type PDEs) are computationally prohibitive for real-time applications. 
Neural operators enable fast inference but are limited by costly data requirements and poor generalization to 3D flows.
We present a data-free operator network for the Navier–Stokes equations that eliminates the need for paired solution data and enables robust, real-time inference for large ensemble forecasting.
The physics-grounded architecture takes initial and boundary conditions as well as forcing functions, yielding solutions robust to high variability and perturbations.
Across 2D benchmarks and 3D test cases, the method surpasses prior neural operators in accuracy and, for ensembles, achieves greater efficiency than conventional numerical solvers. Notably, it delivers accurate solutions of the three-dimensional Navier–Stokes equations—a regime not previously demonstrated for data-free neural operators.
By uniting a numerically grounded architecture with the scalability of machine learning, this approach establishes a practical pathway toward data-free, high-fidelity PDE surrogates for end-to-end scientific simulation and prediction.}




\maketitle

Partial differential equations (PDEs) form the mathematical foundation of physical laws that govern a broad spectrum of scientific and engineering systems. Solving PDEs efficiently and accurately is one of the central interests for science and engineering. In addition, when dealing with various boundary conditions, initial conditions, or external forcing terms of PDEs in fields such as fluid mechanics \cite{anderson1995computational,wilcox1998turbulence,landau2013fluid}, materials science \cite{borden2012phase,giustino2014materials}, weather forecasting \cite{knj03,bauer2015quiet}, and design optimization \cite{li2022machine,martins2021engineering}, PDEs are often required to be solved repeatedly. However, conventional numerical solvers become prohibitively expensive in such settings, particularly for three-dimensional incompressible Navier–Stokes equations (NSEs) \cite{PINN007,vinuesa2022enhancing}. This is because these solvers rely on spatial–temporal discretization and iterative treatment of nonlinear terms, while performing time marching that demands substantial memory and computation. Moreover, they are not well suited for solving large ensembles of scenarios simultaneously, such as those required for uncertainty quantification or design exploration. The resulting computational time, coupled with the need for extensive sampling in ensemble or probabilistic simulations, constitutes a critical bottleneck \cite{bauer2015quiet,price2025probabilistic}.

Neural operator methods such as DeepONet \cite{lu2021learning,lu2022comprehensive} and the Fourier Neural Operator (FNO) \cite{li2020fourier,li2023geometry}, as well as other PDE operator approaches \cite{yin2024scalable,kadeethum2021framework}, have emerged as promising alternatives that learn mappings between infinite-dimensional function spaces to approximate PDE solution operators.
While these approaches offer fast inference once trained, they rely on supervised learning with large datasets of precomputed high-fidelity solutions. This dependence on costly reference data substantially limits their practicality, especially for three-dimensional problems where generating accurate ground-truth solutions is computationally expensive \cite{vinuesa2022enhancing}.
Moreover, most existing frameworks simplify the problem by focusing on periodic domains, where velocity–vorticity formulations can be adopted to make training easier and to achieve better accuracy \cite{jin2021nsfnets,cho2023separable,wang2024respecting,lu2021learning,lu2022comprehensive,ovadia2024vito}. This simplification avoids the challenges inherent in realistic non-periodic settings, where enforcing general boundary conditions such as Dirichlet is more difficult. In practice, existing operator networks often exhibit degraded accuracy when applied with Dirichlet boundaries. Moreover, in three dimensions, the increased complexity and dimensionality render training unstable, and no prior neural operator has yet achieved accurate solutions for the full 3D Navier–Stokes equations.
Physics-informed neural networks (PINNs) \cite{PINN009} mitigate the need for explicit data by embedding PDE residuals into the loss function \cite{cuomo2022scientific,cho2023separable,shukla2024comprehensive}. However, they are not operator networks—they compute single-instance solutions rather than learning the general mapping between input and output functions. As a result, while effective for individual cases, PINNs are not suitable for generating families of solutions or achieving rapid generalization across diverse conditions.

In this work, we introduce a spectral operator network (SpecONet) that overcomes this limitation. 
SpecONet learns solution operators for incompressible NSEs in both two and three dimensions without any precomputed solution data. Unlike prior approaches, it does not approximate PDEs through supervised surrogates but learns directly from the governing equations via variational weak-form losses derived from spectral element discretization. The network predicts spectral coefficients of velocity and pressure fields from diverse PDE inputs—boundary conditions, initial states, or forcing terms—and reconstructs full solutions through known basis functions. This formulation not only ensures physical consistency but also achieves high numerical accuracy and robustness, while providing fast inference that makes the method practical for a wide range of applications.

SpecONet addresses the key limitations of existing neural operator approaches through a unified, physics-structured design.
It is trained entirely without reference data, relying solely on the governing equations, and achieves robust, data-free learning across diverse input modalities—initial, boundary, and forcing conditions—with strong generalization across distributional shifts and robustness to perturbed input functions.
The framework further provides, to our knowledge, the first demonstration of a stable data-free operator network for the three-dimensional incompressible Navier–Stokes equations—where even data-driven operator networks have reported only scarce results—jointly predicting velocity and pressure fields within a unified velocity–pressure formulation.
Finally, by combining spectral efficiency with amortized inference, SpecONet enables scalable, ensemble-level simulations that are orders of magnitude faster to infer NSE solutions than conventional numerical solvers.
Together, these capabilities position SpecONet as a general and reliable foundation for data-free operator learning and scientific simulation.

As shown in Fig.~\ref{f:schematic}a, SpecONet accepts diverse PDE inputs, including initial velocity fields $\mathbf{u}_0$, external forcing $\mathbf{f}$, and boundary conditions $\mathbf{g}$. These inputs are processed through a shallow CNN followed by fully connected layers to produce spectral coefficients ${\widehat{\pmb{\alpha}}_{lmn}}$ and ${\widehat{{\phi}}_{lmn}}$. Consequently, inferences $\widehat{\mathbf{u}}$ and $\widehat{{p}}$ are obtained after combining the coefficients and basis functions $\Psi_{lmn}$ predefined by boundary conditions. Because the basis functions are fixed a priori, the task reduces to coefficient prediction, achieving high accuracy and exactly imposing boundary conditions while keeping the architecture lightweight and fast.
Fig.~\ref{f:schematic}b illustrates the training strategy. Instead of using labeled solutions, the network is optimized through the weak formulation of the Navier–Stokes equations with a spectral element discretization and a rotational pressure-correction scheme (see Section \ref{sec_method}). Momentum balance and incompressibility residuals guide the training, enabling data-free learning that produces physically faithful and robust solutions.
Fig.~\ref{f:schematic}c demonstrates applications. SpecONet solves both 2D and 3D incompressible NSEs, handles perturbed inputs robustly, and enables efficient ensemble computing. Once trained, it rapidly generates thousands of realizations, achieving nearly fifteen-fold acceleration over conventional numerical solvers for $10,000$ 3D flow samples, thereby making large-scale uncertainty quantification and control feasible.
These characteristics position SpecONet not merely as an alternative to conventional solvers but as a general-purpose, high-fidelity surrogate for nonlinear fluid systems.
Built on a PDE-theoretic and numerically consistent foundation, it enables fast and reliable inference for ensemble computing and diverse fluid modeling tasks, paving the way for the results presented next.

\begin{figure*}[th!]
    \centering
    \includegraphics[width=\textwidth]{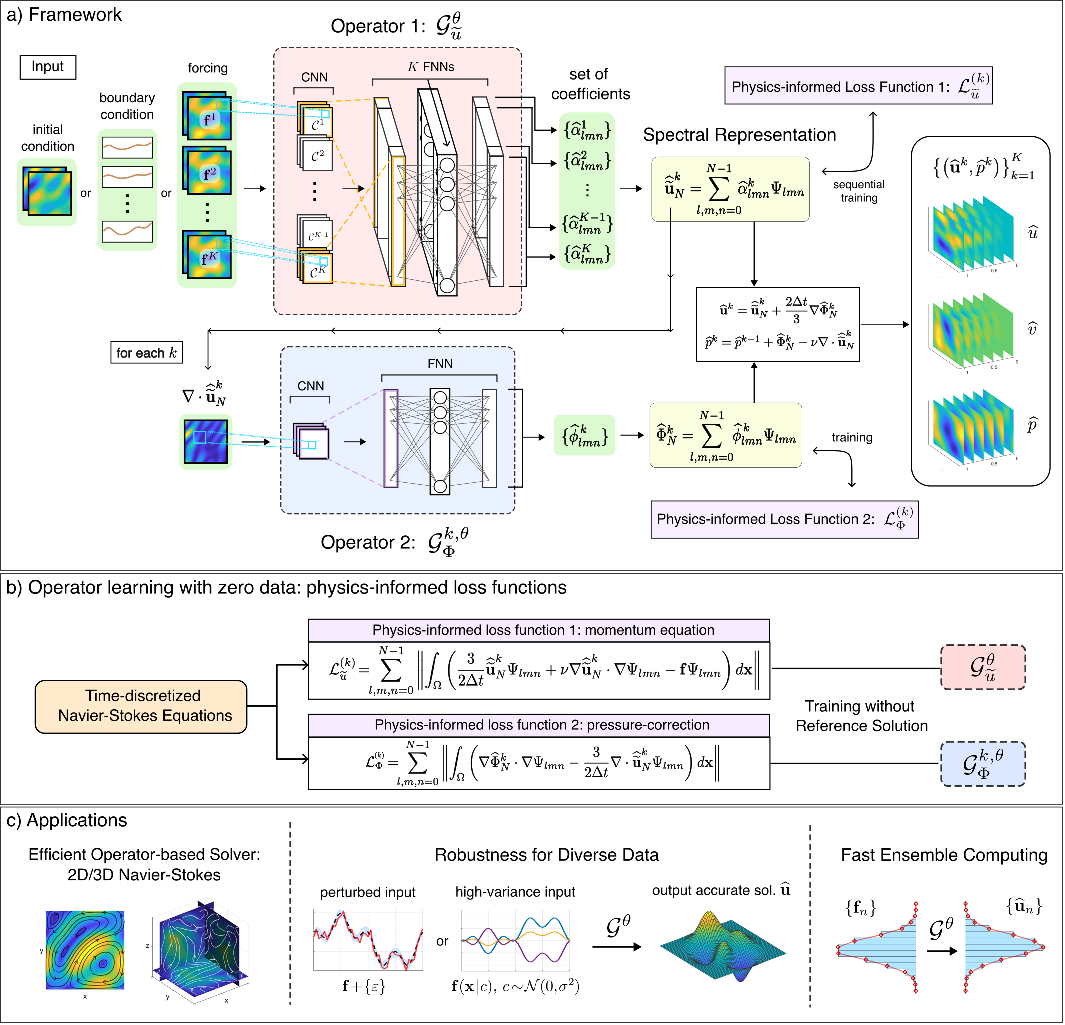}
    \caption{\textbf{Overview of the physics-informed Spectral Operator Network (SpecONet).}
\textbf{a)} Architecture of the SpecONet: The model takes initial data $\bold{u}_{0}$, boundary conditions $g$, or external forcing $\bold{f}$ as input, and outputs a set of coefficients representing the NSE solution. It comprises two operator networks, $\mathcal{G}_{\widetilde{u}}^\theta$ and $\mathcal{G}^{k,\theta}_{\Phi}$, each trained using dedicated physics-based loss functions. Once the spectral representation is reconstructed, the velocity and pressure fields can be directly inferred.
\textbf{b)} Training without reference solutions: SpecONet leverages the physics-informed loss functions derived from time-discretized NSEs via the pressure-correction method. This allows both  $\mathcal{G}_{\widetilde{u}}^\theta$ and $\mathcal{G}^{k,\theta}_{\Phi}$ to be trained without reliance on reference solutions.
\textbf{c)} Applications: The model efficiently solves 2D and 3D NSEs under a variety of conditions. It also demonstrates robustness to diverse input functions: once the model is trained, it remains accuracy even with perturbed or high-variance inputs. In addition, our model enables fast ensemble computation, producing multiple predictions simultaneously with high accuracy.}\label{f:schematic}
\end{figure*}

\section{Results}\label{sec_results}
We evaluate SpecONet on a range of incompressible NSEs to demonstrate that a fully data-free operator network can achieve high accuracy, robustness, and scalability. Across both two- and three-dimensional settings, SpecONet consistently outperforms data-driven neural operators. In 2D benchmarks, data-driven models reach comparable accuracy only when trained on large datasets, whereas their performance drops markedly under limited-data conditions. SpecONet, in contrast, attains strong accuracy and generalization without relying on any reference data, also achieving the first reliable operator-network solutions for full 3D incompressible flows even without prior operator benchmarks. Finally, its efficient data-free inference enables ensemble-based uncertainty quantification and forecasting at a fraction of the computational time of conventional numerical solvers.

\subsection{Accuracy and Robustness on Two-Dimensional Flow Benchmarks}
\label{sec:baseline_comparison}
To evaluate the effectiveness of our approach, Fig.~\ref{f:comparison} presents quantitative and qualitative comparisons of SpecONet against state-of-the-art neural operator models, including the FNO \cite{li2020fourier} and POD-DeepONet (POD-DON) \cite{lu2022comprehensive}, on the 2D incompressible NSEs. We do not compare with existing unsupervised neural operators, as their reported performance is considerably lower than data-driven models. Instead, we focus on benchmarking against competitive supervised methods and demonstrate that SpecONet achieves comparable or superior results despite not using any reference solutions.

To assess robustness and accuracy, we perform two complementary tests. The first involves clean forcing conditions as inputs, while the second introduces perturbed forcing functions. In Fig.~\ref{f:comparison}a, the inputs are clean forcing fields $\mathbf{f}$, and each model $\mathcal{G}^{\theta}$ maps these inputs to predicted velocity fields $\widehat{\mathbf{u}}$. The upper panels display the predicted velocity magnitudes and the magnitudes of pointwise errors for each model. The numerical label following each baseline (e.g., FNO ``300") indicates the number of reference solutions for training the model.
Remarkably, despite being trained with zero reference data, SpecONet achieves Rel.$L^2_x$ errors that are one to two orders of magnitude lower than those of FNO and POD-DON models trained with up to 300 reference solutions. To further highlight temporal accuracy, we plot the Rel.$L^2_x$ error at $t=0.2$, $0.6$, and $1.0$ in the lower bar charts. Across all time points, SpecONet consistently outperforms the baselines, even compared to models trained with 600 solutions. The rightmost plot of Fig.~\ref{f:comparison}a shows the error evolution over time, confirming that our method maintains lower error throughout the simulation. These results demonstrate that our unsupervised framework can rival—and even surpass—data-driven models trained with extensive data resources.

Real-world deployments often involve perturbed input functions, making robustness a critical concern. To evaluate this, we perturb the input forcing functions during inference by adding a sinusoidal disturbance as defined in  Appendix\eqref{force2d_epsilon}. Fig.~\ref{f:comparison}b summarizes this robustness experiment. All models are trained on clean data, but inference is performed on perturbed inputs. While FNO and POD-DON exhibit significant performance degradation under perturbation, SpecONet relatively remains stable and accurate. The upper-right panels showing the predicted velocity fields and their corresponding error maps illustrate that SpecONet preserves the coherent structures of the solution profile, whereas the baselines lose these structures even under mild perturbations.
Importantly, achieving comparable robustness with FNO or POD-DON would require large and diverse training datasets encompassing perturbed scenarios. In contrast, SpecONet generalizes reliably from training that is grounded in the weak formulation and numerical discretization. This highlights a key advantage of our approach, where generalization arises from the enforcement of physical and numerical structure rather than from the breadth of training data.
The entire set of experiments is provided in Appendix \ref{sub:2d_force}.

\begin{figure*}[th!]
    \centering
    \includegraphics[width=\linewidth]{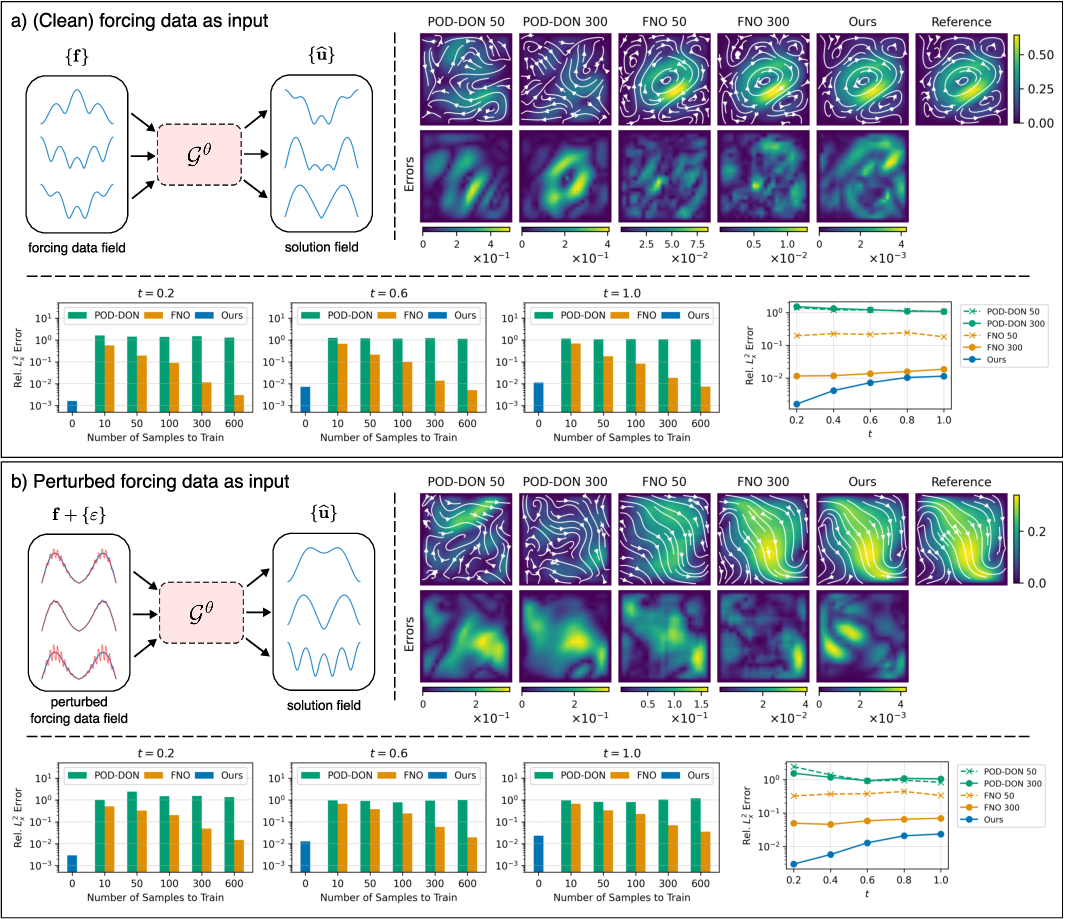}
    \caption{
\textbf{Comparison with baseline models.}
For each subfigure, \textbf{a)} and \textbf{b)}, the upper left panel illustrates the inference process, where a trained model $\mathcal{G}^\theta$—either POD-DON, FNO, or SpecONet—predicts the solution field from the forcing functions (or perturbed forcing data field in b)). 
The upper right panel shows the magnitudes and streamlines of the inferences produced by the different models (top) and the corresponding pointwise error maps between the reference and the inferences (bottom). The number following the POD-DON and FNO model names indicates the number of reference solutions used for training. The colorbar of the reference solution represents the magnitude of the velocity, $\sqrt{u^2 + v^2} $ where $\bold{u} = (u,v)$ is the velocity field. The bottom panel presents the Rel.$L^2_x$ error at time slices $t = $0.2, 0.6, and 1.0 as bar plots, with the rightmost panel showing the evolution of Rel.$L^2_x$ error over time $t$. We note that our model, trained without using reference solutions, yields more accurate results than POD-DON and FNO models even when those are trained with as many as 600 reference solutions, which require significantly more computational resources to generate the corresponding reference solutions. Notably, in b), other models produce even less accurate solutions when given perturbed input functions, whereas our model maintains high accuracy, demonstrating strong robustness to input perturbations.
    } \label{f:comparison}
\end{figure*}

\subsection{Robust Generalization Across Diverse Input Conditions}
\label{sec:input_generalization}

In Fig.~\ref{f:generalization}, we demonstrate the generalization capacity of SpecONet across diverse input modalities and statistical distributions. As shown in Fig.~\ref{f:generalization}a, a key strength of our framework lies in its ability to handle various types of PDE inputs—such as initial conditions, boundary conditions, and external forcing—without any architectural modification. This flexibility is essential for practical deployment in complex physical systems.

To assess generalization under distributional shift, we consider a controlled setup where input functions are sampled from zero-mean Gaussian random fields with varying variances. During training, the model is exposed only to low-variance samples (e.g., $\sigma_0^2 = 5^2$), while test inputs are drawn from higher-variance distributions (e.g., $\sigma_1^2 = 9^2$, $\sigma_2^2 = 13^2$, and up to $\sigma_2^2 = 20^2$ for boundary conditions). This design allows us to examine the model’s ability to generalize to increasingly complex, out-of-distribution inputs beyond the training regime, as illustrated in Fig.~\ref{f:generalization}b.

Fig.~\ref{f:generalization}c shows representative predictions across the three input types.
In the left column (initial condition inputs), SpecONet accurately predicts reference solutions even for test inputs with significantly higher variance than those seen during training.
The middle column highlights the model’s strong generalization to boundary condition inputs whose oscillation on the top boundary grows larger.
The right column demonstrates stable predictions when external forcing fields serve as inputs, indicating the model’s adaptability to unseen and distributionally shifted input patterns.

Quantitative comparisons in Fig.~\ref{f:generalization}d further support these findings. Bar plots of Rel.$L^2_x$ error at $t=0.4$ and $1.0$ show that SpecONet consistently surpasses data-driven baselines, particularly under high-variance test conditions. Remarkably, although SpecONet is trained without any reference solutions, it achieves higher accuracy than FNO and POD-DeepONet trained on high-fidelity labeled solutions.
The related experiments are reported in Appendix \ref{sub:2d_force}, \ref{sub:2d_initial}, and \ref{sub:2d_boundary}.

\begin{figure*}[th!]
    \centering
    \includegraphics[width=\linewidth]{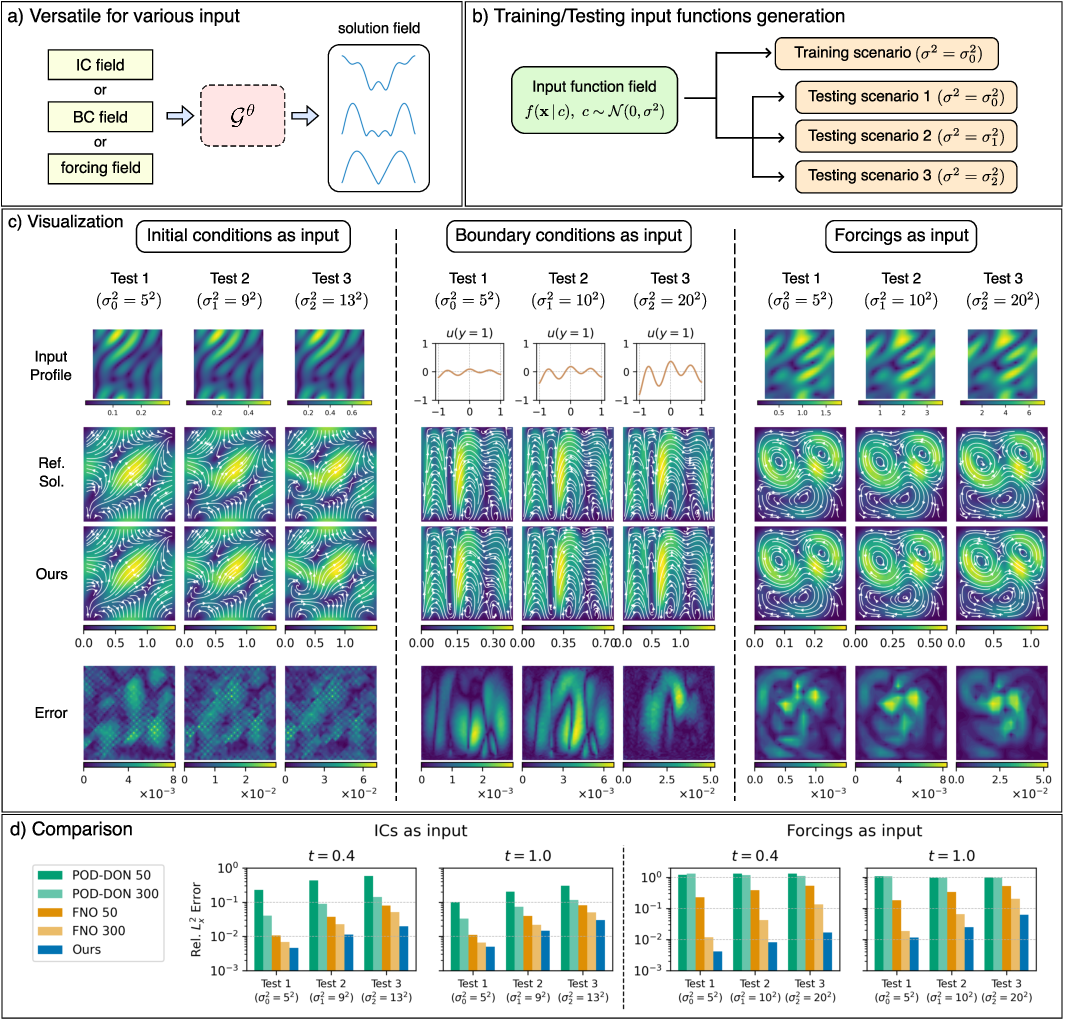}
    \caption{
    \textbf{Versatility of SpecONet for various types of input.}
\textbf{a)} SpecONet is capable of handling different types of input functions, including initial condition fields, boundary condition fields, and external forcing fields.
\textbf{b)} A set of input functions generation for training and testing: Each set is generated from a function field $f(\bold{x} \mid c)$, where the random coefficient $c$ is sampled from a normal distribution $\mathcal{N}(0, \sigma^2)$ with different variances $\sigma^2 = \sigma^2_0$, $\sigma^2_1$, $\sigma^2_2$, satisfying $\sigma^2_0 < \sigma^2_1 < \sigma^2_2$. Separate datasets are used for training and testing.
\textbf{c)} Visualization of model predictions under the three input scenarios: In each case, the model is trained using data generated with variance $\sigma^2_0 = 5^2$.
\textit{Initial condition input}: Tested on inputs generated from different variances $\sigma^2_0 = 5^2$, $\sigma^2_1 = 9^2$, and $\sigma^2_2 = 13^2$.
\textit{Boundary condition input}: Zero Dirichlet conditions are applied to the left, right, and bottom walls, while the top wall velocity is defined by a function with random coefficients sampled from distributions with $\sigma^2_0 = 5^2$, $\sigma^2_1 = 10^2$, and $\sigma^2_2 = 20^2$.
\textit{Forcing input}: The model predicts the solution field given 2D forcing inputs, evaluated with variances $\sigma^2_0 = 5^2$, $\sigma^2_1 = 10^2$, and $\sigma^2_2 = 20^2$.
In all scenarios, the model demonstrates strong predictive accuracy under varying input conditions.
\textbf{d)} Comparison with baseline models: Under different input fields, IC field and forcing field, bar plots show the Rel.$L^2_x$ error at time slices $t = 0.4$ and $t=1.0$ across different testing datasets. Our model consistently outperforms the baseline models, especially under high-variance input conditions.
    }\label{f:generalization}
\end{figure*} 

\subsection{Data-Free Operator Learning for 3D Navier–Stokes Equations}
\label{sec:beltrami}

While most existing neural operator models have demonstrated results primarily on two-dimensional fluid dynamics problems, the three-dimensional incompressible NSEs remain largely unexplored due to their substantially higher computational time and the modeling challenges inherent to operator learning. Notably, even data-driven approaches have reported only limited results on 3D NSEs, and, to our knowledge, there are no prior studies demonstrating data-free operator learning in this regime. In contrast, our framework, {SpecONet}, successfully bridges this gap by producing accurate and stable 3D solutions without relying on any precomputed solutions.

Fig.~\ref{f:beltrami} presents two representative experiments to validate SpecONet's performance on 3D NSEs under different input configurations.
In Fig.~\ref{f:beltrami}a, we consider the cases where the external forcing fields, $\mathbf{f}(\mathbf{x}|c)$ serve as inputs. We employ the model trained on $\mathbf{f}(\mathbf{x}|c)$ where $c$ were sampled from $\mathcal{N}(0, 5^2)$. The leftmost panels display two test cases where $c$ is sampled from $\mathcal{N}(0, 5^2)$ and $\mathcal{N}(0, 10^2)$, which means that they are unseen by the training input samples.
Despite being trained solely on low-variance data, SpecONet accurately predicts the velocity fields, as shown in the prediction and reference columns. The color maps represent the velocity magnitude and the corresponding pointwise error magnitude which remains small relative to the velocity magnitude even though the variance becomes larger. Rel.$L^2_{t,x}$ error bars confirm that this generalization extends consistently across all velocity components.
A distinctive feature of SpecONet lies in its ability to operate directly in the velocity–pressure formulation. Employing a projection-based architecture inspired by the classical pressure-correction schemes~\cite{guermond2004error, shen2011spectral}, the model is not limited to vorticity-based NSEs, but adaptable to velocity-pressure NSEs in periodic domains. This formulation is particularly advantageous for realistic scenarios with Dirichlet boundary conditions, which are imposed on all faces of the cubic domain in this experiment.

Fig.~\ref{f:beltrami}b focuses on the canonical benchmark of the three-dimensional Beltrami flow, where the initial condition serves as the input. SpecONet successfully reconstructs both the velocity and pressure fields at $t = 1$, yielding low pointwise errors $\mathcal{E}_{\boldsymbol{u}}$. The model is trained on Gaussian-random initial conditions with coefficients drawn from $\mathcal{N}^{\ast}(60, 10^2)$ and tested on both in-distribution and multiple variance samples drawn from $\mathcal{N}(60, 5^2)$ $\mathcal{N}(60, 10^2)$ and $\mathcal{N}(60, 20^2)$ (Appendix \ref{sub:3d_initial} provides more details). In all cases, SpecONet maintains high prediction accuracy, as indicated by the bar plots of Rel.$L_{t,x}^2$ error across velocity components. For visualization, both prediction and error maps display velocity magnitude.
Importantly, SpecONet’s predictions preserve essential physical invariants. The rightmost panel plots the temporal evolution of kinetic energy and enstrophy, comparing our predictions with the exact Beltrami solution, which is analytically available for this case. The close alignment of decay trends confirms that the learned operator reproduces key dynamical behaviors of the Beltrami flow.
Additional experiments are discussed in Appendix \ref{sub:3d_initial}.

\begin{figure*}[th!]
    \centering
    \includegraphics[width=\linewidth]{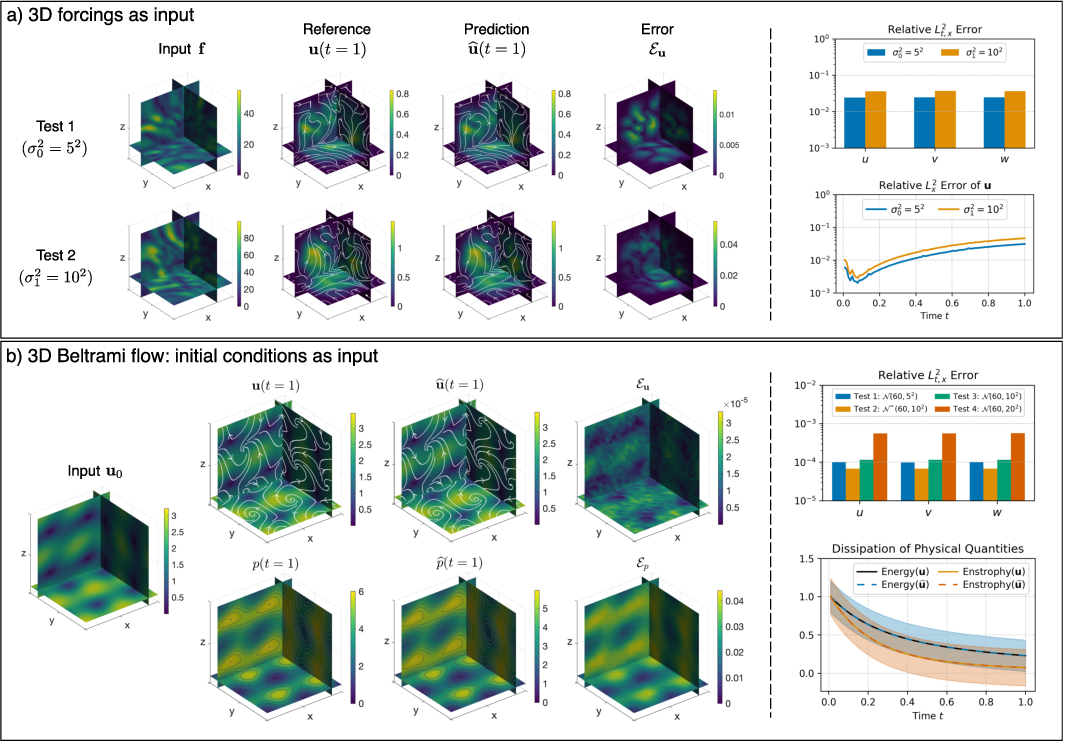}
    \caption{
\textbf{Solving 3D NSEs.} 
SpecONet solves the 3D NSEs under various input fields.
\textbf{a)} 3D forcings as input: Two different input functions, generated with $\sigma_0^2=5^2$ and $\sigma_1^2=10^2$, are utilized to evaluate our model’s performance. The cross sections and streamlines of the inferences $\widehat{\bold{u}}$ at time $t=1$, along with those of the reference solution $\bold{u}$ and the error $\mathcal{E}_{\bold{u}}:=|\bold{u} -\widehat{\bold{u}} |$, are visualized. The bar plot shows the Rel.$L^2_x$ error of each velocity components $u, v, w$ at time $t=1$. The curve plot shows the Rel.$L^2_x$ error of $\widehat{\bold{u}}$ across whole time interval.
\textbf{b)} 3D Beltrami flow: Our model takes initial conditions as input field. The inferences $\widehat{\bold{u}}$ and $\widehat{p}$ at time $t=1$ for $\mathcal{N}^{\ast}(60,10^2)$, together with the corresponding reference solutions $\bold{u}$ and $p$, and the pointwise errors $\mathcal{E}_{\bold{u}}$ and $\mathcal{E}_{p}$, are visualized. The bar plot shows the Rel.$L^2_x$ errors evaluated on testing input datasets with different levels of randomness in the initial condition. Each dataset is generated using coefficients sampled from distinct normal distributions: $\mathcal{N}(60, 5^2)$, $\mathcal{N}^{\ast}(60, 10^2)$, $\mathcal{N}(60, 10^2)$, and $\mathcal{N}(60, 20^2)$ (cf. Appendix \ref{sub:3d_initial}). The curve plot shows the physical quantities: the evolution of average kinetic energy and average enstrophy over time. The solid and dashed lines indicates the averages of the kinetic energies and enstrophies upon 100 test samples from $\mathcal{N}^{\ast}(60,10^2)$. In addition, the shading regions are the standard deviation ranges centered on the averages. The inferences exhibit physically consistent behavior, including the expected decay in both energy and enstrophy, demonstrating the model’s reliability and robustness for real-world fluid dynamics.
    }\label{f:beltrami}
\end{figure*}

\subsection{Scalable and Reliable Ensemble Prediction for 3D NSEs}
\label{sec:ensemble}

Fig.~\ref{f:ensemble} highlights the capability of SpecONet to perform fast and accurate ensemble predictions for the 3D NSEs using external forcing fields as inputs. Ensemble simulations are central to uncertainty quantification and probabilistic forecasting, particularly in fields such as weather prediction~\cite{kalnay2003atmospheric,bauer2015quiet}, biology~\cite{cao2020ensemble}, and finance~\cite{nti2020comprehensive}. However, the computational burden of producing thousands of high-fidelity simulations remains a persistent bottleneck in computational science.
SpecONet mitigates this limitation by providing a data-free, operator-based surrogate that enables rapid and consistent ensemble inference with minimal computational time.

As illustrated in Fig.~\ref{f:ensemble}a, the ensemble inputs comprise multiple realizations of three-dimensional external forcing fields ${\mathbf{f}^i}$, each sampled from a Gaussian random field with zero mean and variance $\sigma_i^2$. During training, SpecONet is exposed only to forcings ($c \sim \mathcal{N}(0, 5^2)$), but during inference, it generalizes to multiple variance forcings with $\sigma_i^2 \in [1^2, 10^2]$—without any retraining or fine-tuning. This setup emulates practical ensemble-forecasting pipelines to assess how uncertainty in the forcings propagates through a single simulator, resulting in variance of the outcomes.
Fig.~\ref{f:ensemble}b visualizes representative predicted velocity magnitudes at $t=0.5$ for different variance levels of the forcing distribution. Each field corresponds to a distinct random realization, and SpecONet maintains consistent prediction quality across the ensemble despite the increased input variability. This behavior reflects the model’s strong generalization capability across high-dimensional and distributionally shifted input spaces. 
To assess ensemble-level statistical behavior, we compute a quantity of interest $Q := \int_{\Omega} \widehat{u}(t,\cdot),d\mathbf{x}$ at $t=1$ for each realization and analyze its empirical distribution across ensemble sizes $S \in \{100, 500, 1000 \}$ referring to \cite{luo2018ensemble}. As shown in Fig.~\ref{f:ensemble}c, the histogram of $Q$ converges to a Gaussian profile as $S$ increases, consistent with the Central Limit Theorem. This convergence behavior is consistent across three  input distributions with variances $\sigma^2 = {1^2, 2^2, 10^2}$. Notably, the convergence toward normal distributions as $S$ gets larger indicates that the operator captures both the variability and the underlying statistical structure of the ensemble dynamics faithfully.
Fig.~\ref{f:ensemble}d provides quantitative comparisons. The top panel reports the Rel.$L^2_{t,x}$ errors of the velocity components $u$, $v$, $w$, and the pressure gradient $\nabla p$ across the three forcing distributions, each corresponding to distinct forcing distributions. In all cases, SpecONet maintains reliable error even under increasing variance, confirming its robustness to input uncertainty (The detailed experiments are articulated in Appendix \ref{sub:3d_force}).
The middle panel compares the total inference time of SpecONet with that of a conventional spectral NSE solver as the ensemble size increases from $S = 10^2$ to $S = 10^4$. Although both scale linearly with $S$, SpecONet achieves approximately a 15× speedup at $S = 10{,}000$, owing to its amortized inference time and compatibility with batch-parallel GPU execution. This efficiency gap is expected to widen further with increasing ensemble size, making the framework particularly suitable for real-time forecasting and risk-sensitive simulations requiring massive forward evaluations.
Finally, the bottom panel in Fig.~\ref{f:ensemble}d tracks the convergence of the ensemble-averaged velocity field $\bar{{u}}^S := \frac{1}{S}\sum_{s=1}^S \hat{{u}}^s$ toward the high-sample reference $\bar{{u}}^{10{,}000}$. The corresponding error $\|\bar{{u}}^S - \bar{{u}}^{10{,}000}\|_{L^2_x}$ scales empirically as $\mathcal{O}(1/\sqrt{S})$, consistent with theoretical expectations under i.i.d. sampling. This confirms both the stability and statistical reliability of SpecONet in ensemble settings.

\begin{figure*}[th!]
    \centering
    \includegraphics[width=\linewidth]{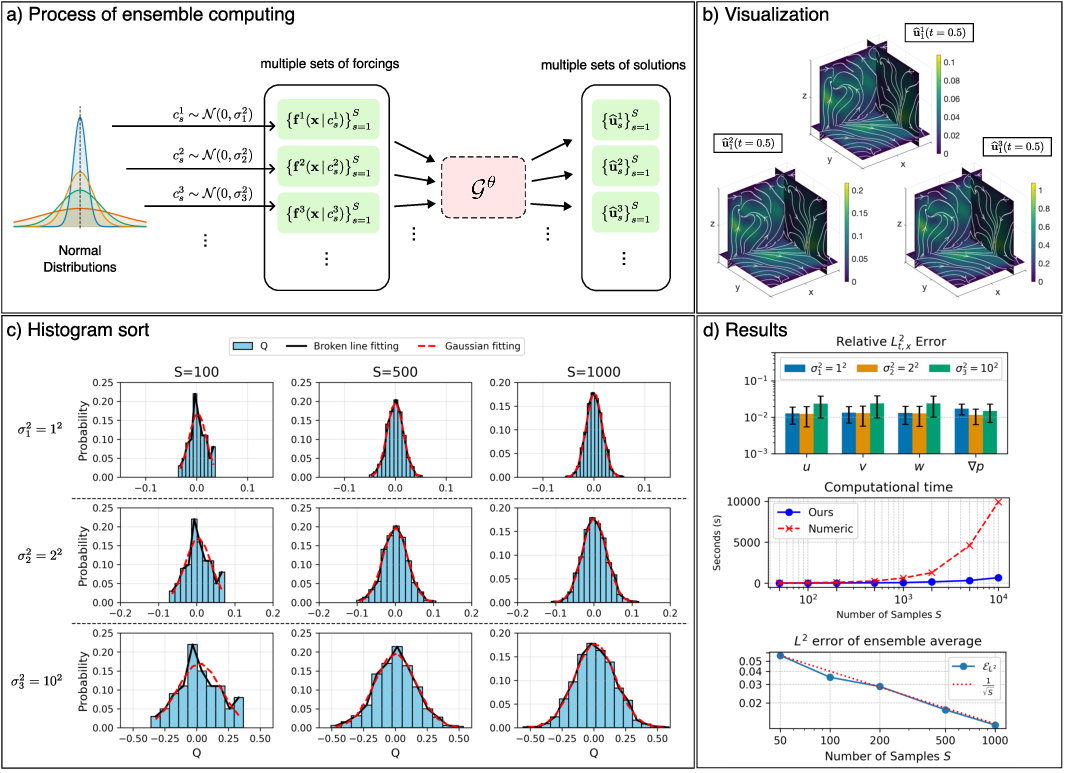}
    \caption{
\textbf{Fast ensemble computing.} 
\textbf{a)} Illustration of the ensemble computing process: The sets $\{\mathbf{f}^i\}$ of input forcing functions are generated from different normal distributions for $i=1,2,3$. Input forcing functions in each set depend on  random coefficients $c^i_s$ drawn from a normal distribution with variances $\sigma^2_i$. By inputting the sets of forcings $\{\mathbf{f}^i\}$, the trained model $\mathcal{G}^\theta$ simultaneously infers multiple solution sets $\{\widehat{\mathbf{u}}^i\}$.
\textbf{b)} The cross sections and stream lines of predicted velocity fields at $t = 0.5$ from $\mathcal{N}(0,1^2)$, $\mathcal{N}(0,2^2)$, and $\mathcal{N}(0,10^2)$.
\textbf{c)} Histogram sorting of ensemble outputs: For each solution set $\{\widehat{{u}}^i\}$ for $t=1$, we compute the quantity of interest $Q = \int_{\Omega} \widehat{{u}} \, d\bold{x}$ for each sample. The histograms show the sorted values of $Q$ for $\{\widehat{{u}}^1\}$, $\{\widehat{{u}}^2\}$, and $\{\widehat{{u}}^{3}\}$, corresponding to variances $\sigma^2_1 = 1$, $\sigma^2_2 = 2$, and $\sigma^2_3 = 10$, respectively. As the number of samples $S$ increases, the distribution of $Q$ converges toward a Gaussian distribution.
\textbf{d)} The top panel shows bar plots of the Rel.$L_{t,x}^2$ errors of the velocity components and pressure gradient across different solution sets $\{\widehat{\mathbf{u}}^1\}$, $\{\widehat{\mathbf{u}}^2\}$, and $\{\widehat{\mathbf{u}}^{3}\}$. The middle panel compares the computational time of our method with conventional numerical solvers. The bottom panel illustrates the convergence behavior of the ensemble-averaged error, defined as $\mathcal{E}_{L^2_x} := \left\| \bar{{u}}^S - \bar{{u}}^{10000} \right\|_{L^2}$ at $t=1$, where $\bar{{u}}^S = \frac{1}{S} \sum_{s=1}^{S} \widehat{{u}}^n$. The error of ensemble average coverages on the order of $\mathcal{O}(1/\sqrt{S})$.
    }\label{f:ensemble}
\end{figure*} 

\section{Discussion}
In this work, we introduced SpecONet, a spectral operator network capable of solving parametric incompressible NSEs in both two and three dimensions without employing any reference solutions. By embedding spectral element discretization directly into the network architecture and training it through a variational loss derived from time-discretized NSEs, SpecONet provides a highly accurate, stable, and fully data-free alternative to existing neural operator frameworks.

Our model addresses several long-standing limitations of operator-learning approaches. Most notably, it eliminates the dependence on large reference solutions that characterize data-driven models such as FNO and DeepONet. In regimes where obtaining high-fidelity solutions is computationally prohibitive—particularly in three-dimensional fluid simulations—this data-free capability offers a major practical advantage.
To our knowledge, SpecONet represents the first demonstration of a stable data-free operator network successfully solving the 3D NSEs, a regime in which even data-driven neural operators have reported only scarce results. Furthermore, unlike many operator networks restricted to two-dimensional flows or vorticity-based approximations under periodic boundaries~\cite{kovachki2023neural,tran2021factorized,ovadia2024vito}, SpecONet jointly predicts velocity and pressure fields within a unified velocity–pressure formulation. This enables stable simulation under general boundary conditions, including Dirichlet boundaries, where enforcing physical constraints near walls is often challenging.

Another key strength of SpecONet lies in its robustness and generalization. The framework accurately handles diverse input modalities—initial, boundary, and forcing conditions—without requiring architectural modification.
It maintains high predictive fidelity even under significantly perturbed inputs or distributional shifts, outperforming data-driven baselines trained with large labeled solutions. These properties extend naturally to ensemble scenarios, where SpecONet enables high-throughput computation and achieves a 15-fold speedup over conventional solvers when generating 10,000 three-dimensional flow realizations. This scalability and stability mark a significant step toward practical ensemble-based forecasting, optimization, and uncertainty quantification in fluid dynamics.

Important directions remain for future work. On the theoretical front, a key goal is to establish a rigorous convergence theory for SpecONet through the lenses of approximation, generalization, and optimization, bridging classical numerical analysis with modern learning theory to derive error bounds and stability guarantees \cite{lu2021learning, kovachki2021universal, lee2025finite}.
From a practical standpoint, SpecONet can be extended to more complex and coupled physical systems. Its architecture and training methodology make it naturally compatible with multi-scale and multi-physics frameworks, such as those used in numerical weather prediction and climate modeling~\cite{lynch2008origins,stocker2014climate,bauer2015quiet}.
By serving as a differentiable, data-free surrogate for expensive simulation components, SpecONet has the potential to accelerate large-scale forecasting systems while maintaining physical consistency.

\section{Methods}\label{sec_method}
We describe here the framework of our model, SpecONet. We begin by introducing the rotational pressure-correction method and spectral element schemes that are the foundation of our network architecture and the associated loss functions. The details of the architecture and the training procedure are then presented.

Throughout this article, we consider the incompressible NSEs, which read:
\begin{align}\label{NSE}
\begin{split}
{\bf{u}}_t+{\bf{u}}\cdot\nabla{\bf{u}}-\nu \Delta{\bf{u}}+\nabla p ={\bf{f}}&,\quad \text{for}\quad t>0, \quad x\in \Omega,\\ 
\nabla\cdot \bold{u}=0&, \quad \text{for}\quad t>0, \quad x\in \Omega,\\
{\bf{u}}={\bf{u}}_0&,\quad \text{for}\quad t=0,\quad x\in \Omega,
\end{split}
\end{align}
given a certain boundary condition, where the domain $\Omega \in \mathbb{R}^{d}$, $d=2$ or $3$, and $\nu > 0$ denotes fluid viscosity. 

SpecONet leverages time-discretized NSEs by applying the rotational pressure-correction scheme to \eqref{NSE}. This scheme is chosen for its temporal accuracy and efficiency in enforcing the divergence-free condition, $\nabla \cdot \bold{u} = 0$, while reducing errors inherent in conventional pressure-correction methods. For spatial discretization, the spectral element method is then employed, providing accurate solution representation and ensuring the given boundary conditions. These schemes are outlined in the following sections.

\subsection{Temporal Scheme: Rotational Pressure-correction Method}\label{sec_temporal}
We first introduce the rotational pressure-correction method \cite{guermond2004error,shen2011spectral}, which achieves second-order accuracy in velocity during time marching and reduces errors inherent to conventional pressure-correction processes. {These advantages inspire the architecture of our model, which is built to emulate this method.} 

The procedure consists of two steps. In the first step, the intermediate velocity $\widetilde{\bold{u}}^{k+1}$ at the $(k+1)$-th time step is obtained from the momentum equation:
\begin{align}\label{NSE_first}
\frac{1}{2\Delta t}(3\widetilde{\bold{u}}^{k+1}-4\bold{u}^{k}+\bold{u}^{k-1})-\nu\Delta\widetilde{\bold{u}}^{k+1}+\nabla p^{k}=\bold{g}(t^{k+1}), \quad \widetilde{\bold{u}}^{k+1}|_{\partial\Omega}=0,
\end{align}
where 
\begin{align}
\bold{g}(t^{k+1})=\bold{f}(t^{k+1})-(2(\bold{u}^{k}\cdot\nabla) \bold{u}^{k}-(\bold{u}^{k-1}\cdot\nabla) \bold{u}^{k-1}), 
\end{align}
with the given forcing $\bold{f}(t^{k+1})$ and the previous time-step solutions $\bold{u}^{k-1}$, $\bold{u}^{k}$, and $p^{k}$. 

The second step is a projection that maps the intermediate velocity onto the divergence-free space, based on the Helmholtz decomposition as
\begin{align}\label{decompse}
&\widetilde{\bold{u}}^{k+1}={\bold{u}}^{k+1}-\frac{2\Delta t}{3}\nabla\Phi^{k+1}, \\
&\nabla\cdot {\bold{u}^{k+1}} = 0, \quad {\bold{u}^{k+1}}\cdot \bold{n}|_{\partial \Omega} = 0.
\end{align}
To solve $\Phi^{k+1}$, we take divergence on \eqref{decompse} and lead a Poisson equation
\begin{align}\label{NSE_second}
&\Delta\Phi^{k+1} = \frac{3}{2\Delta t}\nabla\cdot\widetilde{\bold{u}}^{k+1}, \quad \left. \frac{\partial\Phi}{\partial \bf{n}} \right|_{\partial \Omega} =0.
\end{align}

After obtaining $\widetilde{\bold{u}}^{k+1}$ and $\Phi^{k+1}$ by solving the equations \eqref{NSE_first} and \eqref{NSE_second} respectively, the velocity field ${\bold{u}}^{k+1}$ can be directly inferred from the relation \eqref{decompse} as
\begin{align}\label{solution_first}
\bold{u}^{k+1}=\widetilde{\bold{u}}^{k+1}+\frac{2\Delta t}{3}\nabla\Phi^{k+1}.
\end{align}

Subsequently, the pressure field $p^{k+1}$ can be updated as follows. we first note that \eqref{decompse} and \eqref{NSE_second} yields 
\begin{align}\label{NSE_third}
\begin{split}
\Delta \widetilde{\bold{u}}^{k+1}&= \Delta {\bold{u}}^{k+1} -\frac{2\Delta t} {3}\nabla\Delta\Phi^{k+1}=\Delta {\bold{u}}^{k+1} -\nabla\nabla\cdot\widetilde{\bold{u}}^{k+1}.
\end{split}
\end{align}
When summing up \eqref{NSE_first}, and \eqref{NSE_third} and using \eqref{decompse}, it gives 
\begin{align}
\begin{split}
&\frac{1}{\Delta t}(3{\bold{u}}^{k+1}-4\bold{u}^{k}+\bold{u}^{k-1})-\nu\Delta{\bold{u}}^{k+1}+\nabla (p^{k}+\Phi^{k+1}-\nu \nabla\cdot\widetilde{\bold{u}}^{k+1} )=\bold{g}(t_{k+1}).
\end{split}
\end{align}   
Therefore, the pressure at the next time step is defined as
\begin{align}\label{solution_second}
p^{k+1}=p^{k}+\Phi^{k+1}-\nu\nabla\cdot\widetilde{\bold{u}}^{k+1}.
\end{align}
Once again, using \eqref{solution_first} and \eqref{solution_second}, the scheme at the $(k+2)$-th time step starts.

We now discuss how to apply spectral methods to our network.

\subsection{Spatial Scheme: Spectral Element Methods}\label{sec_spectral}
In the previous subsection, the rotational pressure-correction method reduced the NSEs to two Helmholtz-type equations, \eqref{NSE_first} and \eqref{NSE_third}. Here, we outline how such equations are solved using the spectral method, and subsequently describe its application within our scheme.

For convenience, we denote a generic Helmholtz-type equation as
\begin{align}\label{Helm}
    \tau u -\nu \Delta u= f,
\end{align}
subject to a certain boundary condition. Then, we derive the weak formulation of \eqref{Helm} in a Hilbert space $H^1$. Multiplying a test function $\overline{\Psi} \in H^1$, integrating over $\Omega$, and applying integration by parts yield
\begin{align}\label{weak_form}
    \tau\int_{\Omega} u \overline{\Psi} d{\bf{x}}+\nu\int_{\Omega} \nabla u\cdot\nabla \overline{\Psi} d{\bf{x}}=\int_{\Omega}f \overline{\Psi} d{\bf{x}}.
\end{align}
According to the Galerkin approach, a weak solution $u$ to \eqref{weak_form} can be approximated as $\sum_{n=0}^{N-1} \alpha_n \psi_n$, where $\{\alpha_{n}\}_{n=0}^{N-1}$ is a set of coefficients corresponding to a set of basis (trial) functions $\{\psi_n\}_{n=0}^{N-1}$ in $H$. Different choices of basis functions give rise to numerical schemes such as the finite element, spectral element, or finite volume methods. Here, we employ spectral methods, whose bases are global functions—Legendre polynomials, Chebyshev polynomials, or Fourier modes—that achieve high accuracy with relatively few nodal points, thereby reducing computational time (see, e.g., \cite{shen2011spectral,evans2022partial,trefethen2000spectral}). Using this approach, the approximation of $u$ in $H$ is represented as
\begin{align}\label{sm}
    u_N(x,y,z) =\sum_{l,m,n=0}^{N-1}\alpha_{lmn}\Psi_{lmn}(x,y,z),
\end{align} 
where $\alpha_{lmn}$ are spectral coefficients, $\Psi_{lmn}(x,y,z):=\psi_l(x)\psi_n(y)\psi_m(z)$ are basis functions, and $N$ is the number of basis functions. The choice of basis functions $\Psi_{lmn}$ depends on the boundary condition. In this work, we focus on Dirichlet, Neumann, and periodic boundary conditions for constructing the bases.

For Dirichlet or Neumann boundary conditions, we employ Legendre polynomials on $x \in [-1,1]$ to make basis functions \cite{shen1994efficient,shen2011spectral}:
\begin{align}\label{Legendre_poly}
    \psi_n(x) =\begin{cases} \frac{1}{\sqrt{4n+6}}(L_n(x) - L_{n+2}(x)),~\text{if Dirichlet condition is given},\\
    \frac{1}{\sqrt{b_n(4n+6)}}(L_n(x) -b_n L_{n+2}(x)) ,~\text{if Neumann condition is given}\end{cases}
\end{align}
where \( L_n \) represents the Legendre polynomials of degree \( n \), and $b_n=n(n+1)/((n+2)(n+3))$. Note that $\psi_n$ or $d\psi_n/dx$ in \eqref{Legendre_poly} are zero at $x=\pm1$ when a Dirichlet condition or a Neumann condition is given, respectively. In addition, the test functions in $H$ are identical to \eqref{Legendre_poly}.  

For periodic boundary conditions (cf. \cite{trefethen2000spectral}), the basis functions are the periodic Fourier modes, given by:
\begin{align}\label{period_base}
    \psi_\xi(x)=\frac{1}{2\pi}e^{i\xi x}~\text{on}~[0,2\pi),
\end{align}
where $i$ is the imaginary unit, and $\xi$ is a wave number. Besides, the test functions in $H$ are defined by 
\begin{align}
    \overline{\psi}_\xi(x)=e^{-i\xi x}~\text{on}~[0,2\pi). 
\end{align}

The remaining task is to determine the spectral coefficient $\alpha_{lmn}$ in \eqref{sm}. This requires the computation of the mass and stiffness matrices from \eqref{weak_form}, \eqref{sm}, \eqref{Legendre_poly}, and \eqref{period_base}, which involves linear algebra techniques such as eigenvalue decomposition and preconditioning. As the details depend on the spatial dimension of the domain, they are provided in Appendix \ref{sec:methodology}.

By combining the spectral method with the rotational pressure-correction scheme introduced in Section \ref{sec_temporal}, we construct and train our network to infer solutions of the NSEs, as described in Section \ref{sec_structure}.

\subsection{Details on SpecONet}\label{sec_structure}
In this section, we present the design and training framework of SpecONet. The architecture is constructed by incorporating the temporal and spatial schemes introduced in Sections \ref{sec_temporal} and \ref{sec_spectral} to solve the incompressible NSEs \eqref{NSE}. The training procedure is described thereafter. The overall architecture is illustrated in Fig.~\ref{f:schematic}

Our model consists of two sub-operator networks, denoted $\mathcal{G}_{\widetilde{u}}^\theta$ and $\mathcal{G}_{\Phi}^{k,\theta}$. Each network predicts spectral coefficients arising from the spectral representation of the intermediate velocity $\widetilde{\bold{u}}^{k}$ and the correction variable $\Phi^{k}$, respectively:
\begin{align}\label{velocity_spectral}
	\widetilde{\bold{u}}_{N}^{k} = \sum_{l,m,n=0}^{N-1} \pmb{\alpha}^{k}_{lmn} \Psi_{lmn},
\end{align}
and
\begin{align}\label{pressure_spectral}
	\Phi_{N}^{k} = \sum_{l,m,n=0}^{N-1} \phi^{k}_{lmn} \Psi_{lmn}.
\end{align}
Here, the superscript $k$ denotes the solution at $t = k \Delta t$ step for $k=1,2,\cdots, K$ where $\Delta t$ is the size of a time step, and $K$ is the number of time steps. For convenience, the coefficient $\pmb{\alpha}^{k}_{lmn}$ in \eqref{velocity_spectral} is two- or three-dimensional, depending on the spatial dimension under consideration.

The first operator $\mathcal{G}_{\widetilde{u}}^{\theta}$ is designed to map a set of time-dependent input functions to a corresponding set of coefficients:
\begin{align}\label{network_first}
  \mathcal{G}_{\widetilde{u}}^\theta:\{\mathbf{f}^k\}_{k=1}^{K} \mapsto \{\widehat{\pmb{\alpha}}^{k}_{lmn}\}_{k=1}^{K},
\end{align}
Note that the input functions $\{\mathbf{f}^k\}_{k=1}^{K}$ can be initial data (when $\mathbf{f}^k=\mathbf{u}_0$ for all $k$), boundary conditions, or forcing terms. The network architecture consists of a single convolutional neural network (CNN) with a Swish activation function ($\sigma$), followed by $K$ distinct fully connected neural networks (FNNs) (see Fig.~\ref{f:schematic}a). To achieve the desired mapping, the CNN is employed to compute a feature representation $\mathcal{C}^{k}$ for each input $\mathbf{f}^k$: 
\begin{align}
\mathcal{C}^k:=\sigma(b+\mathcal{K}\star \mathbf{f}^{k}),~\text{for all $k=1,2,\cdots, K$,}
\end{align}
where $\mathcal{K}$ and $b$ are the convolving kernel and bias of the CNN, respectively, and $\star$ denotes the multi-dimensional cross-correlation operator. Subsequently, the $k$-th FNN performs a linear mapping from the feature representation $\mathcal{C}^k$ to the coefficient set $\widehat{\pmb{\alpha}}^{k}_{lmn}$ as:
\begin{align}
 \widehat{\pmb{\alpha}}^{k}_{lmn}=\mathcal{C}^k\mathcal{W}^k,
\end{align}  
 Here, $\mathcal{W}^k$ represents the 2D weight tensor of the $k$-th FNN. By stacking these individual weight tensors $\mathcal{W}^k$ along the $k$ dimension to form a single 3D tensor, denoted by $\mathcal{W}$, the mapping for all $K$ time steps can be efficiently expressed as a single tensor operation:
\begin{align}
\{\widehat{\pmb{\alpha}}^{k}_{lmn}\}_{k=1}^{K}=\sigma(b+\mathcal{K}\star \{\mathbf{f}^k\}_{k=1}^{K})\mathcal{W},
\end{align} 
which formally represents the complete operation conducted in \eqref{network_first}. As a result, the predicted velocity fields for all time steps, $\{\widehat{\widetilde{\bold{u}}}_{N}^{k}\}_{k=1}^K$, are obtained by applying the resulting coefficient set $\{\widehat{\pmb{\alpha}}^{k}_{lmn}\}_{k=1}^{K}$ to the spectral representation \eqref{velocity_spectral} (Sec.\ref{sec:architecture} in Appendix specifically explains the architecture of SPecONet.).

The associated loss function for the operator $\mathcal{G}_{\widetilde{u}}^\theta$ is designed according to the weak form of the time-discretized momentum equation \eqref{NSE_first}, given by \eqref{weak_form}. Subsequently, the loss function for $\mathcal{G}_{\widetilde{u}}^\theta$ is defined as:
\begin{align}\label{loss_legendre}
       \mathcal{L}^{(k+1)}_{\widetilde{u}}=\sum_{l,m,n=0}^{N-1} &\left| \int_{\Omega}\frac{1}{2\Delta t}(3\widehat{\widetilde{\bold{u}}}^{k+1}_N-4\widehat{\bold{u}}^{k}+\widehat{\bold{u}}^{k-1})\overline{\Psi}_{lmn} d\bold{x} +\nu \int_{\Omega} \nabla\widehat{\widetilde{\bold{u}}}^{k+1}_N\cdot\nabla \overline{\Psi}_{lmn} d\bold{x} \right. \nonumber \\
       &\left.  + \int_{\Omega}\nabla \widehat{p}^{k} \overline{\Psi}_{lmn} d\bold{x}
       -\int_{\Omega} \bold{g}^{k+1} \overline{\Psi}_{lmn} d\bold{x}\right|^2,
\end{align}
where
\begin{align}\label{loss_force}
\bold{g}^{k+1}=\bold{f}(t^{k+1})-(2(\widehat{\bold{u}}^{k}\cdot\nabla) \widehat{\bold{u}}^{k}-(\widehat{\bold{u}}^{k-1}\cdot\nabla) \widehat{\bold{u}}^{k-1}).
\end{align}
Based on the design in \eqref{loss_legendre}, sequential training is employed to guide the learning of the network $\mathcal{G}_{\widetilde{u}}^\theta$. The detailed training procedure will be discussed below.

The second operator network, $\mathcal{G}_{\Phi}^{k,\theta}$, is designed to compute the pressure correction related with \eqref{NSE_second} for each time step $k=1,\cdots,K$. Once $\widehat{\widetilde{\bf{u}}}_{N}^{k}$ are reconstructed through the first operator $\mathcal{G}_{\widetilde{u}}^{\theta}$, $\mathcal{G}_{\Phi}^{k,\theta}$ takes $\nabla\cdot\widehat{\widetilde{\bf{u}}}_{N}^{k}$ as an input, and maps it to the spectral coefficients of the correction variable, i.e.,
\begin{align}\label{network_second}
\mathcal{G}_{\Phi}^{k,\theta} :\nabla\cdot\widehat{\widetilde{\bf{u}}}_{N}^{k} \mapsto 
\widehat{\phi}^k_{lmn}.  
\end{align}
This network for each $k$ employs a CNN with a Swish activation function, $\sigma$, and an FNN. Specifically, the CNN and FNN realize \eqref{network_second} as   
\begin{align}
\widehat{\phi}^k_{lmn}=\sigma(b^k+\mathcal{K}^k\star (\nabla\cdot\widehat{\widetilde{\bf{u}}}_{N}^{k}))\mathcal{W}^k,
\end{align} 
where $\mathcal{K}^k$ is a convolving kernel of the CNN, $b^k$ is a bias of the CNN, $\star$ is the multi-dimensional cross-correlation operator, and $\mathcal{W}^k$ is a weight of FNN. 

In a manner similar to deriving \eqref{loss_legendre}, the loss function for training $\mathcal{G}_{\Phi}^{k,\theta}$ is defined based on the weak form of the equation in \eqref{NSE_second}, as follows:
\begin{align}\label{loss_second}
   \mathcal{L}^{(k+1)}_{\Phi}=\sum_{l,m,n=0}^{N-1} \left|\int_{\Omega}\nabla\widehat{\Phi}^{k}_N\cdot\nabla\overline{\Psi}_{lmn}+ \frac{3}{2\Delta t}\nabla\cdot\widehat{\widetilde{\bold{u}}}^{k}_N\overline{\Psi}_{lmn}d\bold{x}\right|^2.
\end{align}

Owing to the design of $\mathcal{G}_{\widetilde{u}}^\theta$ and $\mathcal{G}_{\Phi}^{k,\theta}$, the predicted velocity-pressure fields, $\{ (\widehat{\bold{u}}^k, \ \widehat{p}^{k}) \}^{K}_{k=1}$, can be directly inferred using the relations \eqref{solution_first} and \eqref{solution_second}, alongside the spectral representations \eqref{velocity_spectral} and \eqref{pressure_spectral}.

We now detail the sequential training procedure, which follows \cite{choi2024spectral} with the extension. Starting from $k=0$, where the initial data ($\bold{u}_0$, $p_0$) is provided, we first compute the artificial preceding velocity $\bold{u}^{-1}$ from the Stokes equation:
\begin{align*}
\frac{1}{\Delta t}({\bold{u}}^{0}-\bold{u}^{-1})-\nu\Delta\bold{u}^{0}+\nabla p^{0}=\bold{f}(t^0),
\end{align*}
This calculated $\bold{u}^{-1}$ is then used to determine the known terms within the loss function $\mathcal{L}^{(1)}_{\widetilde{u}}$, as defined in \eqref{loss_legendre}. The training of $\mathcal{G}_{\widetilde{u}}^\theta$ subsequently begins by optimizing its CNN and the first FNN until the loss $\mathcal{L}^{(1)}_{\widetilde{u}}$ plateaus. This yields the first predicted intermediate velocity, $\widehat{\widetilde{\bold{u}}}^{1}_{N}$. Next, by computing the divergence of this first prediction, $\nabla \cdot \widehat{\widetilde{\bold{u}}}^{1}_{N}$, the training of the first pressure correction operator, $\mathcal{G}_{\Phi}^{1, \theta}$, commences. This step continues until its associated loss, $\mathcal{L}^{(1)}_{\Phi}$, is minimized. Consequently, the predicted velocity and pressure fields, $(\widehat{\bold{u}}^1, \ \widehat{p}^{1})$, are directly obtained from the relations \eqref{solution_first} and \eqref{solution_second}. These resulting fields are then prepared to form the term $\bold{g}^{2}$ according to \eqref{loss_force}.

For the second time step, $k=1$, the parameters of the CNN (optimized in the previous step) are frozen. The second FNN of $\mathcal{G}_{\widetilde{u}}^\theta$ is then trained with its corresponding loss, $\mathcal{L}^{(2)}_{\widetilde{u}}$, until convergence. Similarly, the pressure correction operator $\mathcal{G}_{\Phi}^{2, \theta}$ is then trained by inputting $\nabla \cdot \widehat{\widetilde{\bold{u}}}^{2}_{N}$ and minimizing the associated loss $\mathcal{L}^{(2)}_{\Phi}$. This two-stage procedure is repeated sequentially for subsequent time steps until $k$ reaches $K$. In the practical training process, after setting $K$ to 10, we employed new networks of $\mathcal{G}_{\widetilde{u}}^\theta$, and then trained each one for its block of $K$ time steps.

It is important to note that, thanks to the spectral analysis \cite{shen2011spectral,trefethen2000spectral}, the derivatives, including divergence $(\nabla \cdot)$ and gradient $\nabla$, applied to variables involved in the loss functions achieve spectral accuracy with a minimal number of collocation points. On the other hand, the traditional PINNs method often requires numerous collocation points and is sensitive to their specific distribution to achieve higher accuracy (see \cite{yu2022gradient,shin2020convergence}). These advantages collectively render our model more efficient and highly accurate to execute heavy computation such as 3D computing and ensemble computing. In practice, the integrals involved in the loss function are evaluated using numerical integration techniques, depending on the domain and the boundary conditions. Details regarding the implementation for different problems, including the chosen hyperparameters and the numerical strategies used to compute the loss, are provided in Appendix \ref{sec:hyper-parameter} and \ref{sec:methodology}, respectively.

\backmatter






\section*{Acknowledgements}
The work of Y. Hong was supported by the Basic Science Research Program through the National Research Foundation of Korea (NRF) funded by the Korea government (MSIT) (RS-2023-00219980), and by the Institute of Information \& Communications Technology Planning \& Evaluation (IITP) grant funded by the Korea government (MSIT) [NO.RS-2021-II211343, Artificial Intelligence Graduate School Program (Seoul National University)].
This study was supported by the National Research Foundation of Korea (NRF) grant funded by the Korean government (MSIT) (No. 2022R1C1C1009387, No. 2022R1A4A3033320). The work of N. Kim was supported by the National Research Foundation of Korea (NRF) grant funded by the Korea government (MSIT) (RS-2025-16069590).

\begin{center}
    {\Large \textbf{Supporting Information}}\\
    {\large (Supplementary Material for the Main Article)}
\end{center}
\begin{appendices}

\section{Nomenclature}
\begin{table}[h!]\label{notation}
\centering
\begin{tabular}{c l}
 \hline
 Notation & Description \\ \addlinespace
 \hline
 ${\bf{u}}=(u,v,w)$, $p$ &  A solution to Navier-Stokes equation\\\addlinespace
 $\widehat{\bf{u}}=(\widehat{u},\widehat{v},\widehat{w})$, $\widehat{p}$  & An inference of a solution to Navier-Stokes equation \\\addlinespace
 $\widetilde{\bf{u}}=(\widetilde{u},\widetilde{v},\widetilde{w})$  & An numerical solution to \eqref{NSE_first}\\\addlinespace
 $\Phi$  & An numerical solution to \eqref{NSE_second}  \\\addlinespace
 $\pmb{\alpha}$, $\phi$ & Spectral coefficients (see \eqref{velocity_spectral}, and \eqref{pressure_spectral})\\\addlinespace
      $\alpha$, $\beta$, $\gamma$,& Scalar components of $\pmb{\alpha}$ \\\addlinespace
 $\widehat{\pmb{\alpha}}$, $\widehat{\phi}$ &  Outputs of networks which predicts $\pmb{\alpha}$ and $\phi$ (see   \eqref{network_first}, and \eqref{network_second})  \\
\addlinespace
  $x_n$ & The $n$th nodal point on a spatial domain.  \\\addlinespace
  $\xi_n$ & The $n$th wave number in Fourier space. \\\addlinespace
 $\Psi_n(x)=(\psi_n^x,\psi_n^y,\psi_n^z)$ & A basis (trial) function of $n$th order \\\addlinespace
 $e^{\rm{i}\xi_n x}$ & A Fourier basis (trial) function \\ 
\addlinespace
 $\overline{\Psi}_n(x)=(\overline{\psi}_n^x,\overline{\psi}_n^y,\overline{\psi}_n^z)$ & A test function of $n$th order\\\addlinespace
  $Q$ of $f$ & quantity of interest of $f$: $\int_{\Omega}f{d\bf{x}}$   \\ 
   \addlinespace
 $T$ & The final time \\ 
 \addlinespace
 $\Delta t$ & A time step \\ 
\addlinespace
 $N$ & The number of basis functions  \\ 
  $S$ & The number of samples  \\ 
 \addlinespace
  $\mathcal{G}_{\widetilde{u}}^\theta$ & A representation of the model to infer $\widetilde{u}$ \\
  \addlinespace
  $\mathcal{G}_{\Phi}^\theta$ & A representation of the model to infer $\Phi$  \\
    \addlinespace
  $|\mathbf{u}|$ & Magnitude of $\mathbf{u}$: $\sqrt{u^2+v^2+w^2}$\\
   \addlinespace
  Rel.$L_x^2$ error of $u$ & $\sqrt{\int_{\Omega} |\widehat{u}-u|^2{d\bf{x}}/\int_{\Omega} u^2{d\bf{x}}}$  \\  \addlinespace
  Rel.$L_{t,x}^2$ error of $u$ & $\sqrt{\int_{0}^T\int_{\Omega} |\widehat{u}-u|^2{d\bf{x}}dt/\int_{0}^{T}\int_{\Omega} u^2{d\bf{x}}dt}$  \\  \addlinespace
  Rel.$H_x^1$ error of $p$ & $\sqrt{\int_{\Omega} |\nabla\widehat{p}-\nabla p|^2{d\bf{x}}/\int_{\Omega} |\nabla p|^2{d\bf{x}}}$  \\  \addlinespace
  Rel.$H_{t,x}^1$ error of $p$ & $\sqrt{\int_0^T\int_{\Omega} |\nabla\widehat{p}-\nabla p|^2{d\bf{x}}dt/\int_0^T\int_{\Omega} |\nabla p|^2{d\bf{x}}dt}$  \\ \addlinespace
  $\|\mathbf{x}\|_{l^2}$ &  $\sqrt{\sum_{i=1}^Nx_i^2}$ where $\mathbf{x}=(x_1,x_2,\cdots, x_N)$\\
 \hline
\end{tabular}
\end{table}
\newpage
\section{Training procedure}
\begin{algorithm}[h!]
\caption{Training procedure of the SpecONet}\label{alg:cap}
\begin{algorithmic}
\State Refer to the notations of hyper-parameters in Appendix \ref{notation}.
\State {\bfseries Input:} 
Sample $f(x) \sim$ GRF.
\State {\bfseries Step 1:} Map $f$ to a set of coefficients by $\mathcal{G}_{\widetilde{u}}^\theta$ as in
\begin{align}
\mathcal{G}_{\widetilde{u}}^\theta:f\mapsto \Big[\{\widehat{\pmb{\alpha}}_{lmn}^{k+1}\}_{l,m,n=0}^{N-1}\Big]_{k=0}^{K-1},
\end{align}
where $k$ is a time point.
\State {\bfseries Step 2:} Reconstruct inferences to \eqref{NSE_first} as
\begin{align}    \widehat{\widetilde{\mathbf{u}}}_N^{k+1}=\sum_{l,m.n=0}^{N-1}\widehat{\pmb{\alpha}}_{lmn}^{k+1}\Psi_{lmn}.
\end{align}
\For{$k=0$ to $K-1$} 
 \State {\bfseries Step 3:} Minimize the loss \eqref{loss_legendre} of $k+1$ subsequently for $\widehat{\widetilde{\mathbf{u}}}_N^{k+1}$ to get closer to $\widetilde{\mathbf{u}}_N^{k+1}$. 
   
  \State {\bfseries Step 4:} Compute $\nabla\cdot\widehat{\widetilde{\mathbf{u}}}_N^{k+1} $.   
  
\State {\bfseries Step 5:} Map $\nabla\cdot\widehat{\widetilde{\mathbf{u}}}_N^{k+1}$ to a set of coefficients by $\mathcal{G}^{k+1}_{\Phi}$ as in
\begin{align}  \mathcal{G}_{\Phi}^{k+1,\theta}:\nabla\cdot\widehat{\widetilde{\bf{u}}}^{k+1}_N\mapsto \{\widehat{\phi}_{lmn}\}_{l,m,n=0}^{N-1}.  
\end{align}

\State {\bfseries Step 6:} Reconstruct inferences to \eqref{NSE_second} as
\begin{align}
   \widehat{\Phi}_N^{k+1}=\sum_{l,m.n=0}^{N-1}\widehat{\phi}_{lmn}\Psi_{lmn}.
\end{align}

 \State {\bfseries Step 7:} Minimize the loss \eqref{loss_second} of $k+1$ for $\widehat{\Phi}_N^{k+1}$ to get closer to ${\Phi}_N^{k+1}$. 
 \State {\bfseries Step 8:} Reconstruct inferences to NSEs as
\begin{align}\label{inference_first}
\widehat{\bold{u}}^{k+1}&=\widehat{\widetilde{\bold{u}}}^{k+1}+\frac{2\Delta t}{3}\nabla\widehat{\Phi}^{k+1},\\
\widehat{p}^{k+1}&=\widehat{p}^{k}+\widehat{\Phi}^{k+1}-\nu\nabla\cdot\widehat{\widetilde{\bold{u}}}^{k+1}.
\end{align}
\EndFor
\end{algorithmic}
\end{algorithm} 
\newpage

\section{Network architecture and hyper-parameter settings}\label{sec:hyper-parameter}
SpecONet composes two networks, $\mathcal{G}^\theta_{\widetilde{u}}$, $\mathcal{G}^\theta_{\Phi}$ (see Fig.~\ref{f:schematic}). $\mathcal{G}^\theta_{\widetilde{u}}$ is the network to predict $\widetilde{\mathbf{u}}$, which consists of a single convolutional neural network (CNN) equipped with Swish activation function, and $K$ distinct fully connected neural networks (FNNs). In addition, $\mathcal{G}^{k,\theta}_{\Phi}$ is the network to predict $\widetilde{\Phi}$, which consists of a single CNN equipped with Swish activation function, and a FNNs for each $k=1,\cdots, K$. Regrading an optimizer,  Limited-memory BFGS (L-BFGS) was employed. The hyper-parameters used in each examples are articulated in the table.
\begin{table}[h!]
\begin{tabular}{c|c c   c c c c c c } 
 \hline
type of input & BC &Basis&$T$&$\Delta t$&$N$&Filters&Kernels \\ 
 \hline\vspace{2mm}
  2D forcing function\ref{sub:2d_force} &Dirichlet&Legendre type&1&0.01&22&10&9 \\ 
\vspace{2mm}
 2D initial condition\ref{sub:2d_initial} &periodic&Fourier &1&0.01&32&3&9\\ 
\vspace{2mm}
2D boundary condition\ref{sub:2d_boundary}&Dirichlet&Legendre type&1&0.01&62&30&15 \\
\vspace{2mm}
 3D initial condition(Beltrami flow)\ref{sub:3d_initial} &periodic&Fourier&1&0.01&24& 2 &19  \\ 
\vspace{2mm}
  3D forcing function\ref{sub:3d_force} &Dirichlet&Legendre type&1&0.01&18& 3 &9  \\ 
\hline
\end{tabular}
\caption{The hyperparamethers of $\mathcal{G}_{\widetilde{u}}^\theta$}
\end{table}

\begin{table}[h!]
\begin{tabular}{c|c c   c c c c c c } 
 \hline
type of input & BC &Basis&$T$&$\Delta t$&$N$&Filters&Kernels \\ 
 \hline\vspace{2mm}
  2D forcing function\ref{sub:2d_force} &Neumann&Legendre type&1&0.01&22&10&9 \\ 
\vspace{2mm}
 2D initial condition\ref{sub:2d_initial} &Neumann&Fourier &1&0.01&32&3&9\\ 
\vspace{2mm}
2D boundary condition\ref{sub:2d_boundary}&Neumann&Legendre type&1&0.01&62&3&15 \\
\vspace{2mm}
 3D initial condition(Beltrami flow)\ref{sub:3d_initial} &Neumann&Fourier&1&0.01&24& 2 &19  \\ 
\vspace{2mm}
  3D forcing function\ref{sub:3d_force} &Neumann&Legendre type&1&0.01&18& 3 &9  \\ 
\hline
\end{tabular}
\caption{The hyperparamethers of $ \mathcal{G}_{\Phi}^{k+1,\theta}$}
\end{table}

\begin{table}[h!]
\begin{tabular}{c|c c   c c c c c } 
 \hline
type of input & BC &$T$&$\Delta t$&Nodal points&Kernels&Modes& Depth \\ 
 \hline\vspace{2mm}
  2D forcing function\ref{sub:2d_force} &Dirichlet&1&0.2&$24^2$&32&12&5 \\ 
\vspace{2mm}
 2D initial condition\ref{sub:2d_initial} &periodic&1&0.2&$32^2$&32&12&5 \\ 
\vspace{2mm}
2D boundary condition\ref{sub:2d_boundary}&Dirichlet&1&0.2&$64^2$   &32&12&5 \\ 
\hline
\end{tabular}
\caption{The hyper-parameters of FNO. While training the networks, Swish activation and Adaptive Moment Estimation as optimizer were used.}\label{tab:fno_hyper}
\end{table}

\begin{table}[h!]
\begin{tabular}{c|c c   c c c c c } 
 \hline
type of input & BC &$T$&$\Delta t$&Nodal points&Kernels&Modes& Depth \\ 
 \hline\vspace{2mm}
  2D forcing function\ref{sub:2d_force} &Dirichlet&1&0.2&$24^2$&5&30&3 \\ 
\vspace{2mm}
 2D initial condition\ref{sub:2d_initial} &periodic&1&0.2&$32^2$&5&115&3 \\ 
\vspace{2mm}
2D boundary condition\ref{sub:2d_boundary}&Dirichlet&1&0.2&$64^2$   &5&25&3 \\ 
\hline
\end{tabular}
\caption{The hyper-parameters of POD-DON. While training the networks, Swish activation were used. In addition,  Adaptive Moment Estimation and L-BFGS were employed as optimizer. }\label{tab:DON_hyper}
\end{table}

\section{Methodology}\label{sec:methodology}
In this section, we articulate how to apply SpecONet to four cases: \ref{2d_force}. 2D NSE with Dirichlet boundary condition, \ref{2d_initial}. 2D NSE with periodic boundary condition, \ref{3d_initial}. 3D NSE with periodic boundary condition, and \ref{3d_force}. 3D NSE with Dirichlet boundary condition.    

\subsection{Two dimensional NSE with Dirichlet boundary condition}\label{2d_force}We start by presenting how SpecONet works for the case of Dirichlet boundary condition with input functions. SpecONet has two distinct networks; first one denoted by $\mathcal{G}_{\widetilde{u}}^\theta$ is for solving \eqref{NSE_first}, and the second denoted by $\mathcal{G}^{k+1}_{\Phi}$ is for \eqref{NSE_second}. Once a random input function is given to $\mathcal{G}_{\widetilde{u}}^\theta$, the neural network $\mathcal{G}_{\widetilde{u}}^\theta$  computes a set, $\widehat{\pmb{\alpha}}_{lm}^{k+1}:=\{\widehat{\alpha}^{k+1}_{lm},\widehat{\beta}^{k+1}_{lm}\}_{l,m=0}^{N-1}\subset \mathbb{R}$ for $k=1,2,\cdots$. Thereafter, the inferences are reconstructed as $\widehat{\widetilde{{\bf{u}}}}^{k+1}_N(x,y)=(\widehat{\widetilde{u}}^{k+1}_N,\widehat{\widetilde{v}}^{k+1}_N)$ where
\begin{align}\label{inference2d1}
\widehat{\widetilde{{\bf{u}}}}^{k+1}_N(x,y)&=\sum_{l,m=0}^{N-1}\widehat{\pmb{\alpha}}_{lm}\Psi_{lm}(x,y).
\end{align}
Note that the basis functions, $\Psi_{lm}$ are prepared as in  \eqref{Legendre_poly} to satisfy Dirichlet boundary condition. After that, \eqref{inference2d1} are used to compute the loss functions as follows:
\begin{align}\label{loss_legendre2d}
  \begin{split}
       loss=\sum_{l,m=0}^{N-1}&\left|\int_{\Omega}\frac{1}{2\Delta t}(3\widehat{\widetilde{\bold{u}}}^{k+1}_N-4\widehat{\bold{u}}^{k}+\widehat{\bold{u}}^{k-1})\overline{\Psi}_{lm}\right.\\&+\left.\nu\nabla\widehat{\widetilde{\bold{u}}}^{k+1}_N\cdot\nabla\overline{\Psi}_{lm}+\nabla \widehat{p}^{k}\overline{\Psi}_{lm}-\bold{g}^{k+1}\overline{\Psi}_{lm} d\bf{x}\right|^2,
 \end{split}
\end{align}
where 
\begin{align}\label{loss_force2d}
\bold{g}^{k+1}=\bold{f}(t^{k+1})-(2(\widehat{\bold{u}}^{k}\cdot\nabla) \widehat{\bold{u}}^{k}-(\widehat{\bold{u}}^{k-1}\cdot\nabla) \widehat{\bold{u}}^{k-1}).
\end{align}
In addition, the test functions, $\overline{\Psi}_{lm}$ in  \eqref{loss_legendre2d} are identically defined to the basis functions. When the loss plateaus as the number of epoches grows, the inference $\widehat{\widetilde{{\bf{u}}}}^{k+1}_{N}$ are considered close enough to the reference solutions to \eqref{NSE_first}. Subsequently, we design $\mathcal{G}^{k+1}_{\Phi}$ to take $\nabla\cdot \widehat{\widetilde{{\bf{u}}}}^{k+1}_{N}$ as inputs, and to yield    $\{\widehat{\phi}^{k+1}_{lm}\}_{l,m=0}^{N-1}\subset \mathbb{R}$. Then, the inference is calculated as 
\begin{align}\label{inference2d2}
 \widehat{\Phi}^{k+1}_N(
x,y)&=\sum_{l,m=0}^{N-1}\phi_{lm}\Psi_{lm},
\end{align}
where $\Psi_{lm}$ are equipped with Neumann boundary condition as in \eqref{Legendre_poly}.  
Afterwards, the loss \eqref{loss_second2d} with \eqref{inference2d2} is defined as 
\begin{align}\label{loss_second2d}
   loss=\sum_{l,m=0}^{N-1} \left|\int_{\Omega}\nabla\widehat{\Phi}^{k+1}_N\cdot\nabla\overline{\Psi}_{lm}+ \frac{3}{2\Delta t}\nabla\cdot\widehat{\widetilde{\bold{u}}}^{k}_N\overline{\Psi}_{lm}d\bf{x}\right|^2,
\end{align}
where the test functions, $\overline{\Psi}_{lm}$ are identical to the basis functions in \eqref{inference2d2} in the definition. 
Until the loss \eqref{loss_second} gets static, training $\mathcal{G}^{k+1}_{\Phi}$ proceeds. Upon completion of training, the inference of NSEs for the next time step can be reconstructed by calculating as 
\begin{align}
\widehat{\bold{u}}^{k+1}&=\widehat{\widetilde{\bold{u}}}^{k+1}_N+\frac{2\Delta t}{3}\nabla\widehat{\Phi}^{k+1}_N\\
\widehat{p}^{k+1}&=\widehat{p}^{k}+\widehat{\Phi}^{k+1}_N-\nu\nabla\cdot\widehat{\widetilde{\bold{u}}}^{k+1}_N.
\end{align}

Now, we explain how to compute the loss. For a convenience, we simplify all losses \eqref{loss_legendre2d}, and \eqref{loss_second2d} into
\begin{align}\label{loss_helm}
 \begin{split}
       loss=\sum_{l,m=0}^{N-1}&\left|\int_{\Omega}\tau u\overline{\Psi}_{lm}+\nu\nabla u\cdot\nabla\overline{\Psi}_{lm}-{f}^{k+1}\overline{\Psi}_{lm} d\bf{x}\right|^2,
 \end{split}
\end{align}
where $\tau$, $\nu$ is a positive number, $u=\sum_{l,m=0}^{N-1}\alpha_{lm}\Psi_{lm}$ and $\overline{\Psi}_{lm}=\Psi_{lm}$. Thanks to the Gauss-Lobatto
quadrature rule for computing the spatial integration of \eqref{loss_helm}, the loss turns into the linear system as
\begin{align}\label{2d_system1}
   loss=\|\tau B{\alpha} B+ \nu S{\alpha} B+\nu B{\alpha}S-F\|_{l^2}^2.
\end{align}
where B is a $N\times N$ mass matrix,
\begin{align}\label{mass}
B_{lm}=\int_{\Omega}\psi_l\psi_m dx=\begin{cases}& c_lc_m\left(\frac{2}{2l+1}+\frac{2}{2l+5}\right)~\text{if $l=m$,}\\&-c_lc_m\frac{2}{2l+1}\qquad\qquad\text{if $l=m\pm2$,}\\&0\qquad\qquad\qquad\qquad\text{otherwise},\end{cases}
\end{align}
S is a $N\times N$ stiffness matrix,
\begin{align}
S_{lm}=\int_{\Omega}\nabla\psi_l\cdot\nabla\psi_m dx=\begin{cases}&1\quad\text{if $l=m$,}\\
&0\quad\text{otherwise,}\end{cases}
\end{align}
 F is a $N\times N$ matrix,
\begin{align}
F_{lm}=\int_{\Omega}f^{k+1}\psi_l(x)\psi_m(y)dxdy,
\end{align}
and $\alpha$ is the $N\times N$  unknown matrix whose elements reconstruct $u$. Due to $S$ is the identity matrix, \eqref{2d_system1} is identical to
\begin{align}\label{2d_system12}
   loss=\|\tau B{\alpha}B+ \nu{\alpha} B+\nu B{\alpha}-F\|_{l^2}^2.
\end{align}
In fact, \eqref{2d_system12} is a kind of Sylvester equation whose condition number is relatively high. It implies that iterative methods are not efficient to solve  \eqref{2d_system12} for $\alpha$. Thus, using
diagonalization method, we transform the loss into a form of $AX=B$ to reduce the condition number as follows. Let let $\Lambda$ and $E$ be the matrix whose diagonal entries are eigen values $\lambda$ of $B$ and whose columns are orthonomal eigen vectors of $B$, respectively. And let $V:=E^T\alpha$. Then \eqref{2d_system12} becomes
\begin{align}\label{2d_system2}
    \|\tau E\Lambda V B+\nu EVB+\nu E\Lambda V B-F\|_{l^2}^2.
\end{align}
After that, multiplying $E^T$ by \eqref{2d_system2}, we have
\begin{align}\label{2d_system3}
    \|\tau\Lambda V B+\nu VB+\nu \Lambda V B-G\|^2_{l^2},
\end{align}
where $G:=E^TF$.
Subsequently, transpose the \eqref{2d_system3}, it reads
\begin{align}\label{2d_system4}
    \|\tau BV^T\Lambda+\nu BV^T+\nu BV^T\Lambda- G^T\|^2_{l^2}
\end{align}
Then \eqref{2d_system4} is converted to the form $AX=B$ as
\begin{align}\label{2d_system5}
   \|((\tau\lambda_p+\nu)B+\nu\lambda_pI) v_p-g_p\|^2_{l^2},
\end{align}
where the subscript $p$ means the $p$-th eigen value of $\Lambda$ and the column vector of $V$ and $G$. In addition, for reducing the condition number more, we apply a precondition matrix $C$ 
\begin{align}\label{precond}
C_{lm}=\begin{cases}1/\sqrt{((\tau\lambda_p+\nu)B_{ii}+\nu\lambda_p}\quad\text{if}~i=j,\\0\qquad\qquad\qquad\qquad\qquad \text{otherwise}.\end{cases}
\end{align}
If the precondition matrix is multiplied the both sides of $((\tau\lambda_p+\nu)B+\nu\lambda_pI)$, the diagonal entries of the multiplication are all one, which makes the condition number smaller. The way it works is as follows. Let $C^{-1}v_p=w_p$. Then, we get the loss by multiplying $C$ by \eqref{2d_system5} leading to  
\begin{align}\label{2d_system6}
   loss=\|C((\tau\lambda_p+\nu)B+\nu\lambda_pI)C w_p-h_p\|^2_{l^2},
\end{align}
where $h_p:=Cg_p$, which is actually employed to train the networks.
After finding $w_p$ by minimizing the loss \eqref{2d_system6}, $\alpha_p$ can be restored as
 \begin{align}
    \alpha_p=ECw_p.
\end{align}
Finally, the unknown matrix $\alpha$ is obtained by composing the column vectors, $\alpha_p$.

\subsection{Two dimensional NSE with periodic boundary conditions}\label{2d_initial}
Now, we deal with how SpecONet is designed if periodic boundary conditions are imposed. For this experiment, let a random input datum $(u_0,v_0)$ prepared. As the same manner as in section \ref{2d_force}, when inserting the input datum, the first network of SpecONet produces $\widehat{\pmb{\alpha}}^{k+1}_{\xi_l\xi_m}:=\{\widehat{\alpha}^{k+1}_{\xi_l\xi_m},\widehat{\beta}^{k}_{\xi_l\xi_m}\}_{l,m=0}^{N-1}\subset \mathbb{C}$ for $(\xi_l,\xi_m)=-N/2+1+(l,m)$ and $k=0,1,\cdots$. Subsequently, the inferences for \eqref{NSE_first} are composed as
\begin{align}\label{inference_fourier}
\widehat{\widetilde{\bold{u}}}^{k+1}_N(x,y)=\left(\frac{1}{2\pi}\right)^2\sum_ {l,m=0}^{ N-1}\widehat{\pmb{\alpha}}_{\xi_l\xi_m}^{k+1}e^{\mathrm{i}(\xi_lx+\xi_my)},
\end{align}
where $\widehat{\widetilde{\bold{u}}}^{k+1}_N=(\widehat{\widetilde{{u}}}^{k+1}_N,\widehat{\widetilde{{v}}}^{k+1}_N)$.
Then, the loss function should be
\begin{align}\label{loss_period2d}
   \begin{split}
       loss=\sum_{l,m=0}^{N-1}&\left|\int_{\Omega}\frac{1}{2\Delta t}(3\widehat{\widetilde{\bold{u}}}^{k+1}_N-4\widehat{\bold{u}}^{k}+\widehat{\bold{u}}^{k-1})\overline{\Psi}_{lm}\right.\\
&\left.+\nu\nabla\widehat{\widetilde{\bold{u}}}^{k+1}_N\cdot\nabla\overline{\Psi}_{lm}+\nabla \widehat{p}^{k}\overline{\Psi}_{lm}-\bold{g}^{k+1}\overline{\Psi}_{lm}d\bf{x}\right|^2,
 \end{split}
\end{align}
where 
\begin{align}\label{loss_force_period}
\bold{g}^{k+1}=\bold{f}(t^{k+1})-(2(\widehat{\bold{u}}^{k}\cdot\nabla) \widehat{\bold{u}}^{k}-(\widehat{\bold{u}}^{k-1}\cdot\nabla) \widehat{\bold{u}}^{k-1}),
\end{align}
and the test functions are $\overline{\Psi}_{lm}:=e^{-\mathrm{i}(\xi_lx+\xi_my)}$. When the loss gets static as training proceeds, the inference $\widehat{\widetilde{{\bf{u}}}}^{k+1}_{N}$ are expected to be close to the reference solutions to \eqref{NSE_first}. Afterwards, $\mathcal{G}^{k+1}_{\Phi}$ maps $\nabla\cdot \widehat{\widetilde{{\bf{u}}}}^{k+1}_{N}$ to $\{\widehat{\phi}^{k+1}_{lm}\}_{l,m=0}^{N-1}\subset \mathbb{C}$. Then, the inference is calculated as 
\begin{align}\label{inference2d2period}
 \widehat{\Phi}^{k+1}_N(
x,y)&=\left(\frac{1}{2\pi}\right)^2\sum_{l,m=0}^{N-1}\phi_{lm}e^{\mathrm{i}(\xi_lx+\xi_my)}.
\end{align}
Following that, the loss is defined as 
\begin{align}\label{loss_second2d_period}
   loss_{k+1}=\sum_{l,m=0}^{N-1} \left|\int_{\Omega}\nabla\widehat{\Phi}^{k+1}_N\cdot\nabla\overline{\Psi}_{lm}+ \frac{3}{2\Delta t}\nabla\cdot\widehat{\widetilde{\bold{u}}}^{k}_N\overline{\Psi}_{lm}d\bf{x}\right|^2,
\end{align}
where
the test functions are $\overline{\Psi}_{lm}:=e^{-\mathrm{i}(\xi_lx+\xi_my)}$. Consequently, the inferences for the next time step read 
\begin{align}
\begin{split}
\widehat{\bold{u}}^{k+1}&=\widehat{\widetilde{\bold{u}}}^{k+1}_N+\frac{2\Delta t}{3}\nabla\widehat{\Phi}^{k+1}_N\\
\widehat{p}^{k+1}&=\widehat{p}^{k}+\widehat{\Phi}^{k+1}_N-\nu\nabla\cdot\widehat{\widetilde{\bold{u}}}^{k+1}_N.
\end{split}
\end{align}

Now, based on linear algebra, the losses are computed as follows. Let \eqref{loss_period2d} and \eqref{loss_second2d_period} simplified as
\begin{align}\label{loss_helm_period}
 \begin{split}
       loss=\sum_{l,m=0}^{N-1}&\left|\int_{\Omega}\tau u\overline{\Psi}_{lm}+\nu\nabla u\cdot\nabla\overline{\Psi}_{lm}-{f}\overline{\Psi}_{lm} d\bf{x}\right|^2,
 \end{split}
\end{align}
where $\tau$, $\nu$ is a positive number, $u:=\left(\frac{1}{2\pi}\right)^2\sum_{l,m=0}^{N-1}\alpha_{lm}e^{\mathrm{i}(\xi_lx+\xi_my)}$, and $\overline{\Psi}_{lm}:=e^{-\mathrm{i}(\xi_lx+\xi_my)}$. Then, substituting all to the loss, \eqref{loss_helm_period} turns into
\begin{align}\label{loss_form}
loss=\sum_ {l,m=0}^{ N-1}\|\tau{\alpha}_{\xi_l\xi_m}+ \nu(\xi_l^2+\xi_m^2){\alpha}_{\xi_l\xi_m}-\mathcal{F}_{\xi_l\xi_m} (f)\|^2_{l^2}.
\end{align}
Here, $\mathcal{F}_{\xi_l\xi_m}$ for $(\xi_l,\xi_m)=-N/2+1+(l,m)$ is defined by
\begin{align}
 \mathcal{F}_{\xi_l\xi_m} (f)=h^2\sum_{l,m=0}^{N-1}f(x_l,y_m)e^{-\mathrm{i}(\xi_lx_l+\xi_my_m)},
\end{align}
where $x_l=lh$, $y_m=mh$ on $[0, 2\pi)$, and $h=\frac{2\pi}{N}$.

\subsection{Three dimensional NSE with periodic boundary conditions}\label{3d_initial}
Now, we design SpecONet to infer 3D NSE solutions provided that periodic boundary conditions are imposed. Let a random input datum given. As the same manner as in section \ref{2d_force}, when the input datum goes through the first network of SpecONet to generate $\widehat{\pmb{\alpha}}^{k+1}_{\xi_l\xi_m\xi_n}:=\{\widehat{\alpha}^{k+1}_{\xi_l\xi_m\xi_n},\widehat{\beta}^{k+1}_{\xi_l\xi_m},\widehat{\gamma}^{k+1}_{\xi_l\xi_m\xi_n}\}_{l,m=0}^{N-1}\subset \mathbb{C}$ for $(\xi_l,\xi_m,\xi_n)=-N/2+1+(l,m,n)$ and $k=0,1,\cdots$. After that, the inferences for \eqref{NSE_first} are made as
\begin{align}\label{inference_fourier3d}
\widehat{\widetilde{\pmb{u}}}^{k+1}(x,y,z)&=\left(\frac{1}{2\pi}\right)^3\sum_ {l,m,n=0}^{ N-1}\widehat{\pmb{\alpha}}_{\xi_l\xi_m\xi_n}^{k+1}e^{\mathrm{i}(\xi_lx+\xi_my+\xi_nz)}.
\end{align}
Then, the loss function should be
\begin{align}\label{loss_period3d}
  \begin{split}
       loss=\sum_{l,m.n=0}^{N-1}&\left|\int_{\Omega}\frac{1}{2\Delta t}(3\widehat{\widetilde{\bold{u}}}^{k+1}_N-4\widehat{\bold{u}}^{k}+\widehat{\bold{u}}^{k-1})\overline{\Psi}_{lmn}\right.\\
&\left.+\nu\nabla\widehat{\widetilde{\bold{u}}}^{k+1}\cdot\nabla\overline{\Psi}_{lmn}+\nabla \widehat{p}^{k}\overline{\Psi}_{lmn}-\bold{g}^{k+1}\overline{\Psi}_{lmn} d\bf{x}\right|^2,
 \end{split}
\end{align}
where 
\begin{align}\label{loss_force_period3d}
\bold{g}^{k+1}=\bold{f}(t^{k+1})-(2(\widehat{\bold{u}}^{k}\cdot\nabla) \widehat{\bold{u}}^{k}-(\widehat{\bold{u}}^{k-1}\cdot\nabla) \widehat{\bold{u}}^{k-1}),
\end{align}
and the test functions are $\overline{\Psi}_{lmn}:=e^{-\mathrm{i}(\xi_lx+\xi_my+\xi_nz)}$. When the loss get flat as training continues, the inference $\widehat{\widetilde{{\bf{u}}}}^{k+1}_{N}$ are regarded to be close to the reference solutions to \eqref{NSE_first}. Thereafter, $\mathcal{G}^{k+1}_{\Phi}$ links $\nabla\cdot \widehat{\widetilde{{\bf{u}}}}^{k+1}_{N}$ to $\{\widehat{\phi}^{k+1}_{lmn}\}_{l,m,n=0}^{N-1}\subset \mathbb{C}$. Then, the inference is calculated as 
\begin{align}\label{inference3d2period}
 \widehat{\Phi}^{k+1}_N(
x,y)&=\left(\frac{1}{2\pi}\right)^3\sum_{l,m,n=0}^{N-1}\phi_{lmn}e^{\mathrm{i}(\xi_lx+\xi_my+\xi_nz)}.
\end{align}
Following that, the loss is defined as 
\begin{align}\label{loss_second3d_period}
   loss_{k+1}=\sum_{l,m,n=0}^{N-1} \left|\int_{\Omega}\nabla\widehat{\Phi}^{k+1}_N\cdot\nabla\overline{\Psi}_{lmn}+ \frac{3}{2\Delta t}\nabla\cdot\widehat{\widetilde{\bold{u}}}^{k}_N\overline{\Psi}_{lmn}d\bf{x}\right|^2,
\end{align}
where
the test functions are $\overline{\Psi}_{lmn}:=e^{-\mathrm{i}(\xi_lx+\xi_my+\xi_nz)}$.
Therefore, the inferences for the next time step are computed by
\begin{align}
\begin{split}
\widehat{\bold{u}}^{k+1}&=\widehat{\widetilde{\bold{u}}}^{k+1}_N+\frac{2\Delta t}{3}\nabla\widehat{\Phi}^{k+1}_N\\
\widehat{p}^{k+1}&=\widehat{p}^{k}+\widehat{\Phi}^{k+1}_N-\nu\nabla\cdot\widehat{\widetilde{\bold{u}}}^{k+1}_N.
\end{split}
\end{align}

Now, the losses are computed as follows. Let \eqref{loss_period3d} and \eqref{loss_second3d_period} simplified as
\begin{align}\label{loss_helm_period3d}
 \begin{split}
       loss=\sum_{i,j,k=0}^{N-1}&\left|\int_{\Omega}\tau u\Psi_{ijk}+\nu\nabla u\cdot\nabla\Psi_{ijk}-{f}\Psi_{ijk} d\bf{x}\right|^2,
 \end{split}
\end{align}
where $\tau$, $\nu$ is a positive number, $u:=\sum_{i,j,k=0}^{N-1}\alpha_{ijk}e^{\mathrm{i}(\xi_lx+\xi_my+\xi_nz)}$, and $\overline{\Psi}_{lmn}:=e^{-\mathrm{i}(\xi_lx+\xi_my+\xi_nz)}$. Then, substituting all to the loss, \eqref{loss_helm_period3d} becomes
\begin{align}\label{loss_form3d}
loss=\sum_ {l,m,n=0}^{ N-1}\|\tau{\alpha}_{\xi_l\xi_m\xi_n}+ \nu(\xi_l^2+\xi_m^2+\xi_n^2){\alpha}_{\xi_l\xi_m\xi_n}-\mathcal{F}_{\xi_l\xi_m\xi_n} (f)\|^2_{l^2}.
\end{align}
Here, $\mathcal{F}_{\xi_l\xi_m\xi_n}$ for $(\xi_l,\xi_m,\xi_n)=-N/2+1+(l,m,n)$ is defined by
\begin{align}
 \mathcal{F}_{\xi_l\xi_m\xi_n} (f)=h^3\sum_{l,m,n=0}^{N-1}f(x_l,y_m,z_n)e^{-\mathrm{i}(\xi_lx_l+\xi_my_m+\xi_nz_n)},
\end{align}
where $x_l=lh$, $y_m=mh$, $z_n=nh$ on $[0, 2\pi)$, and $h=\frac{2\pi}{N}$.

\subsection{Three dimensional NSE with Dirichlet boundary condition}\label{3d_force}
Finally, we display SpecONet that infers 3D NSE solutions imposing the homogeneous Dirichlet boundary condition. Likewise, two distinct networks should be constructed. The first network, $\mathcal{G}_{\widetilde{u}}^\theta$ takes an input datum, it yields a set of coefficients $\widehat{\pmb{\alpha}}^{k+1}_{lmn}:=\{\widehat{\alpha}^{k+1}_{lmn},\widehat{\beta}^{k+1}_{lmn},\widehat{\gamma}^{k+1}_{lmn}\}_{l,m,n=0}^{N-1}$ for $k=0,1,\cdots$. 
Consequently, the inference to \eqref{NSE_first} is constructed as $\widehat{\widetilde{{\bf{u}}}}_N^{k+1}:=(\widehat{\widetilde{u}}^{k+1}_N,\widehat{\widetilde{v}}^{k+1}_N,\widehat{\widetilde{w}}^{k+1}_N)$ where
\begin{align}\label{inference3d1}
\widehat{\widetilde{{\bf{u}}}}_N^{k+1}=\sum_{l,m,n=0}^{N-1}\widehat{\pmb{\alpha}}_{lmn}^{k+1}\Psi_{lmn}.
\end{align}
Note that $\Psi_{lmn}$ are prepared basis functions as in \eqref{Legendre_poly} to impose Dirichlet boundary condition. Subsequently, we define the loss as
\begin{align}\label{loss_helm3d}
  \begin{split}
       loss=\sum_{l,m.n=0}^{N-1}&\left|\int_{\Omega}\frac{1}{2\Delta t}(3\widehat{\widetilde{\bold{u}}}^{k+1}_N-4\widehat{\bold{u}}^{k}_N+\widehat{\bold{u}}^{k-1}_N)\overline{\Psi}_{lmn}\right.\\&\left.+\nu\nabla\widehat{\widetilde{\bold{u}}}^{k+1}\cdot\nabla\overline{\Psi}_{lmn}+\nabla \widehat{p}^{k}\overline{\Psi}_{lmn}-\bold{g}^{k+1}\overline{\Psi}_{lmn} d\bf{x}\right|^2,
 \end{split}
\end{align}
where 
\begin{align}
\bold{g}(t^{k+1})=\bold{f}(t^{k+1})-(2(\widehat{{{\bf{u}}}}^{k}\cdot\nabla) \widehat{{{\bf{u}}}}^{k}-(\widehat{{{\bf{u}}}}^{k-1}\cdot\nabla) \widehat{{{\bf{u}}}}^{k-1}).
\end{align}
Here, the test functions $\overline{\Psi}_{lmn}$ are defined identically to the basis functions in \eqref{inference3d1}.  
When the loss stops moving even though training lasts, the inference $\widehat{\widetilde{{\bf{u}}}}^{k+1}_{N}$ are considered close enough to the reference solutions to \eqref{NSE_first}. Subsequently, we set up $\mathcal{G}^{k+1}_{\Phi}$ to connect $\nabla\cdot \widehat{\widetilde{{\bf{u}}}}^{k+1}_{N}$ to $\{\widehat{\phi}^{k+1}_{lm}\}_{l,m=0}^{N-1}\subset \mathbb{R}$. Then, the inference is calculated as 
\begin{align}\label{inference3d2}
 \widehat{\Phi}^{k+1}_N(
x,y)&=\sum_{l,m,n=0}^{N-1}\phi_{lmn}\Psi_{lmn},
\end{align}
where $\Psi_{lmn}$ satisfy Neumann boundary condition. 
Afterwards, the loss with \eqref{inference3d2} is defined as 
\begin{align}\label{loss_second3d}
   loss=\sum_{l,m,n=0}^{N-1} \left|\int_{\Omega}\nabla\widehat{\Phi}^{k+1}_N\cdot\nabla\overline{\Psi}_{lmn}+ \frac{3}{2\Delta t}\nabla\cdot\widehat{\widetilde{\bold{u}}}^{k+1}_N\overline{\Psi}_{lmn}d\bf{x}\right|^2,
\end{align}
where $\overline{\Psi}_{lmn}$ are identically generated to the basis functions used in \eqref{inference3d2}. 
If the loss \eqref{loss_second3d} decreases, training $\mathcal{G}^{k+1}_{\Phi}$ continues. After end of training, the inference of NSEs for the next time step can be reconstructed by calculating as 
\begin{align}
\begin{split}
\widehat{\bold{u}}^{k+1}&=\widehat{\widetilde{\bold{u}}}^{k+1}_N+\frac{2\Delta t}{3}\nabla\widehat{\Phi}^{k+1}_N\\
\widehat{p}^{k+1}&=\widehat{p}^{k}+\widehat{\Phi}^{k+1}_N-\nu\nabla\cdot\widehat{\widetilde{\bold{u}}}^{k+1}_N.
\end{split}
\end{align}

Now, we specify how to compute the loss. For a convenience, we simply put all losses \eqref{loss_helm3d}, and \eqref{loss_second3d} into
\begin{align}\label{loss_helm}
 \begin{split}
       loss=\sum_{l,m,n=0}^{N-1}&\left|\int_{\Omega}\tau u\overline{\Psi}_{lmn}+\nu\nabla u\cdot\nabla\overline{\Psi}_{lmn}-{f}\overline{\Psi}_{lmn} d\bf{x}\right|^2,
 \end{split}
\end{align}
where $\tau$, $\nu$ is a positive number and $u=\sum_{l,m,n=0}^{N-1}\alpha_{lmn}\Psi_{lmn}$. Then, due to the Gauss-Lobatto quadrature rule, the integration in \eqref{loss_helm3d} is equivalent to the following linear system 
\begin{align}\label{3d_system1}
   \tau B_{pl}\alpha_{lmn}B_{qm}B_{rn}+\nu\alpha_{pmn}B_{qm}B_{rn}+\nu B_{pl}\alpha_{lqn}B_{rn}+\nu B_{pl}\alpha_{lmr}B_{qm}-f_{pqr}
\end{align}
where $B_{pl}$ is the $pl$th element of the mass matrix \eqref{mass}, and  $f_{pqr}=\int_{\Omega}f\Psi_{pqr}d\bf{x}$. Note that the multiplication in \eqref{3d_system1}, and in subsequent equations follow the Einstein summation:
\begin{align*}
    B_{ni}B_{nj}:=\sum_{n=0}^{N-1}B_{ni}B_{nj}.
\end{align*}
Subsequently, note that the definition of the eigenvalues, $\lambda$, and the orthonormal matrix $E$ of $B$ reads  
\begin{align}\label{3d_system2}
    B_{pl}E_{li}=\lambda_{q}E_{pi},~E_{pi}E_{pj}=\delta_{ij}.
\end{align}
Let $v_{imn}:=\alpha_{lmn}/E_{li}$. Then, the above definition leads \eqref{3d_system1} to   
\begin{align}\label{3d_system3}
   \tau \lambda_{i} E_{pi}v_{imn}B_{qm}B_{rn}+\nu E_{li}v_{imn}B_{qm}B_{rn}+\nu\lambda_{i} E_{pi}v_{iqn}B_{rn}+\nu\lambda_{i} E_{pi}v_{lmr}B_{qm}-f_{pqr}.
\end{align}
Applying the Einstein summation to \eqref{3d_system3} with $E_{pj}$, it turns into
\begin{align}\label{3d_system4}
   (\tau \lambda_{j}+\nu )v_{jmn}b_{qm}b_{rn}+\nu\lambda_{j}( v_{jqn}b_{rn}+ v_{jmr}b_{qm})=e_{pj}f_{pqr}:=g_{jqr}.
\end{align}
If $V^j:=v_{jmn}$ and $G^j:=g_{jmn}$, \eqref{3d_system1} is equivalent to
\begin{align}\label{3d_system5}    (\tau\lambda_j+\nu)BV^jB+\nu\lambda_j(V^jB+BV^j)=G^j.
\end{align}
Note that for each $0\leq j\leq N-1$, \eqref{3d_system5} is the same format to \eqref{2d_system1} of the 2d system. Thus, when applying the procedure as in the 2d system, it becomes the same format to \eqref{2d_system5},
\begin{align}\label{3d_system5}
   \|(\tau\lambda_j\lambda_k+\nu\lambda_k+\nu\lambda_j)B+\nu\lambda_j\lambda_kI) v_k^j-g_k^j\|^2_{l^2},
\end{align}
where $v_k^j$ and $g_k^j$ are the $k$-th column of $V^j$ and $G^j$, respectively. 
Moreover, precondition matrix \eqref{precond}, $C^j$ is applied for reducing the condition number, which leads to $(C^j)^{-1}v_{k}^j=w_{k}^j$. 
\begin{align}\label{3d_system6}   loss=\|C^j(\tau\lambda_j\lambda_k+\nu\lambda_j+\nu\lambda_k)B+\nu\lambda_j\lambda_kI)C^j w_p^j-h_k^j\|^2_{l^2},
\end{align}
where $h_k:=C^jg_k$, which is actually employed to train the networks.

After obtaining $w_k$ by solving the linear system, we have the solution, 
\begin{align}
    \alpha_k^j=E^jC^jw_k^j.
\end{align}
In the long run, the unknown matrix, $\alpha$, is composed of the column vectors, $\alpha_k^j$. 
\subsection{Architecture of SPecONet}\label{sec:architecture}
In this section, we explain the architecture of SPecONet which is made of $\mathcal{G}_{\widetilde{u}}^\theta$ \eqref{network_first} and $\mathcal{G}_{\Phi}^\theta$ \eqref{network_second}. 

The architecture of $\mathcal{G}_{\widetilde{u}}^\theta$ is described as follows. $\mathcal{G}_{\widetilde{u}}^\theta$ consists of a single CNN with a Swish activation function and $K$ distinct FNNs (see Fig.~\ref{f:schematic}a). Let $S$ input functions given with a size of
\begin{align}
S\times K\times d\times N_x~&\text{if the input functions are}~b(x),  \\
S\times K\times d\times N_x\times N_y\times N_z~&\text{if the input functions are}~\bf{f}(x),\\
S\times d\times N_x\times N_y\times N_z~&\text{if the input functions are}~\bf{u}_0(x).
\end{align}
Here, $d$ is the dimension of the input functions; $N_x$, $N_y$, $N_z$ are the number of nodal points in the $x$, $y$, $z$ directions, respectively. For convenience, we denote an input function to $\mathbf{f}:=(f_1,f_2,f_3)$. Then, the CNN maps the input functions to $S\cdot K$ intermediate outputs denoted by $\mathcal{C}_{\widetilde{u}}$ as
\begin{align}
\mathcal{C}^{s,k}_{\widetilde{u}}:=\sigma(b+\sum_{i=1}^d\mathcal{K}_i\star f^{s,k}_i),~\text{for all $k=1,2,\cdots, K$,}
\end{align}
where $\mathcal{K}$ and $b$ are the convolving kernel and bias of the CNN, respectively, and $\star$
denotes the multi-dimensional cross-correlation operator.
In addition, the size of $\mathcal{C}_{\widetilde{u}}$ is
 \begin{align}\label{app:size_cnn}
S\times K\times C^{out}\times L^{out}_x\times L^{out}_y\times L^{out}_z,
\end{align}
where the dimension of the input functions, $d$ is regarded as   the number of the input's channels; $C^{out}$ is the number of the output's channels; $L^{out}_x$,$L^{out}_y$,$L^{out}_z$ are the number of the nodal points produced by the convolution in the $x$, $y$, $z$ directions, respectively. Note that the values of $L^{out}_x$,$L^{out}_y$,$L^{out}_z$ depend on kernel size, padding size, stride, and so on. After that, $\mathcal{C}_{\widetilde{u}}$ is transformed by $K$ FNNs as
 \begin{align}\label{app:size_fnn}
\{\pmb{\alpha}^{k}_{lmn}\}_{k=0}^{K-1}= \mathcal{C}_{\widetilde{u}}\mathcal{W}_{\widetilde{u}}
 \end{align}
if the $K$ FNNs' weight is denoted by $\mathcal{W}_{\widetilde{u}}$ with the size of  
\begin{align*}
K\times(C^{out}\cdot L^{out}_x \cdot L^{out}_y\cdot L^{out}_z)\times (d\cdot N^3),
\end{align*}
and $\mathcal{C}_{\widetilde{u}}$ is reshaped from \eqref{app:size_cnn} to 
\begin{align*}
K\times S\times  (C^{out}\cdot L^{out}_x\cdot L^{out}_y\cdot L^{out}_z).
\end{align*}
Accordingly, the size of $\mathcal{C}_{\widetilde{u}}$ and $\mathcal{W}_{\widetilde{u}}$ ensures that 3D tensor multiplication in \eqref{app:size_fnn} is well defined. Thus, the size of $\{\pmb{\alpha}^{k}_{lmn}\}_{k=0}^{K-1}$ should be  
 \begin{align}
K\times S\times (d\cdot N^3),
\end{align}
or it is reshaped to have a size of 
 \begin{align}
S\times K\times d\times N\times N\times N.
\end{align}

The second operator network, $\mathcal{G}_{\Phi}^{k,\theta}$ \eqref{network_second}, is designed as follows. This network is constructed by a CNN and an FNN as follows. Let the size of the input $\nabla\cdot\mathbf{u}^k$ be
\begin{align}
S\times  1\times N_x\times N_y\times N_z,
\end{align}
where $1$ is regarded as the number of the input's channel.
Then, the CNN maps the input functions to $S$ intermediate outputs denoted by $\mathcal{C}_{\Phi}$ as
\begin{align}
\mathcal{C}^{s,k}_{\Phi}:=\sigma(b+\mathcal{K}\star (\nabla\cdot\mathbf{u}^{s,k})),
\end{align}
where $\mathcal{K}$ and $b$ are the convolving kernel and bias of the CNN, respectively, and $\star$
denotes the multi-dimensional cross-correlation operator. Then, the size of $\mathcal{C}_{\Phi}$ is
 \begin{align}\label{app:size_cnn2}
S\times  C^{out}\times L^{out}_x\times L^{out}_y\times L^{out}_z.
\end{align}
Subsequently, $\mathcal{C}_{\Phi}$ is transformed by the FNN as
 \begin{align}\label{app:size_fnn2}
\widehat{\phi}^k_{lmn}= \mathcal{C}_{\Phi}\mathcal{W}_{\Phi}
 \end{align}
if the weight of the FNN is denoted by $\mathcal{W}_{\Phi}$ with the size of  
\begin{align*}
(C^{out}\cdot L^{out}_x \cdot L^{out}_y\cdot L^{out}_z)\times N^3,
\end{align*}
and $\mathcal{C}_{\Phi}$ is reshaped from \eqref{app:size_cnn2} to 
\begin{align*}
 S\times  (C^{out}\cdot L^{out}_x\cdot L^{out}_y\cdot L^{out}_z).
\end{align*}
Accordingly, $\widehat{\phi}^k_{lmn}$ in \eqref{network_second} is found with the size of 
 \begin{align}
 S\times  N^3,
\end{align}
or it is reshaped to have a size of 
 \begin{align}
S\times N\times N\times N.
\end{align}

\newpage
\section{Numerical experiments}
In this section, we design the numerical experiments to demonstrate three features of SpecONet: 1. flexibility on types of input, 2. accomplishment of accuracy without reference data reliance, and 3. robustness for more complex input functions.

To this end, we are going to show and analyse results of five experiments on different types of input data: \ref{sub:2d_force}. 2D forcing functions; \ref{sub:2d_initial}. 2D initial conditions; \ref{sub:2d_boundary}. 2D boundary conditions; \ref{sub:2d_initial}. 3D initial conditions; \ref{sub:3d_force}. 3D forcing functions. 

Moreover, in order to demonstrate that SpecONet can accomplish accuracy without reference data reliance, we will compare accuracy of SpecONet to accuracy of FNO and POD-DON which rely on reference data. In particular, we designed the FNO \cite{lu2021learning} and the POD-DON \cite{lu2022comprehensive} to map an input function $v(x)$ to reference solutions $u(T,x)$ at a specific time T as
\begin{align}
   \mathcal{G}:v(x) \longmapsto u(T,x).
\end{align}
The hyper-parameters of the networks can be refered to table \ref{tab:fno_hyper} and \ref{tab:DON_hyper}.  For FNO+RNN \cite{lu2021learning}, it was trained to map as
\begin{align}
   \mathcal{G}:\{v(T_\ast-9\Delta t,x),v(T_\ast-8\Delta t,x),\cdots,v(T_\ast,x)\} \longmapsto u(T_\ast,x).
\end{align}
for a specific time $T_\ast$.
Once training was complete, the FNO+RNN was used to infer NSE solutions at a time $T$ in time marching sense as   
\begin{align}
   \mathcal{G}:\{v(T-9\Delta t,x),v(T-8\Delta t,x),\cdots,v(T,x)\} \longmapsto u(T,x).
\end{align}

Lastly, we are going to exhibit robustness of SpecONet by conducting experiments on more complex input data.

Before observing specific experiments, table \ref{tab:error_all} and \ref{tab:error_3d} provide an overview of accuracy with respect to the five types of input. In order to evaluate the errors as in \ref{tab:error_all} and \ref{tab:error_3d}, we generated 100 test data for each type, which were not used in the training process. After that, the 100 errors between inferences and reference solutions were measured on spacial-temporal domains in the norms as in \ref{tab:error_all} and \ref{tab:error_3d}. The values in table \ref{tab:error_all} and \ref{tab:error_3d} were the average of the 100 errors.

\begin{table}[h!]
\centering
\begin{tabular}{|c|c|c|c|}
\hline
\makecell{Types of input}& \makecell{Two dimensional \\forcing function}   & \makecell{Two dimensional \\initial condition}  & \makecell{Two dimensional \\boundary condition}  \\
\hline\hline
Rel.$L^2_{t,x}$ error of $u$&6.91e-03 &4.42e-03 &3.03e-03 \\
\hline
Rel.$L^2_{t,x}$ error of $v$&7.18e-03 &4.45e-03 &3.98e-03  \\
\hline
Rel.$H^1_{t,x}$ error of $p$&3.86e-03 &1.09e-01 &4.51e-03  \\
\hline
\end{tabular}
\caption{The Rel.$L^2_{t,x}$ error of 2d NSEs.  }\label{tab:error_all}
\end{table}

\begin{table}[h!]
\centering
\begin{tabular}{|c|c|c|c|c|c|}
\hline
\makecell{Types of input} & \makecell{Three dimensional \\initial condition} & \makecell{Three dimensional \\forcing function} \\
\hline\hline
Rel.$L^2_{t,x}$ error of $u$  &1.26e-04 &2.42e-02  \\
\hline
Rel.$L^2_{t,x}$ error of $v$ &1.26e-04 &2.47e-02  \\
\hline
Rel.$L^2_{t,x}$ error of $w$ &1.26e-04 &2.46e-02  \\
\hline
Rel.$H^1_{t,x}$ error of $p$  &1.17e-01 &1.83e-02  \\
\hline
\end{tabular}
\caption{The Rel.$L^2_{t,x}$ error of 3D NSEs.  }\label{tab:error_3d}
\end{table}

\subsection{Two dimensional NSE with random forcing functions}\label{sub:2d_force}
We conducted experiments to evaluate the accuracy of inference of our method about random forcing functions as input. Random forcing functions, $f_x$ and $f_y$ were generated by the real part of
\begin{align}\label{generation:2d_force}
\frac{1}{12}\sin(t)\sum_{k_x,k_y=0}^{2}c_{k_xk_y}\exp(i(k_xx+k_yy)),
\end{align}
where $c_{k_xk_y}=a_{k_xk_y}+ib_{k_xk_y}$ and $a_{k_xk_y}$ and $b_{k_xk_y}$ are random variables. We generated 600 training samples whose $a_{k_xk_y}$ and $b_{k_xk_y}$ were drawn from $\mathcal{N}(0,5^2)$. 

The other information is as follows:
\begin{table}[h!]
    \centering    
    \begin{tabular}{|c|c||c|c|}
    \hline
    Domain   & $[-1,1]^2$  & Boundary condition   & all zero (Dirichlet condition).   \\ \hline
    Bases type   & Legendre   & Initial condition   &$u_0=v_0=0$   \\ \hline
   $\Delta t$&0.01& The number of time steps& 100\\ \hline
    $\nu$&0.1& N& 22\\ \hline
    \end{tabular}
    \caption{Information on the numerical methods for 2D forcing functions.}
    \label{tab:2d_forcing}
\end{table}

In order for comparison, we carried out training on POD-DON, FNO, and FNO+RNN with the forcing functions and the corresponding reference solutions at $T=0.2,0.4,0.6,0.8$ and $1$. In addition, we set $T_\ast=0.2$ for FNO+RNN. Note that the number of training samples for FNO and POD-DON varied from 10 to 600, which will underscore that accuracy of POD-DON and FNO are sensitive to the number of training samples whereas accuracy of SpecONet does not depend on the number.

 For test samples, we made 100 forcing functions from three cases, $\mathcal{N}(0,5^2)$, $\mathcal{N}(0,10^2)$, and $\mathcal{N}(0,20^2)$. Afterwards, velocity solutions to NSEs at $T=0.2,0.4,0.6,0.8$ and $1$ were computed to employ them as reference solutions. Then, 3 sets of 100 errors were computed in Rel.$L^2_x$ sense between inferences of each methods and the velocity reference solutions. The comparison on the errors made by each networks are discussed in the following subsections: a) $\mathcal{N}(0,5^2)$ (see table \ref{tab:comparison_2dforce_sigma5}, Fig.~ \ref{fig:comparison_2dforce_sigma5}); b) $\mathcal{N}(0,10^2)$ (see table \ref{tab:comparison_2dforce_sigma10}; Fig.~\ref{fig:comparison_2dforce_sigma10}); c) $\mathcal{N}(0,20^2)$ (see table \ref{tab:comparison_2dforce_sigma20}; Fig.~\ref{fig:comparison_2dforce_sigma20}). Lastly, we tested the networks by providing forcing functions with random perturbed data as input (see table \ref{tab:comparison_force_noise}, Fig.~\ref{fig:comparison_2dforce_noise}).

\subsubsection{\bf{Random forcing functions on $\mathcal{N}(0,5^2)$}}
As shown in table \ref{tab:comparison_2dforce_sigma5}, SpecONet achieves the comparable accuracy to the benchmarking networks despite of not relying on reference solutions. Particularly, the errors of SpecONet surpass those of POD-DON and FNO+RNN for all time and for all the number of reference solutions. For POD-DON, it fails to generalize on the test samples; in contrast, its training goes well as in table \ref{tab:DON_training}. FNO+RNN also fails to infer NSE solutions for $T>T_\ast=0.2$ in time marching sense. Thus, the application of FNO+RNN along with time marching is not effective in predicting NSE solutions. Only if training sample is 600 and $T=T_\ast=0.2$, it has the error under 3.5\%, which is somewhat good. We note that SpecONet has the better performance than FNO for most of the cases since the accuracy FNO is sensitive to the number. However, the errors of SpecONet are slightly over those of FNO with 600 training solutions for $t\geq 0.6$. That is because SpecONet is designed to emulate the time marching numerical scheme, which is the same to error accumulation of the schemes as time progresses. 
\newpage
\begin{table}[hp!]
\centering
\begin{tabular}{|c|c||c||c|c|c|}
\hline
Time&  \makecell{SpecONet \\ (ours)} & \makecell{The number of \\ references} & POD-DON & FNO&FNO+RNN \\
\hline\hline
\multirow{5}{*}{0.2} & \multirow{5}{*}{1.61e-03} & 10 & 1.59e+00 & 5.65e-01 & 1.01e+00 \\
 &  & 50 & 1.41e+00 & 1.96e-01 & 6.34e-01 \\
 &  & 100 & 1.38e+00 & 8.92e-02 & 5.20e-01 \\
 &  & 300 & 1.51e+00 & 1.16e-02 & 4.06e-01 \\
 &  & 600 & 1.27e+00 & 3.02e-03 & 3.45e-02 \\
\hline
\multirow{5}{*}{0.4} & \multirow{5}{*}{4.14e-03} & 10 & 1.30e+00 & 6.42e-01 & 1.03e+00 \\
 &  & 50 & 1.21e+00 & 2.27e-01 & 6.61e-01 \\
 &  & 100 & 1.32e+00 & 8.98e-02 & 5.67e-01 \\
 &  & 300 & 1.32e+00 & 1.18e-02 & 4.66e-01 \\
 &  & 600 & 1.40e+00 & 4.32e-03 & 2.61e-01 \\
\hline
\multirow{5}{*}{0.6} & \multirow{5}{*}{7.24e-03} & 10 & 1.25e+00 & 6.67e-01 & 1.04e+00 \\
 &  & 50 & 1.20e+00 & 2.15e-01 & 7.01e-01 \\
 &  & 100 & 1.16e+00 & 1.00e-01 & 6.27e-01 \\
 &  & 300 & 1.22e+00 & 1.37e-02 & 5.34e-01 \\
 &  & 600 & 1.14e+00 & 5.09e-03 & 3.86e-01 \\
\hline
\multirow{5}{*}{0.8} & \multirow{5}{*}{1.04e-02} & 10 & 1.20e+00 & 6.69e-01 & 1.05e+00 \\
 &  & 50 & 1.14e+00 & 2.45e-01 & 7.58e-01 \\
 &  & 100 & 1.14e+00 & 7.96e-02 & 6.97e-01 \\
 &  & 300 & 1.10e+00 & 1.59e-02 & 6.17e-01 \\
 &  & 600 & 1.20e+00 & 6.07e-03 & 5.03e-01 \\
\hline
\multirow{5}{*}{1.0} & \multirow{5}{*}{1.14e-02} & 10 & 1.17e+00 & 6.87e-01 & 1.06e+00 \\
 &  & 50 & 1.07e+00 & 1.81e-01 & 8.34e-01 \\
 &  & 100 & 1.10e+00 & 8.22e-02 & 7.89e-01 \\
 &  & 300 & 1.08e+00 & 1.86e-02 & 7.24e-01 \\
 &  & 600 & 1.09e+00 & 7.49e-03 & 6.38e-01 \\
\hline
\end{tabular}
\caption{Comparison on errors of various networks over 2D forcing functions generated in $\mathcal{N}(0,5^2)$. The errors are averaged over the Rel.$L^2_x$ errors of inferences from 100 unseen data samples at each time step: 0.2, 0.4, 0.6, 0.8, and 1. Note that whereas our method did not use reference solutions to train, POD-DON, FNO and FNO+RNN used 10, 50, 100, 300, or 600 reference solutions to train. 
}\label{tab:comparison_2dforce_sigma5}
\end{table}
\begin{table}[hp!]
\centering
\begin{tabular}{|c||c|c|c|c|c|}
\hline
Time & 0.2& 0.4&0.6&0.8&1.0 \\
\hline\hline
\makecell{POD-DNO \\ (training data)}  &1.69e-03& 2.37e-03&2.68e-03&3.38e-03&3.97e-03  \\
\hline
\end{tabular}
\caption{The training errors of POD-DON using 600 reference solutions in Rel.$L^2_x$ sense.  }\label{tab:DON_training}
\end{table}

\newpage
 \begin{figure}[th!]
 \centering
 \includegraphics[width=\linewidth]{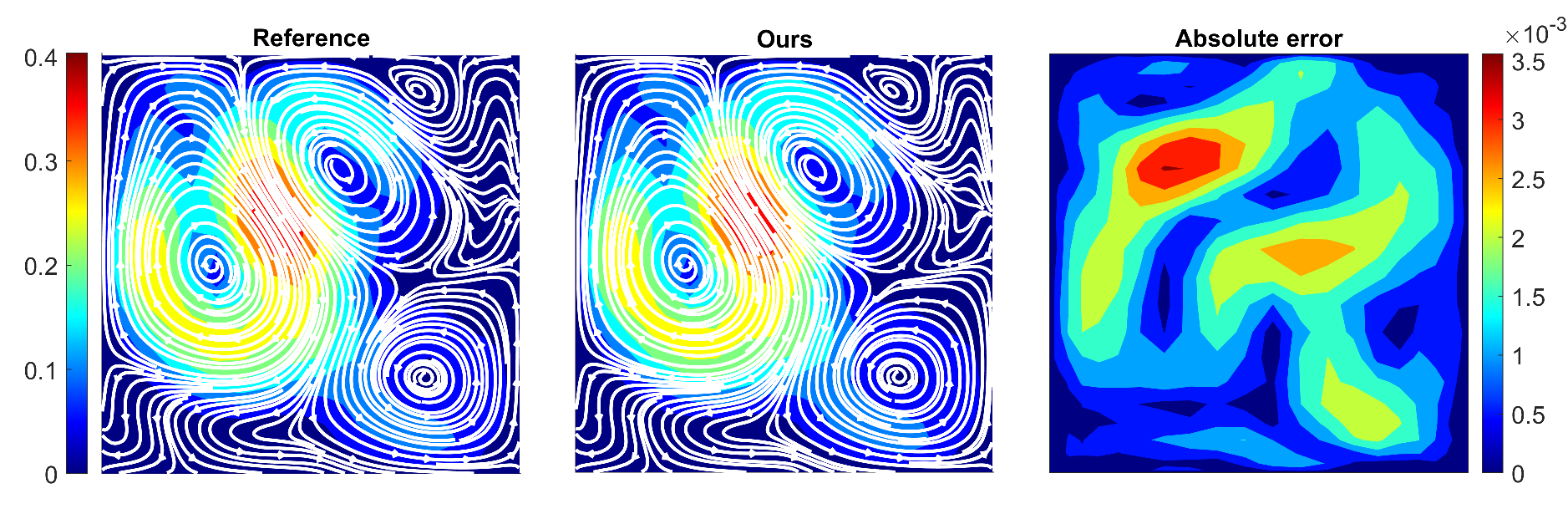}\\
 \includegraphics[width=\linewidth]{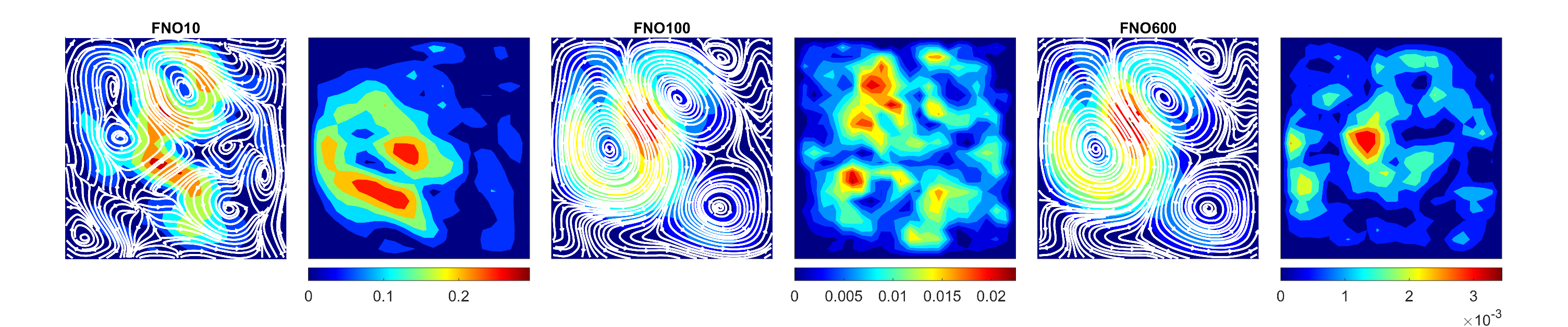}\\
 \includegraphics[width=\linewidth]{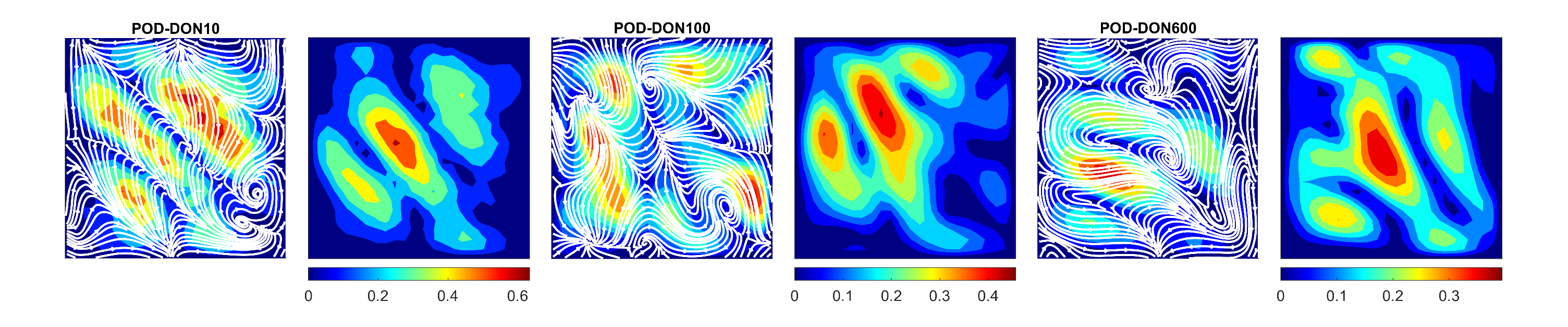}\\
 \includegraphics[width=\linewidth]{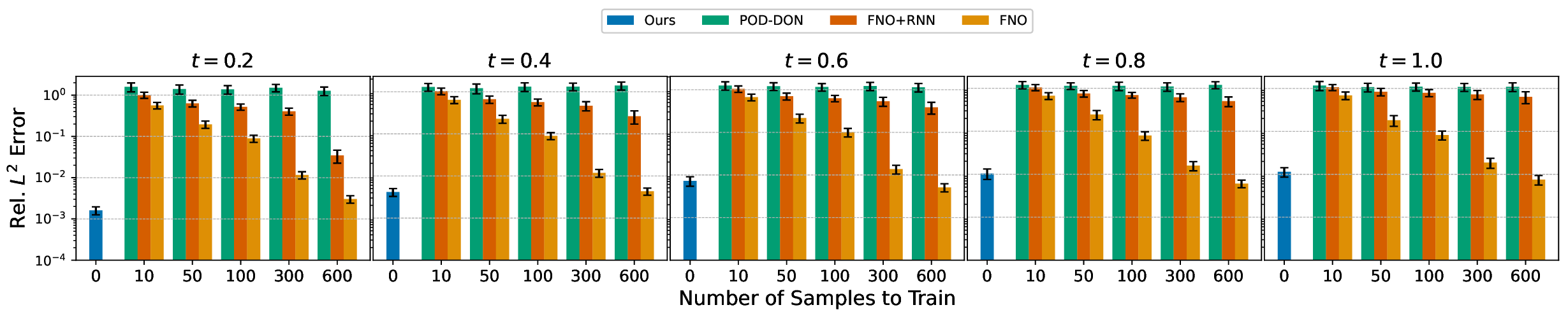}   
 \caption{\textbf{Experiments for 2D forcing functions randomly generated by \eqref{generation:2d_force}  from $N(0,5^2)$.} \textbf{(Top)} Magnitude of a reference solution with its stream line at $T=1$ in the left panel,  magnitude of the corresponding inference of SpecONet with its stream line in the middle panel, and magnitude of the pointwise error between them in the right panel. \textbf{(Middle)} The Inferences and errors of FNO and POD-DON upon the same input. Note that FNO100 and POD-DON100 both used 100 reference solutions to train, and FNO600 and POD-DON600 both did 600 ones. \textbf{(Bottom)} the Rel.$L^2_x$ errors are displayed for different networks, varying numbers of reference solutions to train, and at each time point (see table \ref{tab:comparison_2dforce_sigma5}).}\label{fig:comparison_2dforce_sigma5}
 \end{figure}

\newpage
\subsubsection{\bf{Random forcing functions on $\mathcal{N}(0,10^2)$}} This test data are sampled in a more complicated manner than the training data, which was sampled from $\mathcal{N}(0,5^2)$. In this case, the Rel.$L^2_x$ errors of SpecONet are, at largest, 2.5\% better than all the errors of POD-DON and FNO. It implies that SpecONet is more robust for more complicated test samples than POD-DON and FNO. 

\begin{table}[!h]
\centering
\begin{tabular}{|c|c||c||c|c|}
\hline
Time&  \makecell{SpecONet \\ (ours)} & \makecell{The number of \\references}  & POD-DON & FNO \\
\hline\hline
\multicolumn{1}{|c|}{\multirow{5}{*}{0.2}}&\multicolumn{1}{c||}{\multirow{5}{*}{3.19e-03}} & 10&1.72e+00 & 6.06e-01\\
 && 50&2.13e+00 & 3.37e-01\\
 && 100&1.80e+00 & 2.10e-01\\
 && 300&1.60e+00 & 4.04e-02\\
 && 600&1.49e+00 & 1.03e-02\\
\hline
\multicolumn{1}{|c|}{\multirow{5}{*}{0.4}}&\multicolumn{1}{c||}{\multirow{5}{*}{8.20e-03}} & 10&1.35e+00 & 6.81e-01\\
 && 50&1.31e+00 & 3.89e-01\\
 && 100&1.21e+00 & 2.28e-01\\
 && 300&1.17e+00 & 4.04e-02\\
 && 600&1.32e+00 & 1.03e-02\\
\hline
\multicolumn{1}{|c|}{\multirow{5}{*}{0.6}}&\multicolumn{1}{c||}{\multirow{5}{*}{1.45e-02}} & 10&1.19e+00 & 6.88e-01\\
 && 50&1.09e+00 & 3.77e-01\\
 && 100&1.06e+00 & 2.46e-01\\
 && 300&1.08e+00 & 4.79e-02\\
 && 600&1.06e+00 & 1.74e-02\\
\hline
\multicolumn{1}{|c|}{\multirow{5}{*}{0.8}}&\multicolumn{1}{c||}{\multirow{5}{*}{2.18e-02}} & 10&1.06e+00 & 7.13e-01\\
 && 50&1.03e+00 & 4.06e-01\\
 && 100&1.00e+00 & 2.13e-01\\
 && 300&9.96e-01 & 5.57e-02\\
 && 600&1.10e+00 & 2.09e-02\\
\hline
\multicolumn{1}{|c|}{\multirow{5}{*}{1.0}}&\multicolumn{1}{c||}{\multirow{5}{*}{2.49e-02}} & 10&1.02e+00 & 7.20e-01\\
 && 50&9.74e-01 & 3.42e-01\\
 && 100&9.82e-01 & 2.07e-02\\
 && 300&9.76e-01 & 6.55e-02\\
 && 600&1.04e+00 & 2.64e-02\\
\hline
\end{tabular}
\caption{Comparison on errors of various networks over 2D forcing functions generated in $\mathcal{N}(0,10^2)$. The errors are averaged over the Rel.$L^2_x$ errors of inferences from 100 unseen data samples at each time step: 0.2, 0.4, 0.6, 0.8, and 1. Note that whereas our method did not employ reference solutions to train, POD-DON and FNO employed 10, 50, 100, 300, or 600 references to train. 
}\label{tab:comparison_2dforce_sigma10}
\end{table}

\newpage
 \begin{figure}[th!]
 \centering
 \includegraphics[width=\linewidth]{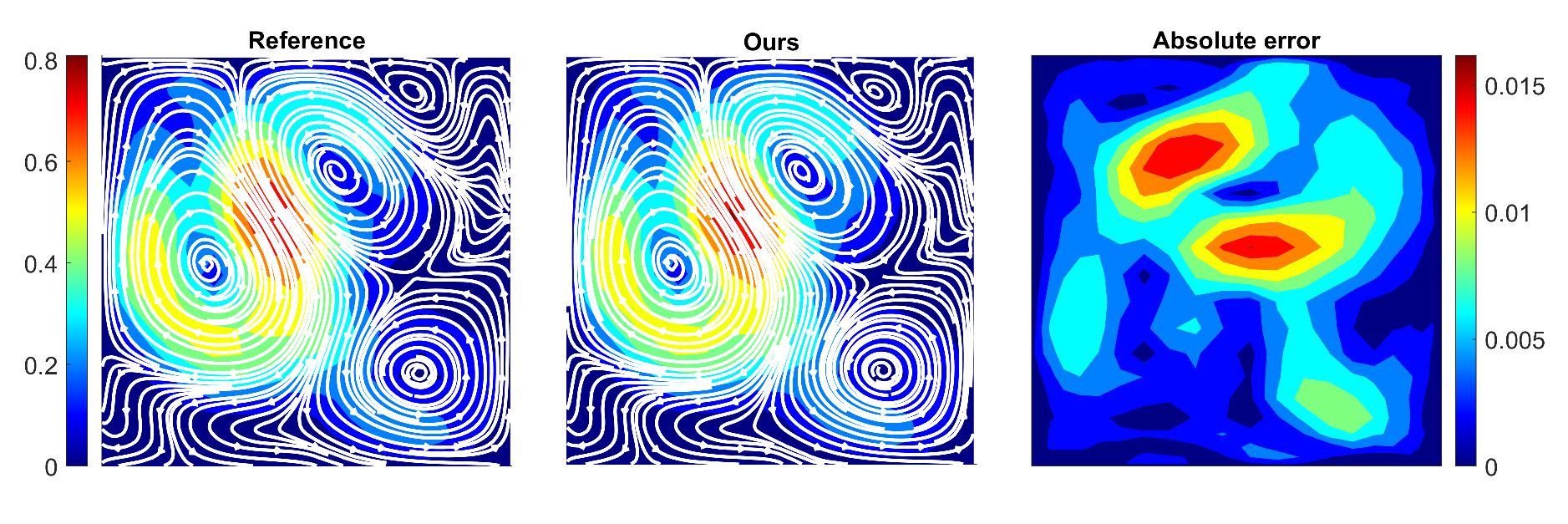}\\
 \includegraphics[width=\linewidth]{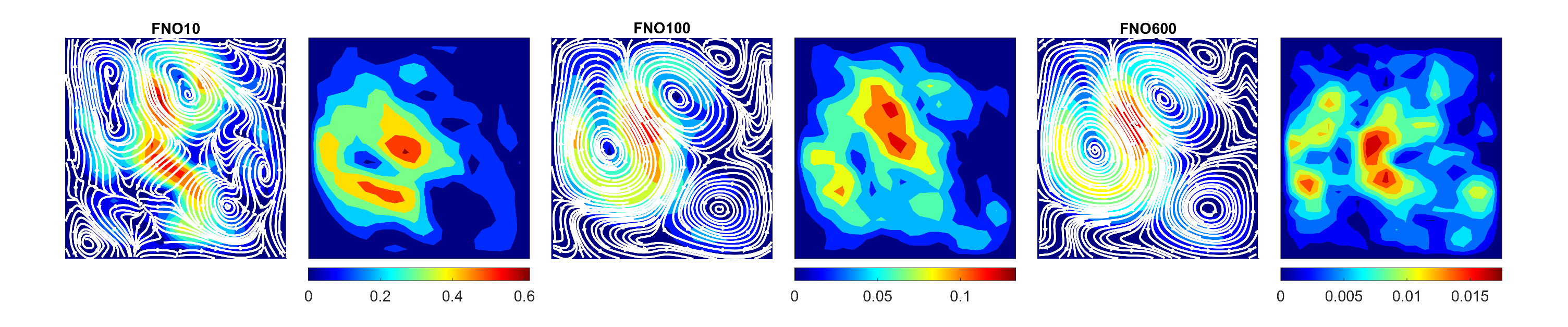}\\
 \includegraphics[width=\linewidth]{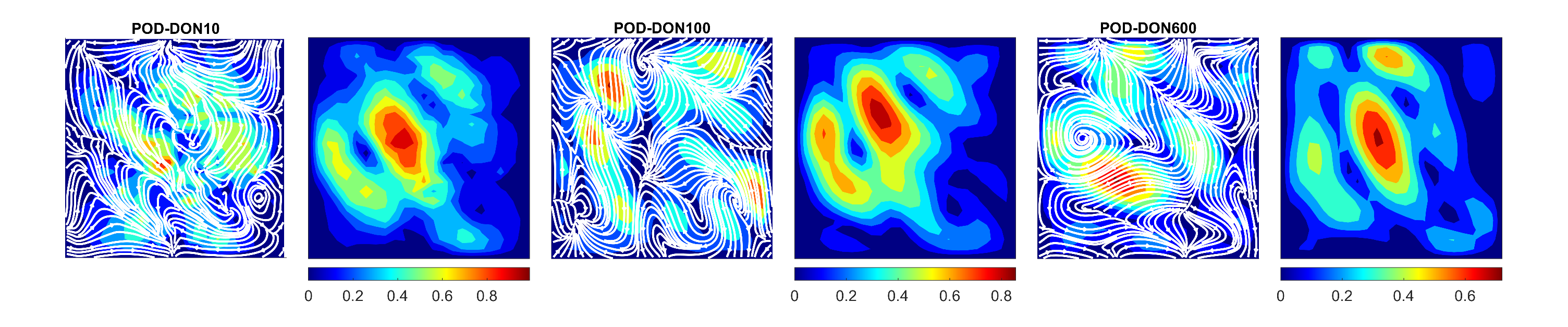}\\
 \includegraphics[width=\linewidth]{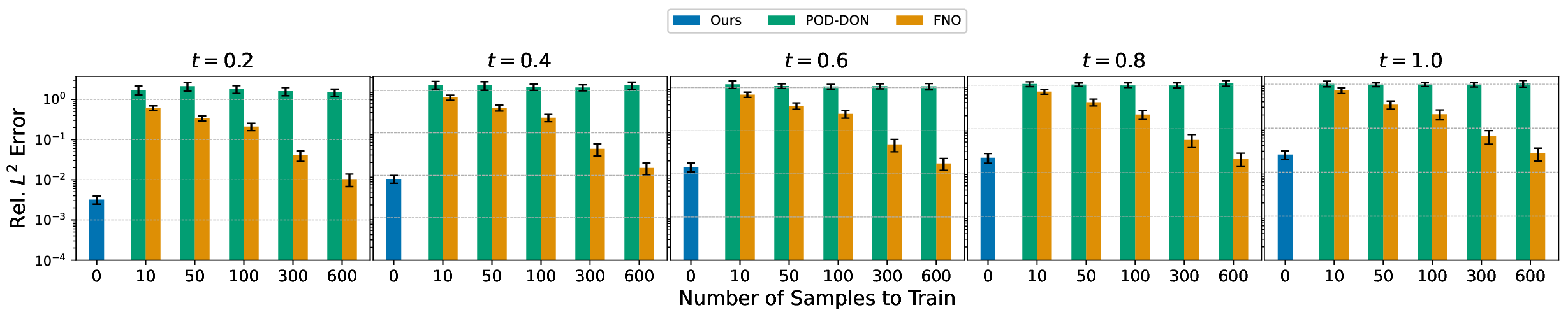}   
 \caption{\textbf{Experiments for 2D forcing functions randomly generated by \eqref{generation:2d_force}  from $N(0,10^2)$.} \textbf{(Top)} Magnitude of a reference solution with its stream line at $T=1$ in the left panel,  magnitude of the corresponding inference of SpecONet with its stream line in the middle panel, and magnitude of the pointwise error between them in the right panel.\textbf{(Middle)} The Inferences and errors of FNO and POD-DON upon the same input. Note that FNO100 and POD-DON100 both used 100 reference solutions to train, and FNO600 and POD-DON600 both did 600 ones. \textbf{(Bottom)} the Rel.$L^2_x$ errors are displayed for different networks, varying numbers of reference solutions to train, and at each time point (see table\ref{tab:comparison_2dforce_sigma10}).}\label{fig:comparison_2dforce_sigma10}
 \end{figure}

\newpage
\subsubsection{\bf{Random forcing functions on $\mathcal{N}(0,20^2)$}} This test data are sampled even more complicatedly than the training data sampled from $\mathcal{N}(0,5^2)$. In this case, the Rel.$L^2_x$ errors of SpecONet grows less from the errors of $\mathcal{N}(0,5^2)$ than those of POD-DON and FNO. It means that SpecONet is relatively reliable comparing to POD-DON and FNO even though test samples becomes more complicated. 
\begin{table}[h!]
\centering
\begin{tabular}{|c|c||c||c|c|}
\hline
Time&  \makecell{SpecONet \\ (ours)} & \makecell{The number of \\references}  & POD-DON & FNO \\
\hline\hline
\multicolumn{1}{|c|}{\multirow{5}{*}{0.2}}&\multicolumn{1}{c||}{\multirow{5}{*}{6.39e-03}} & 10&2.17e+00 & 6.39e-01\\
 && 50&3.04e+00 & 4.57e-01\\
 && 100&2.05e+00 & 3.49e-01\\
 && 300&1.50e+00 & 1.26e-02\\
 && 600&1.53e+00 & 4.13e-02\\
\hline
\multicolumn{1}{|c|}{\multirow{5}{*}{0.4}}&\multicolumn{1}{c||}{\multirow{5}{*}{1.66e-02}} & 10&1.53e+00 & 7.26e-01\\
 && 50&1.31e+00 & 5.31e-01\\
 && 100&1.12e+00 & 3.98e-01\\
 && 300&1.10e+00 & 1.36e-02\\
 && 600&1.30e+00 & 5.76e-02\\
\hline
\multicolumn{1}{|c|}{\multirow{5}{*}{0.6}}&\multicolumn{1}{c||}{\multirow{5}{*}{3.18e-02}} & 10&1.17e+00 & 7.19e-01\\
 && 50&1.03e+00 & 5.19e-01\\
 && 100&1.02e+00 & 4.18e-01\\
 && 300&1.04e+00 & 1.52e-02\\
 && 600&1.07e+00 & 6.68e-02\\
\hline
\multicolumn{1}{|c|}{\multirow{5}{*}{0.8}}&\multicolumn{1}{c||}{\multirow{5}{*}{5.20e-02}} & 10&1.02e+00 & 7.48e-01\\
 && 50&1.01e+00 & 5.59e-01\\
 && 100&9.89e-01 & 4.00e-01\\
 && 300&9.82e-01 & 1.72e-01\\
 && 600&1.10e+00 & 8.03e-02\\
\hline
\multicolumn{1}{|c|}{\multirow{5}{*}{1.0}}&\multicolumn{1}{c||}{\multirow{5}{*}{6.31e-02}} & 10&9.95e-01 & 7.58e-01\\
 && 50&9.82e-01 & 5.20e-01\\
 && 100&9.75e-01 & 3.90e-01\\
 && 300&9.66e-01 & 2.02e-01\\
 && 600&1.04e+00 & 1.02e-01\\
\hline
\end{tabular}
\caption{Comparison on errors of various networks over 2D forcing functions generated in $\mathcal{N}(0,20^2)$. The errors are averaged over the Rel.$L^2_x$ errors of inferences from 100 unseen data samples at each time step: 0.2, 0.4, 0.6, 0.8, and 1. Note that whereas our method did not employ reference solutions to train, POD-DON and FNO employed 10, 50, 100, 300, or 600 references to train. 
}\label{tab:comparison_2dforce_sigma20}
\end{table}
\newpage
 \begin{figure}[th!]
 \centering
 \includegraphics[width=\linewidth]{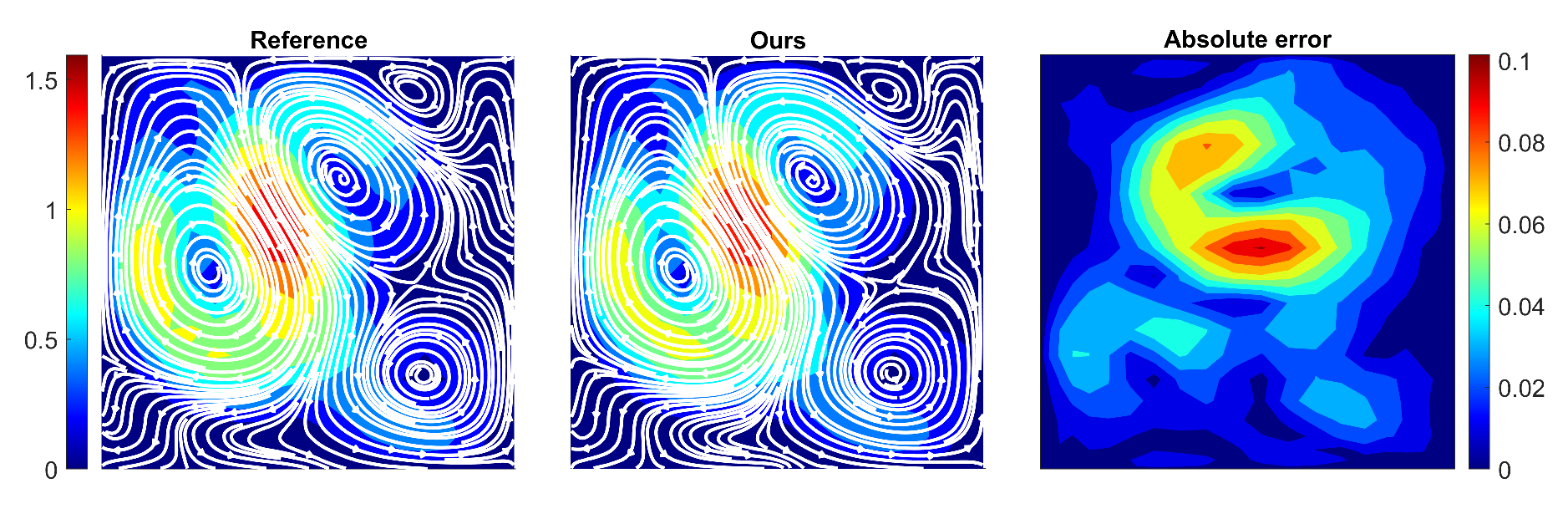}\\
 \includegraphics[width=\linewidth]{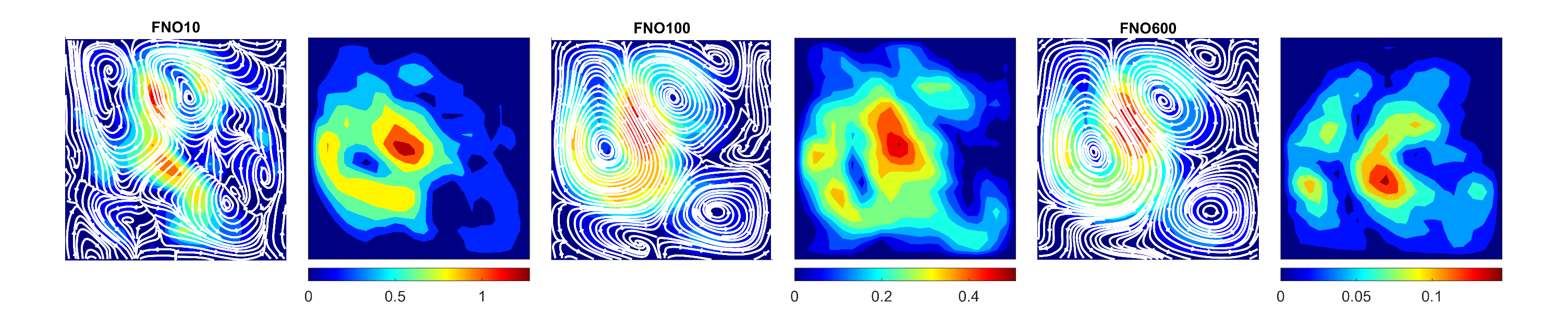}\\
 \includegraphics[width=\linewidth]{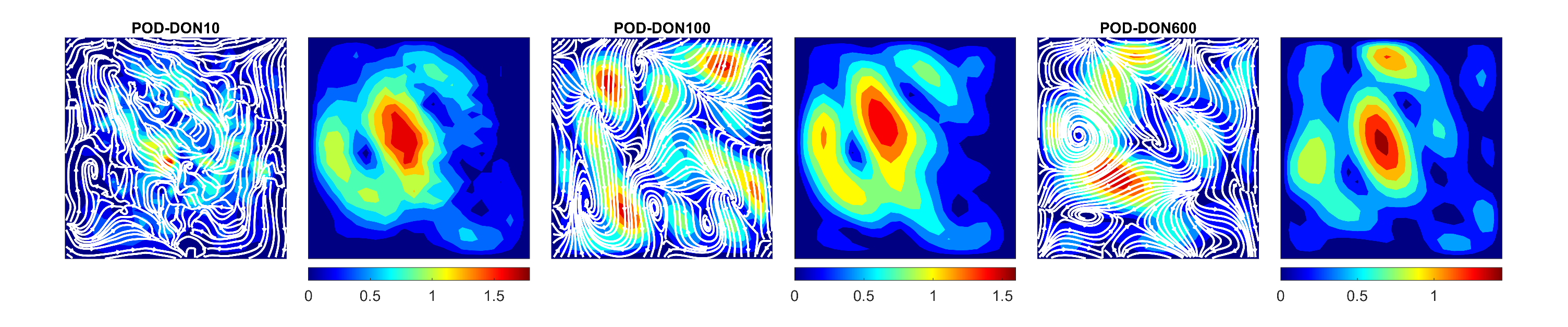}\\
 \includegraphics[width=\linewidth]{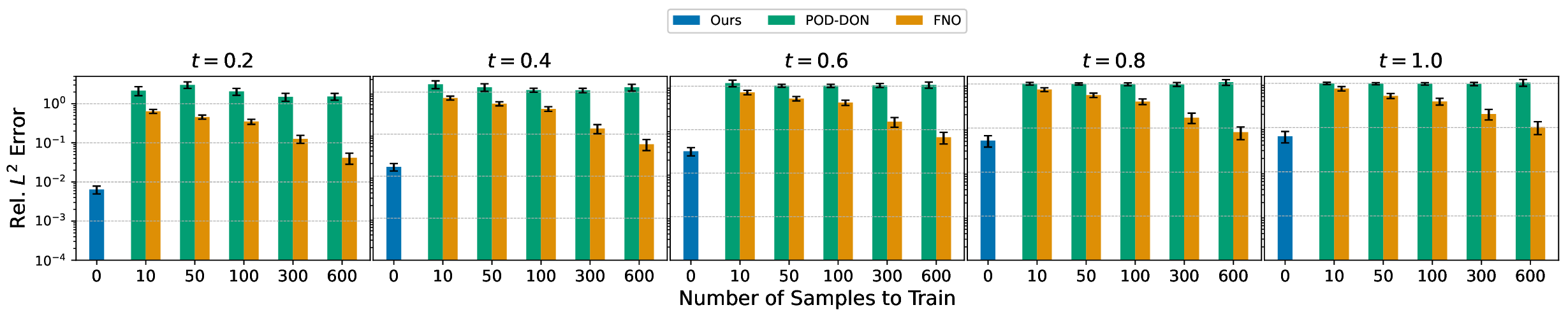}   
 \caption{\textbf{Experiments for 2D forcing functions randomly generated by \eqref{generation:2d_force} from $N(0,20^2)$.} \textbf{(Top)} Magnitude of a reference solution with its stream line at $T=1$ in the left panel,  magnitude of the corresponding inference of SpecONet with its stream line in the middle panel, and magnitude of the pointwise error between them in the right panel.\textbf{(Middle)} The Inferences and errors of FNO and POD-DON upon the same input. Note that FNO100 and POD-DON100 both used 100 reference solutions to train, and FNO600 and POD-DON600 both did 600 ones. \textbf{(Bottom)} the Rel.$L^2_x$ errors are displayed for different networks, varying numbers of reference solutions to train, and at each time point (see table\ref{tab:comparison_2dforce_sigma20}).}\label{fig:comparison_2dforce_sigma20}
 \end{figure}

\newpage
\subsubsection{\bf{Forcing functions with random perturbed data}}\label{sub:perturbed data}
In this section, we tested networks on randomly perturbed data. The framework of this test can be applied to ensemble computing on various fields such as weather prediction \cite{kalnay2003atmospheric,bauer2015quiet}, biology \cite{cao2020ensemble}, stock-market prediction \cite{nti2020comprehensive}, and so fourth. To this end, first, we considered a clean forcing function as 
\begin{align}
   \tilde{f}_x&=1.5\sin(t)((1+\cos(y)-\sin(x)-\sin(x+y))\\
   \tilde{f}_y&=1.5\sin(t)((1+\sin(y)-\cos(x)-\cos(x+y)).
\end{align}
Subsequently, we added artificial perturbations to the clean function denoted by $\epsilon_x$, $\epsilon_y$ as the real part of
\begin{align}  \label{force2d_epsilon}
\frac{1}{24}\sin(t)\sum_{k_x,k_y=0}^{2}c_{k_xk_y}\exp(i(k_xx+k_yy)) .
\end{align}
Here $c_{k_xk_y}$ were constructed as $a_{k_xk_y}+ib_{k_xk_y}$ after randomly choosing $a_{k_xk_y}$ and $b_{k_xk_y}$ from $\mathcal{N}(0,5^2)$. Based on the formulations above, we produced 100 sets of a forcing pair, $(f_x^m,f_y^m)$ such that
\begin{align}\label{generation:force2d_noise}
\begin{split}
   f_x^m&=\tilde{f}_x+\epsilon_{x}^m\\
   f_y^m&=\tilde{f}_y+\epsilon_{y}^m,
\end{split}
\end{align}
for $m=1,2,\cdots, 100$. Note that this 100 sets were totally unseen data from the training data .

As a result, the impact of the perturbations on accuracy is the least for SpecONet compared to FNO and POD-DON. This leads to the fact that SpecONet is a more reliable for ensemble computing than the others.

\newpage
\begin{table}[h!]
\centering
\begin{tabular}{|c|c||c|c|c|}
\hline
Time&  \makecell{SpecONet \\ (ours)} & \makecell{The number of \\ references}  & POD-DON & FNO \\
\hline\hline
\multirow{5}{*}{0.2} & \multirow{5}{*}{2.97e-03} & 10 & 1.02e+00 & 5.13e-01 \\
 &  & 50 & 2.43e+00 & 3.24e-01 \\
 &  & 100 & 1.51e+00 & 2.03e-01 \\
 &  & 300 & 1.54e+00 & 4.93e-02 \\
 &  & 600 & 1.36e+00 & 1.49e-02 \\
\hline
\multirow{5}{*}{0.4} & \multirow{5}{*}{5.73e-03} & 10 & 9.62e-01 & 6.29e-01 \\
 &  & 50 & 1.38e+00 & 3.68e-01 \\
 &  & 100 & 1.02e+00 & 2.37e-01 \\
 &  & 300 & 1.16e+00 & 4.57e-02 \\
 &  & 600 & 1.26e+00 & 1.86e-02 \\
\hline
\multirow{5}{*}{0.6} & \multirow{5}{*}{1.28e-02} & 10 & 9.44e-01 & 6.73e-01 \\
 &  & 50 & 9.16e-01 & 3.77e-01 \\
 &  & 100 & 7.92e-01 & 2.45e-01 \\
 &  & 300 & 9.25e-01 & 5.85e-02 \\
 &  & 600 & 1.01e+00 & 1.97e-02 \\
\hline
\multirow{5}{*}{0.8} & \multirow{5}{*}{2.08e-02} & 10 & 9.10e-01 & 6.66e-01 \\
 &  & 50 & 9.52e-01 & 4.39e-01 \\
 &  & 100 & 8.85e-01 & 2.14e-01 \\
 &  & 300 & 1.08e+00 & 6.53e-02 \\
 &  & 600 & 1.27e+00 & 2.62e-02 \\
\hline
\multirow{5}{*}{1.0} & \multirow{5}{*}{2.35e-02} & 10 & 9.53e-01 & 6.72e-01 \\
 &  & 50 & 8.23e-01 & 3.34e-01 \\
 &  & 100 & 7.99e-01 & 2.32e-01 \\
 &  & 300 & 1.05e+00 & 6.92e-02 \\
 &  & 600 & 1.18e+00 & 3.59e-02 \\
\hline
\end{tabular}
\caption{Comparison on errors of various networks over 2D forcing functions generated by \eqref{generation:force2d_noise}. The errors are averaged over the Rel.$L^2_x$ errors of inferences from 100 unseen data samples at each time step: 0.2, 0.4, 0.6, 0.8, and 1. Note that whereas our method did not employ reference solutions to train, POD-DON and FNO employed 10, 50, 100, 300, or 600 references to train. 
}\label{tab:comparison_force_noise}
\end{table}
\newpage
 \begin{figure}[th!]
 \centering
 \includegraphics[width=\linewidth]{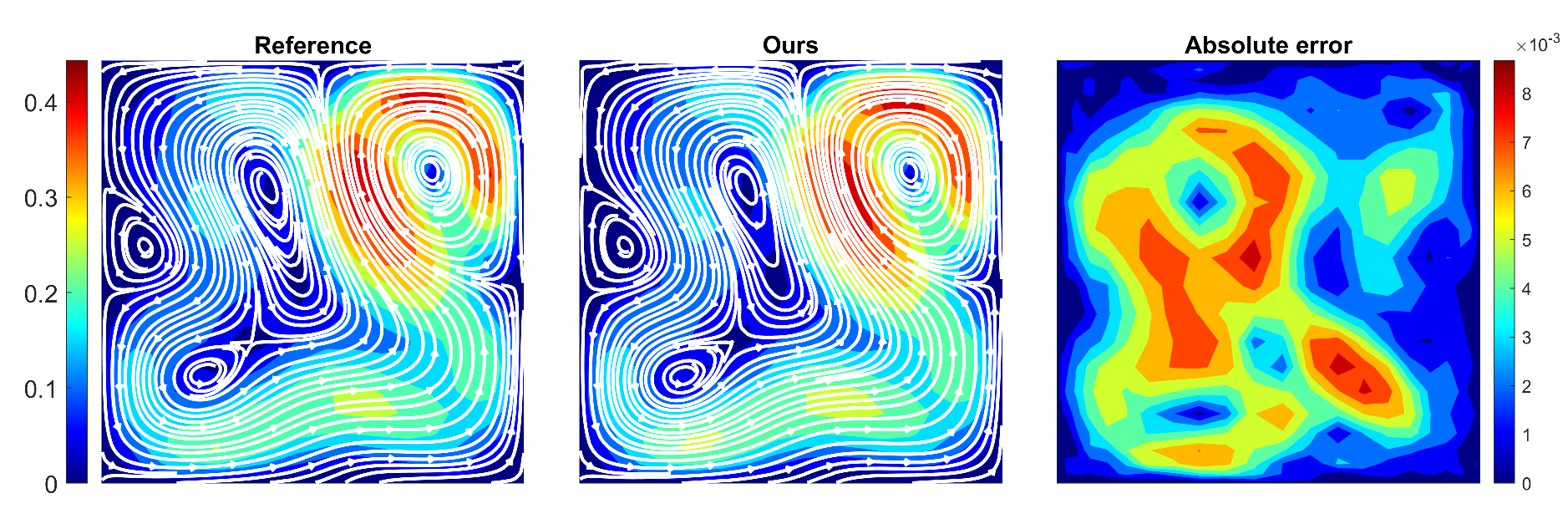}\\
 \includegraphics[width=\linewidth]{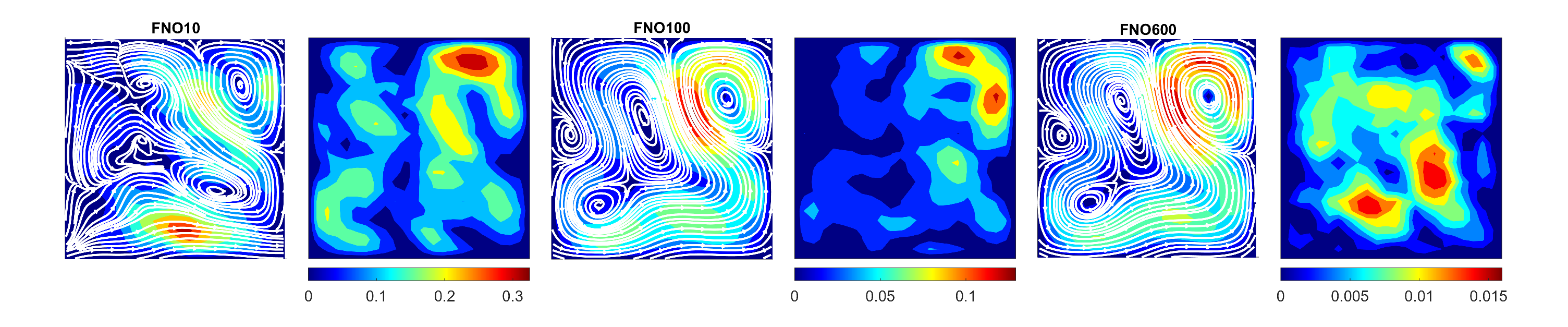}\\
 \includegraphics[width=\linewidth]{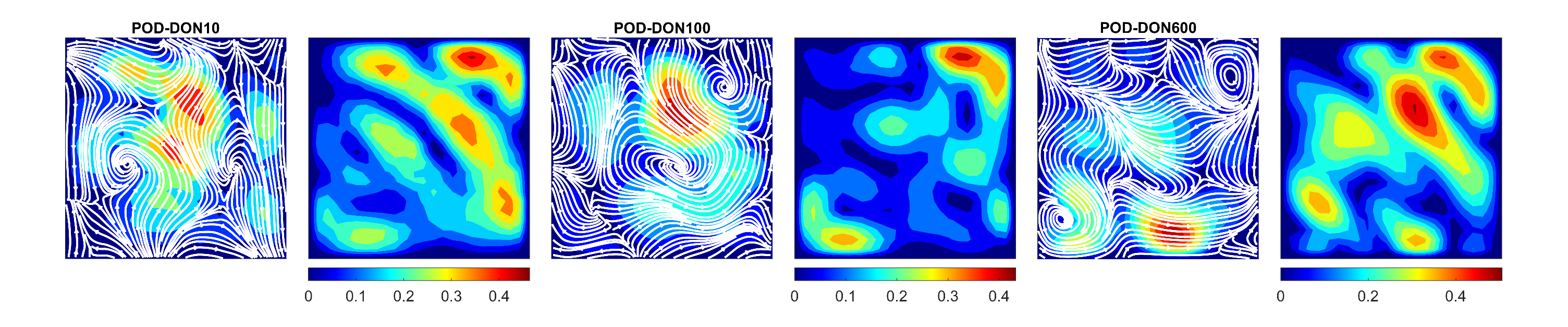}\\
 \includegraphics[width=\linewidth]{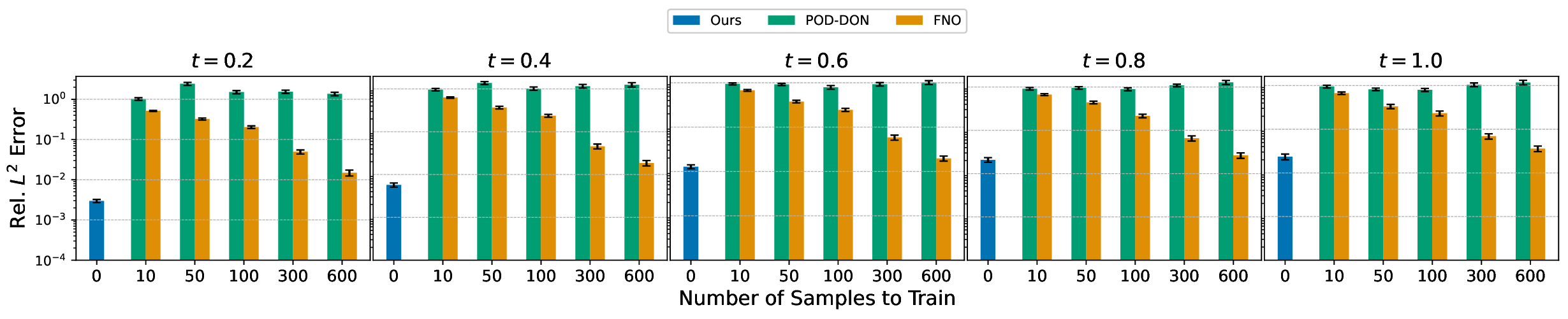}   
 \caption{\textbf{Experiments for 2D perturbed forcing functions generated by \eqref{generation:force2d_noise}.} \textbf{(Top)} Magnitude of a reference solution with its stream line at $T=1$ in the left panel,  magnitude of the corresponding inference of SpecONet with its stream line in the middle panel, and magnitude of the pointwise error between them in the right panel.\textbf{(Middle)} The Inferences and errors of FNO and POD-DON upon the same input. Note that FNO100 and POD-DON100 both used 100 reference solutions to train, and FNO600 and POD-DON600 both did 600 ones. \textbf{(Bottom)} the Rel.$L^2_x$ errors are displayed for different networks, varying numbers of reference solutions to train, and at each time point (see table \ref{tab:comparison_force_noise}).}\label{fig:comparison_2dforce_noise}
 \end{figure}

\newpage
\subsection{Two dimensional NSE with random initial conditions}\label{sub:2d_initial}
We executed experiments to evaluate the accuracy of inference upon random initial conditions as input. The random initial conditions were defined by
\begin{align}\label{generation:2dinitial}
   (u_0,v_0)=(-\partial_y\Psi,\partial_x\Psi)
\end{align}
where $\Psi(x,y)$ is the real part of $\frac{1}{240}\sin(t)\sum_{k_x,k_y=0}^{2}c_{k_xk_y}\exp(i(k_xx+k_yy))$. In addition, $c_{k_xk_y}=a_{k_xk_y}+ib_{k_xk_y}$ where $a_{k_xk_y}$, and $b_{k_xk_y}$ were random variables. For training input data, we produced 600 initial conditions whose the random variables were sampled from $\mathcal{N}(0,5^2)$. The other information was set as in table \ref{tab:2d_initial}. 

In order for comparison, we carried out training on POD-DON and FNO with the initial conditions and the corresponding reference solutions at $T = 0.2, 0.4, 0.6, 0.8$ and $1$, which
were identical to the initial conditions used for training SpecONet. Note that the number of training samples for FNO and POD-DON varied from 10 to 600. This variance will underscore that accuracy of SpecONet does not depend on the number of samples whereas accuracy of POD-DON and FNO are sensitive to the number of training samples.   

For test samples, we made 100 initial conditions from three cases, $\mathcal{N}(0,5^2)$,  $\mathcal{N}(0,9^2)$, and $\mathcal{N}(0,13^2)$. Afterwards, velocity
solutions to NSEs at $T = 0.2, 0.4, 0.6, 0.8$, and $1$ were computed to employ them as reference solutions. Then, the sets of 100 errors from the three distributions were computed in Rel.$L^2_x$ sense between inferences of
each methods and the velocity reference solutions for each time step, $T = 0.2, 0.4, 0.6, 0.8$
and $1$. The comparison on the errors made by each method are discussed in
the following subsections: a) $\mathcal{N}(0,5^2)$ (see table \ref{tab:comparison_2dinitial_sigma5}, Fig.~\ref{fig:comparison_2dinitial_sigma5}); b) $\mathcal{N}(0,9^2)$ (see table \ref{tab:comparison_2dinitial_sigma9}; Fig.~\ref{fig:comparison_2dinitial_sigma9}); c) $\mathcal{N}(0,13^2)$ (see table \ref{tab:comparison_2dinitial_sigma13}; Fig.~\ref{fig:comparison_2dinitial_sigma13}).

\begin{table}[h!]
    \centering    
    \begin{tabular}{|c|c||c|c|}
    \hline    
    Domain   & $[0,2\pi]^2$  & Boundary condition   & periodic   \\ \hline
    Bases type   & Fourier   & Forcing function   &$f_x=f_y=\sin(x)\sin(y)$    \\ \hline
    $\Delta t$   & 0.01   & The number of time steps& 100       \\ \hline
    $\nu$   & 0.01   & $N$   & 24    \\ \hline
    \end{tabular}
    \caption{Information on the numerical schemes for 2D initial conditions as input}
    \label{tab:2d_initial}
\end{table}

\newpage
\subsubsection{\bf{Random initial conditions on $\mathcal{N}(0,5^2)$}}The Rel.$L^2_x$ errors of SpecONet are less than 0.5\% better than all the errors of POD-DON and FNO. Moreover, whereas the errors of POD-DON, and FNO are sensitive to the number of reference solutions, the errors of SpecONet does not rely on the number of reference solutions. However, SpecONet accumulates errors as time progresses because it emulates the time marching numerical scheme. 

\begin{table}[h!]
\centering
\begin{tabular}{|c|c||c|c|c|}
\hline
Time&  \makecell{SpecONet \\ (ours)} & \makecell{The number of \\ references}  & POD-DON & FNO \\
\hline\hline
\multirow{5}{*}{0.2} & \multirow{5}{*}{4.49e-03} & 10 & 5.19e-01 & 8.59e-02 \\
 &  & 50 & 4.35e-01 & 1.32e-02 \\
 &  & 100 & 3.00e-01 & 9.36e-03 \\
 &  & 300 & 6.88e-02 & 6.41e-03 \\
 &  & 600 & 3.71e-02 & 6.49e-03 \\
\hline
\multirow{5}{*}{0.4} & \multirow{5}{*}{4.64e-03} & 10 & 2.89e-01 & 4.27e-02 \\
 &  & 50 & 2.28e-01 & 1.05e-02 \\
 &  & 100 & 1.73e-01 & 8.27e-03 \\
 &  & 300 & 4.04e-02 & 6.88e-03 \\
 &  & 600 & 2.70e-02 & 6.64e-03 \\
\hline
\multirow{5}{*}{0.6} & \multirow{5}{*}{4.66e-03} & 10 & 1.95e-01 & 3.65e-02 \\
 &  & 50 & 1.80e-01 & 1.25e-02 \\
 &  & 100 & 1.15e-01 & 8.79e-03 \\
 &  & 300 & 3.74e-02 & 6.57e-03 \\
 &  & 600 & 2.64e-02 & 7.13e-03 \\
\hline
\multirow{5}{*}{0.8} & \multirow{5}{*}{4.76e-03} & 10 & 1.34e-01 & 3.47e-02 \\
 &  & 50 & 1.28e-01 & 1.05e-02 \\
 &  & 100 & 9.11e-02 & 7.88e-03 \\
 &  & 300 & 3.40e-02 & 6.79e-03 \\
 &  & 600 & 2.85e-02 & 6.71e-03 \\
\hline
\multirow{5}{*}{1.0} & \multirow{5}{*}{5.01e-03} & 10 & 1.12e-01 & 3.47e-02 \\
 &  & 50 & 1.00e-01 & 1.09e-02 \\
 &  & 100 & 7.65e-02 & 8.17e-03 \\
 &  & 300 & 3.30e-02 & 6.59e-03 \\
 &  & 600 & 2.97e-02 & 6.52e-03 \\
\hline
\end{tabular}
\caption{Comparison on errors of various networks over 2D initial conditions generated in $\mathcal{N}(0,5^2)$. The errors are averaged over the Rel.$L^2_x$ errors of inferences from 100 unseen data samples at each time step: 0.2, 0.4, 0.6, 0.8, and 1. Note that whereas our method did not employ reference solutions to train, POD-DON and FNO employed 10, 50, 100, 300, or 600 references to train. 
}\label{tab:comparison_2dinitial_sigma5}
\end{table}

\newpage
 \begin{figure}[th!]
 \begin{center}
 {\textbf{Numerical results for $\sigma=5$, $T=1$}}
 \includegraphics[width=\linewidth]{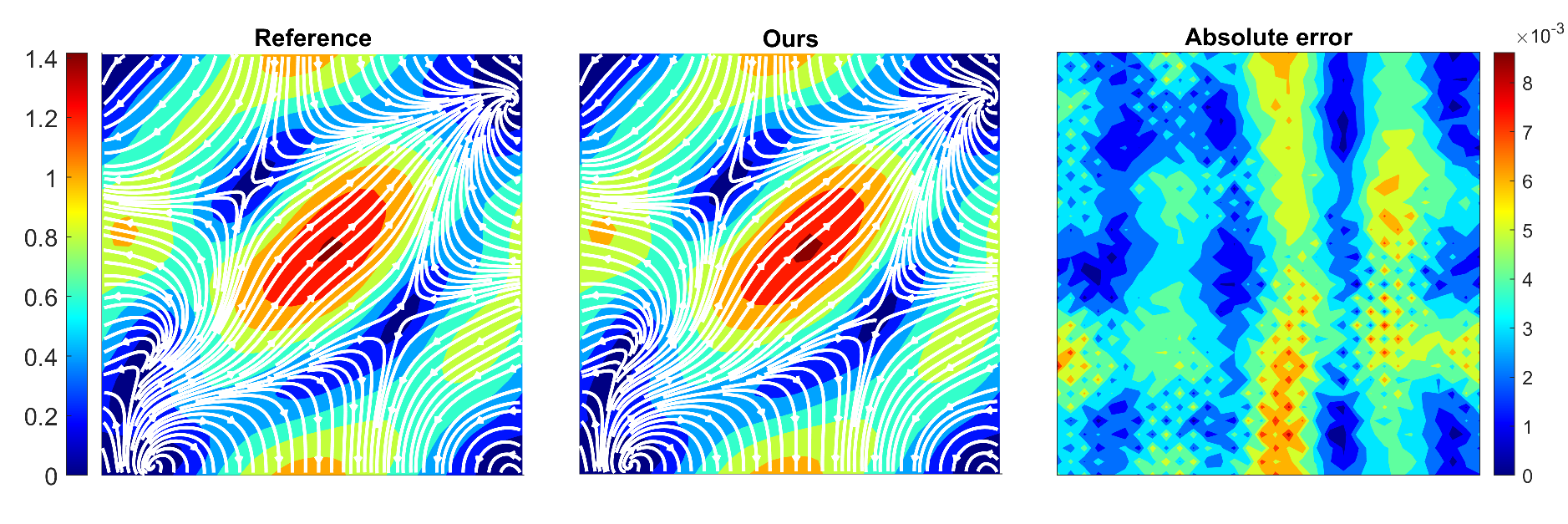}\\
 \includegraphics[width=\linewidth]{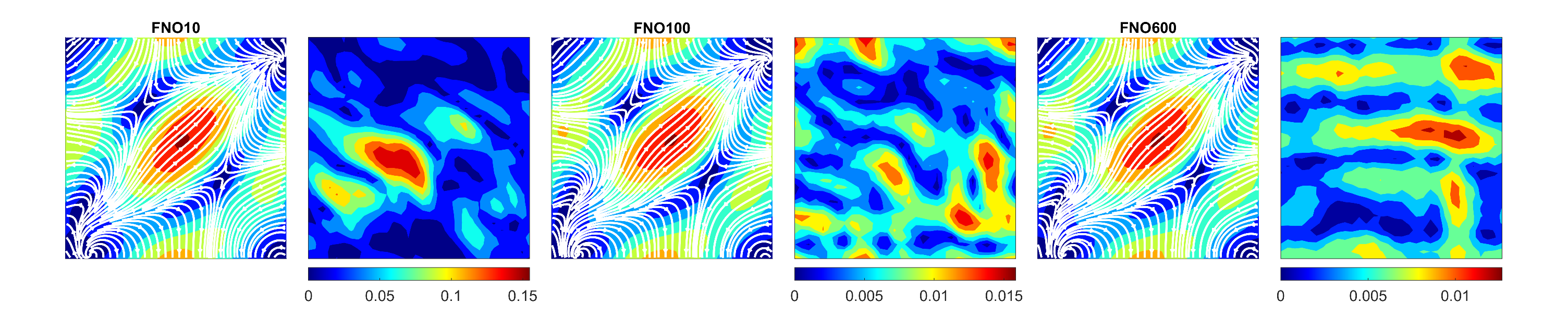}\\
 \includegraphics[width=\linewidth]{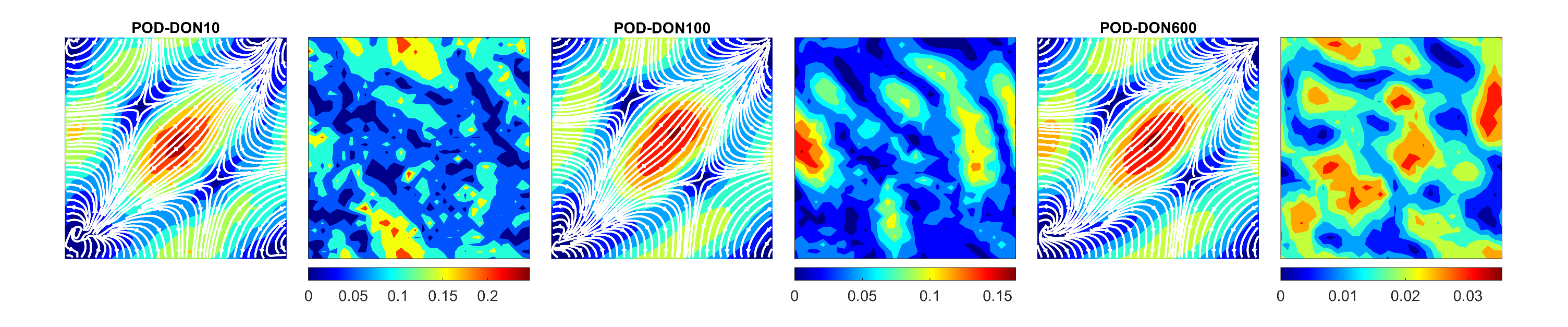}\\
 \includegraphics[width=\linewidth]{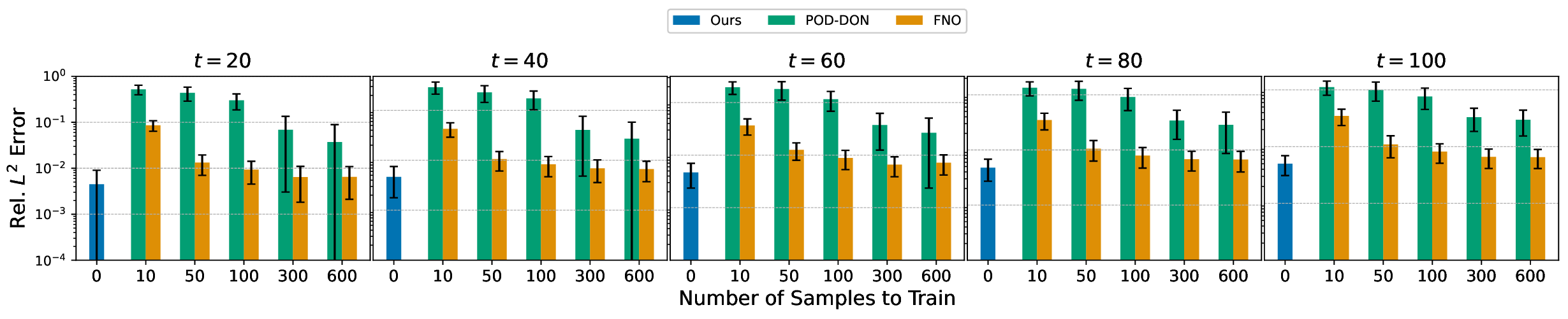}   
 \end{center}
 \caption{\textbf{Experiments for 2D initial conditions randomly generated by \eqref{generation:2dinitial} from $N(0,5^2)$.} \textbf{(Top)} Magnitude of a reference solution with its stream line at $T=1$ in the left panel,  magnitude of the corresponding inference of SpecONet with its stream line in the middle panel, and magnitude of the pointwise error between them in the right panel.\textbf{(Middle)} The Inferences and errors of FNO and POD-DON against the same input. Note that FNO100 and POD-DON100 both used 100 reference solutions to train, and FNO600 and POD-DON600 both did 600 ones. \textbf{(Bottom)} The Rel.$L^2_x$ errors are displayed for different networks, varying numbers of reference solutions to train, and at each time point (see table \ref{tab:comparison_2dinitial_sigma5}).}\label{fig:comparison_2dinitial_sigma5}
 \end{figure}

\newpage
\subsubsection{\bf{Random initial conditions on \texorpdfstring{$\mathcal{N}(0,9^2)$}{N(0,9²)}}}This test data are sampled in a more complicated manner than the training data, which was sampled from $\mathcal{N}(0,5^2)$. In this case, the Rel.$L^2_x$ errors of SpecONet are, at largest, 1.5\% better than all the errors of POD-DON and FNO. It implies that SpecONet is more robust for more complicated test samples than POD-DON and FNO.

\begin{table}[h!]
\centering
\begin{tabular}{|c|c||c|c|c|}
\hline
Time&  \makecell{SpecONet \\ (ours)} & \makecell{The number of \\ references}  & POD-DON & FNO \\
\hline\hline
\multirow{5}{*}{0.2} & \multirow{5}{*}{9.16e-03} & 10 & 7.68e-01 & 1.71e-01 \\
 &  & 50 & 6.99e-01 & 4.65e-02 \\
 &  & 100 & 4.93e-01 & 3.29e-02 \\
 &  & 300 & 1.27e-01 & 2.10e-02 \\
 &  & 600 & 8.43e-02 & 1.82e-02 \\
\hline
\multirow{5}{*}{0.4} & \multirow{5}{*}{1.12e-02} & 10 & 4.97e-01 & 1.08e-01 \\
 &  & 50 & 4.31e-01 & 3.70e-02 \\
 &  & 100 & 3.26e-01 & 3.05e-02 \\
 &  & 300 & 9.05e-02 & 2.25e-02 \\
 &  & 600 & 7.27e-02 & 1.81e-02 \\
\hline
\multirow{5}{*}{0.6} & \multirow{5}{*}{1.19e-02} & 10 & 3.44e-01 & 9.43e-02 \\
 &  & 50 & 3.50e-01 & 4.32e-02 \\
 &  & 100 & 2.30e-01 & 3.17e-02 \\
 &  & 300 & 8.26e-02 & 2.02e-02 \\
 &  & 600 & 7.05e-02 & 2.12e-02 \\
\hline
\multirow{5}{*}{0.8} & \multirow{5}{*}{1.25e-02} & 10 & 2.42e-01 & 8.51e-02 \\
 &  & 50 & 2.53e-01 & 3.81e-02 \\
 &  & 100 & 1.84e-01 & 2.73e-02 \\
 &  & 300 & 7.62e-02 & 2.14e-02 \\
 &  & 600 & 7.35e-02 & 1.97e-02 \\
\hline
\multirow{5}{*}{1.0} & \multirow{5}{*}{1.46e-02} & 10 & 2.06e-01 & 7.84e-02 \\
 &  & 50 & 2.02e-01 & 3.92e-02 \\
 &  & 100 & 1.57e-01 & 2.95e-02 \\
 &  & 300 & 7.39e-02 & 2.18e-02 \\
 &  & 600 & 7.30e-02 & 2.02e-02 \\
\hline
\end{tabular}
\caption{Comparison on errors of various networks over 2D initial conditions generated in $\mathcal{N}(0,9^2)$. The errors are averaged over the Rel.$L^2_x$ errors of inferences from 100 unseen data samples at each time step: 0.2, 0.4, 0.6, 0.8, and 1. Note that whereas our method did not employ reference solutions to train, POD-DON and FNO employed 10, 50, 100, 300, or 600 references to train. 
}\label{tab:comparison_2dinitial_sigma9}
\end{table}

\newpage
 \begin{figure}[th!]
 \begin{center}
 {\textbf{Numerical results for $\sigma=9$, $T=1$}}
 \includegraphics[width=\linewidth]{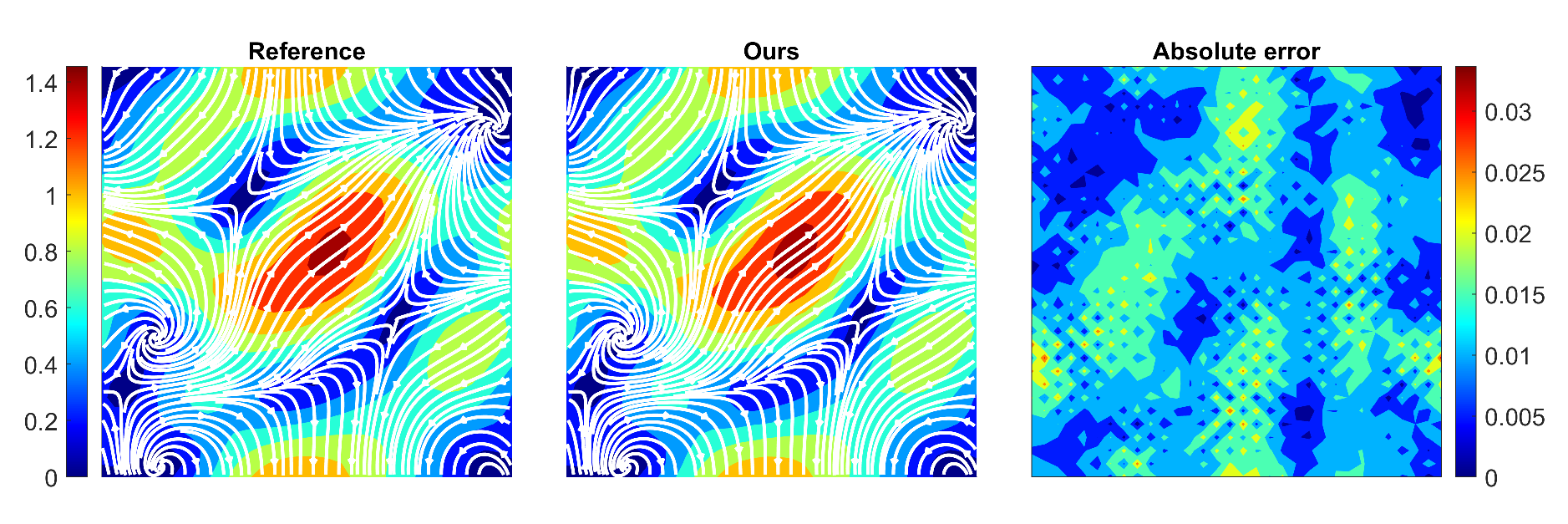}\\
 \includegraphics[width=\linewidth]{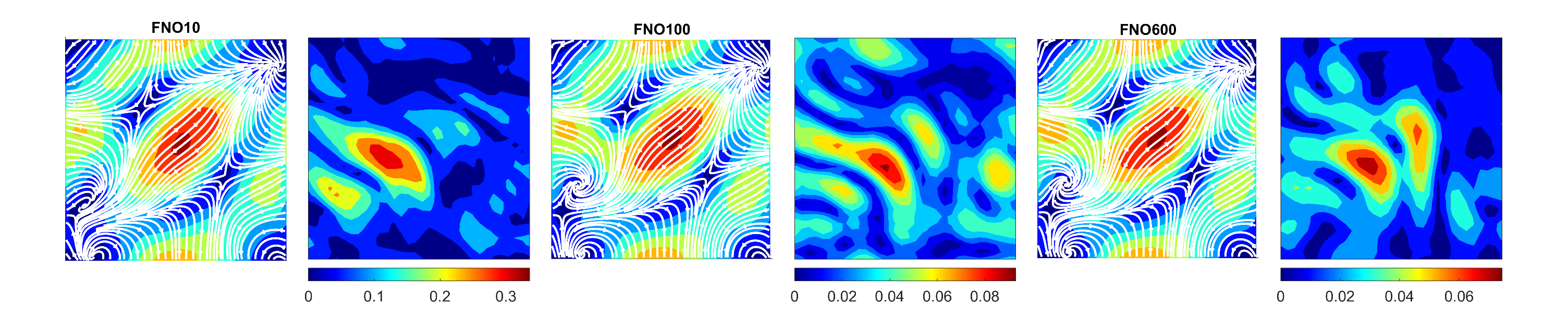}\\
 \includegraphics[width=\linewidth]{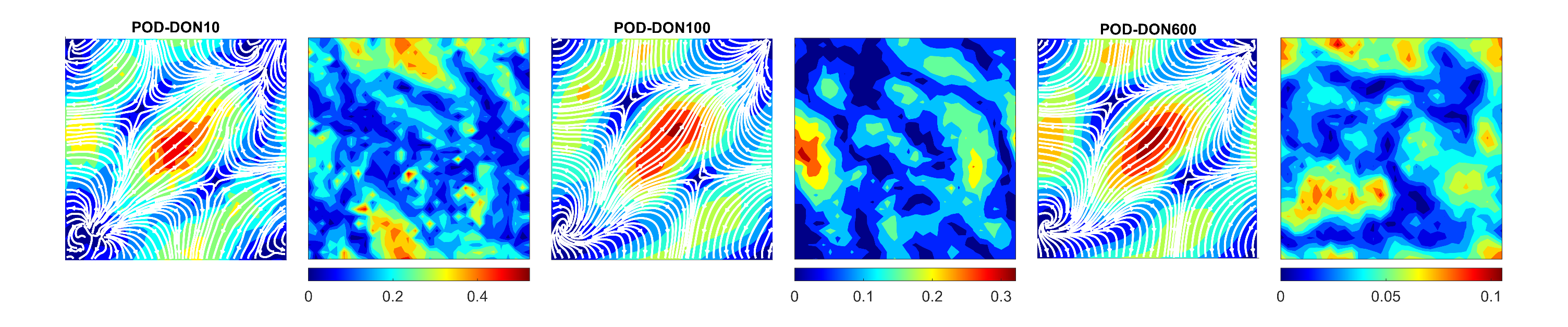}\\
 \includegraphics[width=\linewidth]{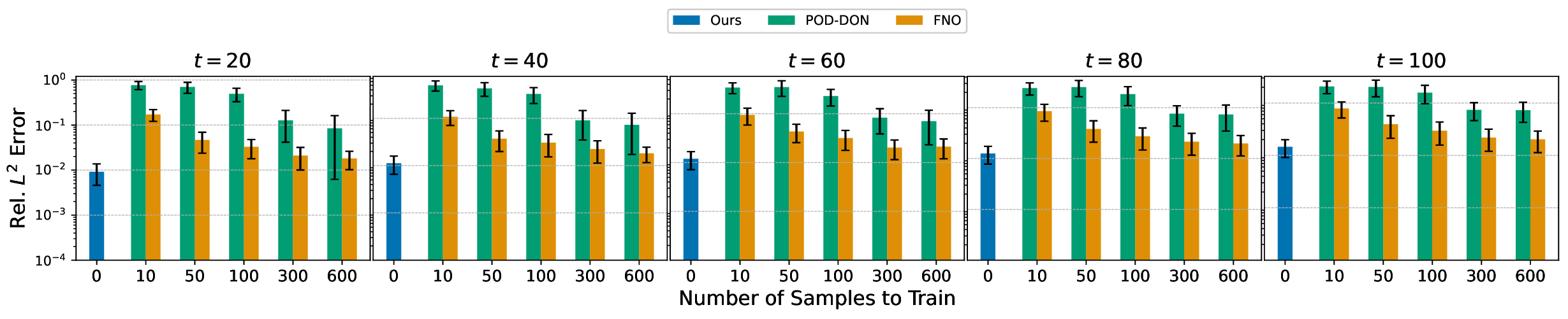}    
 \end{center}\caption{\textbf{Experiments for 2D initial conditions randomly generated by \eqref{generation:2dinitial} on $N(0,9^2)$.} \textbf{(Top)} Magnitude of a reference solution with its stream line at $T=1$ in the left panel,  magnitude of the corresponding inference of SpecONet with its stream line in the middle panel, and magnitude of the pointwise error between them in the right panel. \textbf{(Middle)} The Inferences and errors of FNO and POD-DON over the same input above. Note that FNO100 and POD-DON100 both used 100 reference solutions to train, and FNO600 and POD-DON600 both did 600 ones. \textbf{(Bottom)} The Rel.$L^2_x$ errors are displayed for different networks, varying numbers of training samples, and at each time point (see table\ref{tab:comparison_2dinitial_sigma9}).}\label{fig:comparison_2dinitial_sigma9}
 \end{figure}

\newpage
\subsubsection{\bf{Random initial conditions on $\mathcal{N}(0,13^2)$}}This test data are sampled even more complicatedly than the training data sampled from $\mathcal{N}(0,5^2)$. In this case, the Rel.$L^2_x$ errors of SpecONet increased less than those of POD-DON and FNO from the errors of $\mathcal{N}(0,5^2)$. It means that SpecONet is relatively reliable comparing to POD-DON and FNO even though test samples becomes more complicated.
\begin{table}[h!]
\centering
\begin{tabular}{|c|c||c|c|c|}
\hline
Time&  \makecell{SpecONet \\ (ours)} & \makecell{The number of \\ references}  & POD-DON & FNO \\
\hline\hline
\multirow{5}{*}{0.2} & \multirow{5}{*}{1.42e-02} & 10 & 9.04e-01 & 2.57e-01 \\
 &  & 50 & 8.43e-01 & 9.72e-02 \\
 &  & 100 & 6.07e-01 & 7.04e-02 \\
 &  & 300 & 1.81e-01 & 4.95e-02 \\
 &  & 600 & 1.31e-01 & 3.86e-02 \\
\hline
\multirow{5}{*}{0.4} & \multirow{5}{*}{1.98e-02} & 10 & 6.63e-01 & 1.81e-01 \\
 &  & 50 & 5.87e-01 & 7.93e-02 \\
 &  & 100 & 4.50e-01 & 6.87e-02 \\
 &  & 300 & 1.42e-01 & 5.09e-02 \\
 &  & 600 & 1.24e-01 & 3.55e-02 \\
\hline
\multirow{5}{*}{0.6} & \multirow{5}{*}{2.23e-02} & 10 & 4.76e-01 & 1.66e-01 \\
 &  & 50 & 4.97e-01 & 8.57e-02 \\
 &  & 100 & 3.38e-01 & 6.64e-02 \\
 &  & 300 & 1.32e-01 & 4.53e-02 \\
 &  & 600 & 1.22e-01 & 4.48e-02 \\
\hline
\multirow{5}{*}{0.8} & \multirow{5}{*}{2.41e-02} & 10 & 3.45e-01 & 1.46e-01 \\
 &  & 50 & 3.71e-01 & 7.98e-02 \\
 &  & 100 & 2.75e-01 & 5.97e-02 \\
 &  & 300 & 1.21e-01 & 4.76e-02 \\
 &  & 600 & 1.25e-01 & 4.24e-02 \\
\hline
\multirow{5}{*}{1.0} & \multirow{5}{*}{3.00e-02} & 10 & 2.98e-01 & 1.30e-01 \\
 &  & 50 & 3.00e-01 & 8.09e-02 \\
 &  & 100 & 2.37e-01 & 6.38e-02 \\
 &  & 300 & 1.17e-01 & 5.03e-02 \\
 &  & 600 & 1.22e-01 & 4.73e-02 \\
\hline
\end{tabular}
\caption{Comparison on errors of various networks over 2D initial conditions generated in $\mathcal{N}(0,13^2)$. The errors are averaged over the Rel.$L^2_x$ errors of inferences from 100 unseen data samples at each time step: 0.2, 0.4, 0.6, 0.8, and 1. Note that whereas our method did not employ reference solutions to train, POD-DON and FNO employed 10, 50, 100, 300, or 600 references to train. 
}\label{tab:comparison_2dinitial_sigma13}
\end{table}
\newpage
 \begin{figure}[th!]
 \begin{center}
 {\textbf{Numerical results for $\sigma=13$, $T=1$}}
 \includegraphics[width=\linewidth]{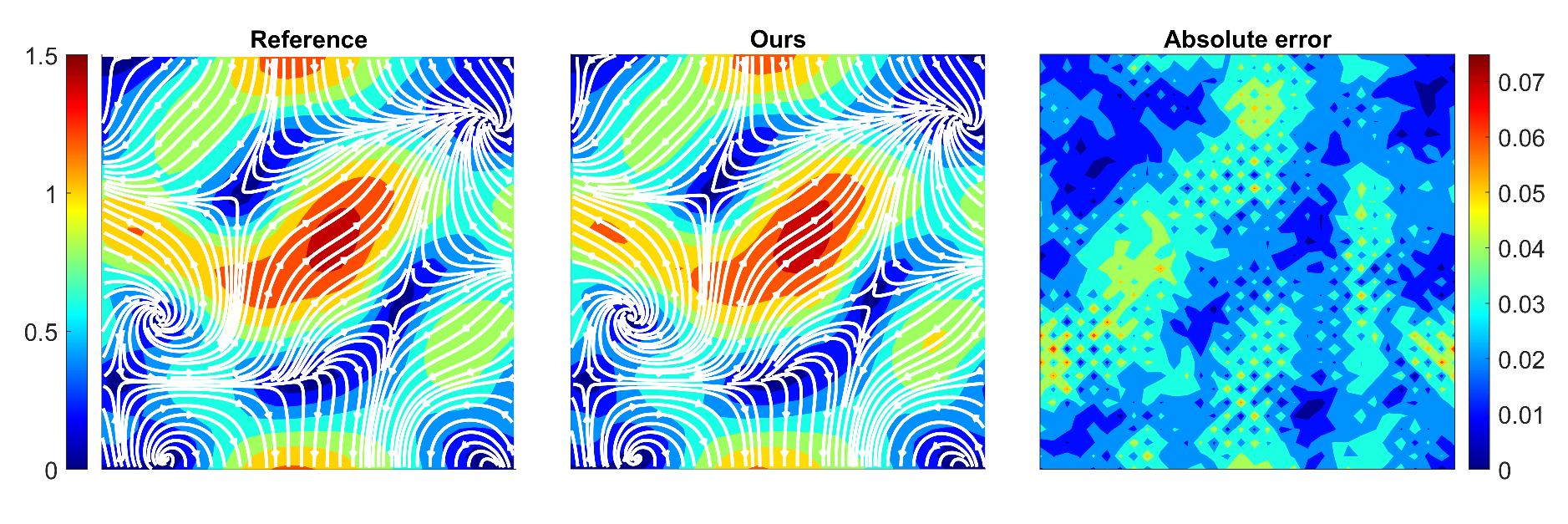}\\
 \includegraphics[width=\linewidth]{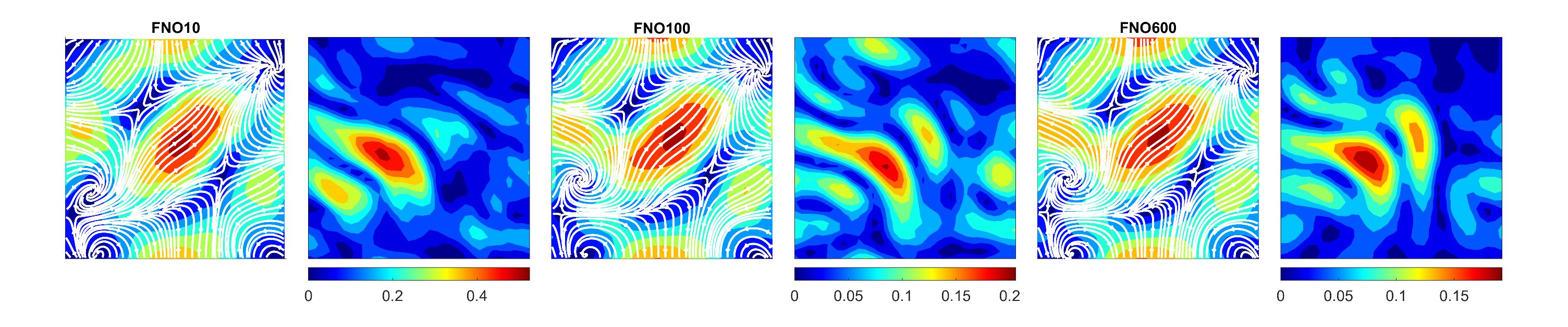}\\
 \includegraphics[width=\linewidth]{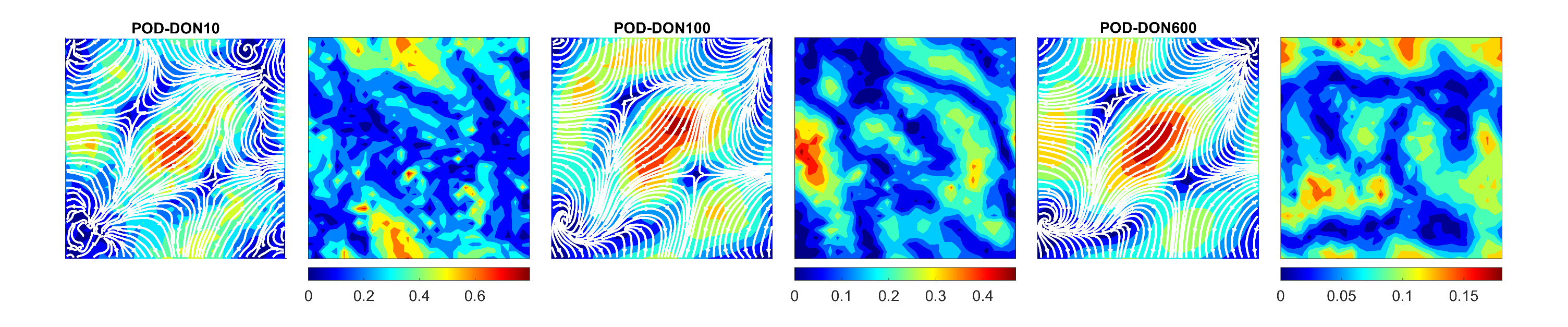}\\ 
 \includegraphics[width=\linewidth]{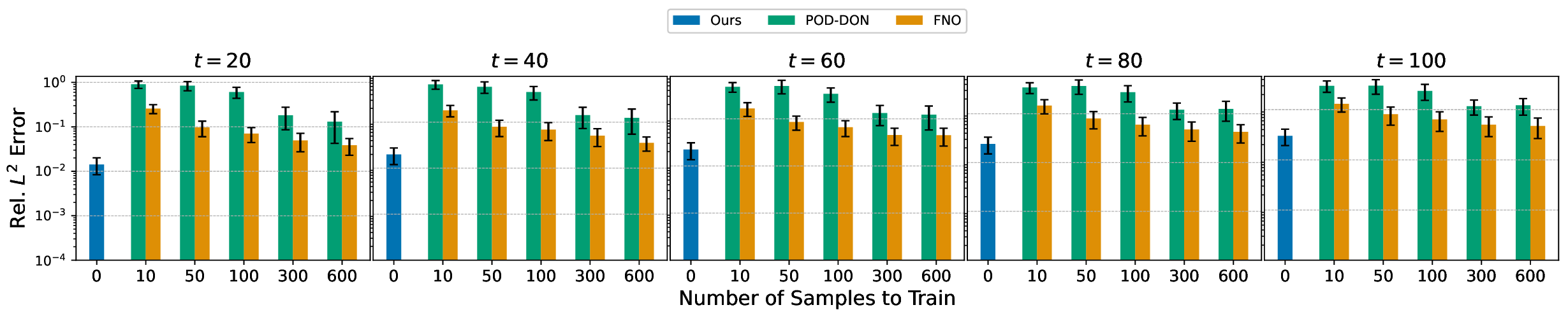} 
 \end{center}
 \caption{\textbf{Experiments for 2D initial conditions randomly generated by \eqref{generation:2dinitial} from $N(0,13^2)$.} \textbf{(Top)} Magnitude of a reference solution with its stream line at $T=1$ in the left panel,  magnitude of the corresponding inference of SpecONet with its stream line in the middle panel, and magnitude of the pointwise error between them in the right panel.\textbf{(Middle)} The Inferences and errors of FNO and POD-DON against the same input. Note that FNO100 and POD-DON100 both used 100 reference solutions to train, and FNO600 and POD-DON600 both did 600 ones. \textbf{(Bottom)} The Rel.$L^2_x$ errors are displayed for different networks, varying numbers of reference solutions to train, and at each time point (see table \ref{tab:comparison_2dinitial_sigma13}).}\label{fig:comparison_2dinitial_sigma13}
 \end{figure}

\newpage
\subsection{Two dimensional NSE with random boundary conditions}\label{sub:2d_boundary}We carried out experiments to evaluate the accuracy of inference against random boundary conditions as input. The random boundary conditions of $u$ were generated by the real part of
\begin{align}\label{generation:2d_boundary}
0.015\sin(t)\sum_{k=0}^{9}c_{k}\exp(ikx)~\text{on}~ y=1,
\end{align}
where $a_{k}$, and $b_{k}$ were random parameters to make  $c_k=a_{k}+ib_{k}$. We made 300 training samples whose $a_{k}$ and $b_k$ were drawn from $\mathcal{N}(0,5^2)$. On the other boundary, $u$ are set to zero. In addition, the boundary conditions of $v$ were set to zero. The other information is as in table \ref{tab:2d_boundary}.
\begin{table}[h!]
    \centering
    \begin{tabular}{|c|c||c|c|}
    \hline    
    Domain   & $[-1,1]^2$  & Boundary condition   & zero on $x=\pm 1$, $y=-1$ (Dirichlet condition).   \\ \hline
    Bases type   & Legendre   & Initial condition   &$u_0=v_0=0$   \\ \hline
    $\Delta t$   & 0.01   & The number of time steps& 100       \\ \hline
    $\nu$   & 0.5   & $N$   & 62    \\ \hline
    \end{tabular}
    \caption{Information on the numerical schemes for 2D NES with random boundary conditions.}
    \label{tab:2d_boundary}
\end{table}

In order for comparison, we carried out training on POD-DON and FNO with the forcing functions and the corresponding reference solutions at $T=0.2,0.4,0.6,0.8$ and $1$, which were identical to the boundary conditions used for training SpecONet. Note that the number of training samples for FNO and POD-DON varied from 10 to 300. These various training samples will underscore that accuracy of POD-DON and FNO is sensitive to the number of training samples whereas accuracy of SpecONet does not depend on the number of samples.

 For test samples, we created 100 boundary conditions from three cases, $\mathcal{N}(0,5^2)$, $\mathcal{N}(0,10^2)$, and $\mathcal{N}(0,20^2)$. Afterwards, velocity solutions to NSEs at $T=0.2,0.4,0.6,0.8$ and $1$ were computed to employ them as reference solutions. Then, 3 sets of 100 errors were computed in Rel.$L^2_x$ sense between inferences of each methods and the velocity reference solutions for each time step, $T=0.2,0.4,0.6,0.8$ and $1$. The comparison on the average of the errors made by each method are discussed in the following subsections: a) $\mathcal{N}(0,5^2)$ (see table \ref{tab:comparison_cavity_sigma5}, Fig.~\ref{fig:comparison_cavity_sigma5}); b) $\mathcal{N}(0,10^2)$ (see table \ref{tab:comparison_cavity_sigma10}; Fig.~\ref{fig:comparison_cavity_sigma10}); c) $\mathcal{N}(0,20^2)$ (see table \ref{tab:comparison_cavity_sigma20}; Fig.~\ref{fig:comparison_cavity_sigma20}).

\newpage
\subsubsection{\bf{Random boundary conditions on $\mathcal{N}(0,5^2)$}} As shown in table \ref{tab:comparison_cavity_sigma5}, the accuracy of SpecONet is close to that of the benchmarking
networks despite not depending on reference solutions. Particularly, the errors of
SpecONet outperform those of POD-DON and FNO for most of the cases. However, the errors of SpecONet are slightly over
those of FNO and POD-DON when training samples are 300 or $T\geq0.8$. That is because SpecONet is designed to emulate the time marching
numerical scheme, which is the same to error accumulation of the schemes as time
progresses. Meanwhile, as the number of samples decreases, the accuracy of POD-DNO and FNO worsens, whereas that of that of SpecONet does not.
\begin{table}[h!]
\centering
\begin{tabular}{|c|c||c||c|c|}
\hline
Time&  \makecell{SpecONet \\ (ours)} & \makecell{The number of \\training samples}  & POD-DON & FNO \\
\hline\hline
\multicolumn{1}{|c|}{\multirow{5}{*}{0.2}}&\multicolumn{1}{c||}{\multirow{5}{*}{7.83e-04}} & 10&3.69e-01 & 3.66e-01\\
 && 50&3.32e-02 & 4.53e-02\\
 && 100&2.50e-03 & 8.08e-03\\
 && 150&1.40e-03 & 5.33e-03\\
 && 300&9.03e-04 & 3.48e-03\\
\hline
\multicolumn{1}{|c|}{\multirow{5}{*}{0.4}}&\multicolumn{1}{c||}{\multirow{5}{*}{1.80e-03}} & 10&4.51e-01 & 4.77e-01\\
 && 50&6.63e-02 & 4.86e-02\\
 && 100&7.76e-03 & 9.70e-03\\
 && 200&2.30e-03 & 5.41e-03\\
 && 300&1.06e-03 & 3.32e-03\\
\hline
\multicolumn{1}{|c|}{\multirow{5}{*}{0.6}}&\multicolumn{1}{c||}{\multirow{5}{*}{3.02e-03}} & 10&5.27e-01 & 4.46e-01\\
 && 50&1.05e-01 & 5.55e-02\\
 && 100&1.52e-02 & 8.74e-03\\
 && 150&4.22e-03 & 5.43e-03\\
 && 300&1.39e-03 & 3.39e-03\\
\hline
\multicolumn{1}{|c|}{\multirow{5}{*}{0.8}}&\multicolumn{1}{c||}{\multirow{5}{*}{4.15e-03}} & 10&5.18e-01 & 4.55e-01\\
 && 50&1.86e-01 & 5.65e-02\\
 && 100&2.93e-02 & 9.25e-03\\
 && 150&6.70e-03 & 5.20e-03\\
 && 300&2.03e-03 & 3.25e-03\\
\hline
\multicolumn{1}{|c|}{\multirow{5}{*}{1.0}}&\multicolumn{1}{c||}{\multirow{5}{*}{6.48e-03}} & 10&5.02e-01 & 4.38e-01\\
 && 50&1.8e-01 & 4.90e-02\\
 && 100&4.53e-02 & 9.86e-03\\
 && 150&1.50e-02 & 5.21e-03\\
 && 300&3.31e-03 & 3.33e-03\\
\hline
\end{tabular}
\caption{Comparison on errors of various networks over boundary conditions generated in $\mathcal{N}(0,5^2)$. The errors are averaged over the Rel.$L^2_x$ errors of inferences from 100 unseen data samples at each time step: 0.2, 0.4, 0.6, 0.8, and 1. Note that whereas our method did not employ reference solutions to train, POD-DON and FNO employed 10, 50, 100, 150, or 300 references to train. 
}\label{tab:comparison_cavity_sigma5}
\end{table}
\newpage

 \begin{figure}[th!]
 \begin{center}
 {\textbf{Numerical results for $\sigma=5$, $T=1$}}
 \begin{tabular}{cll} 
 \includegraphics[width=0.3\linewidth]{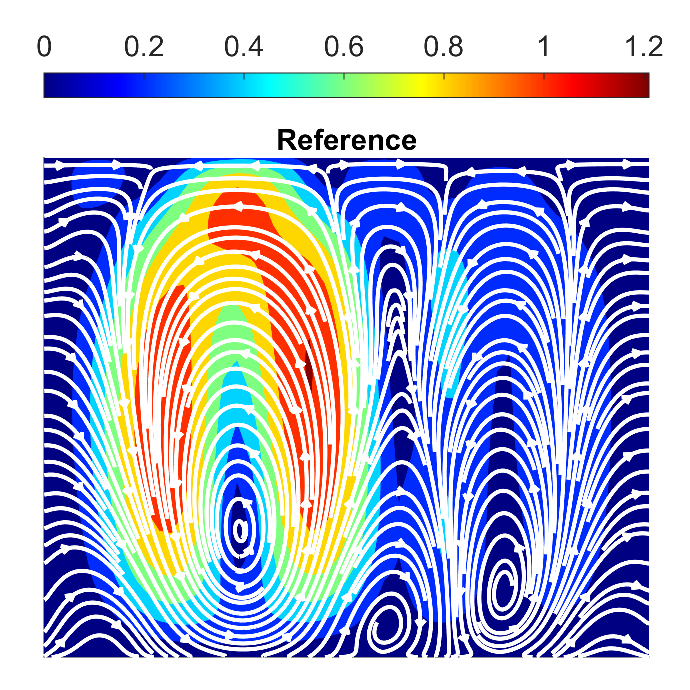}&\includegraphics[width=0.3\linewidth]{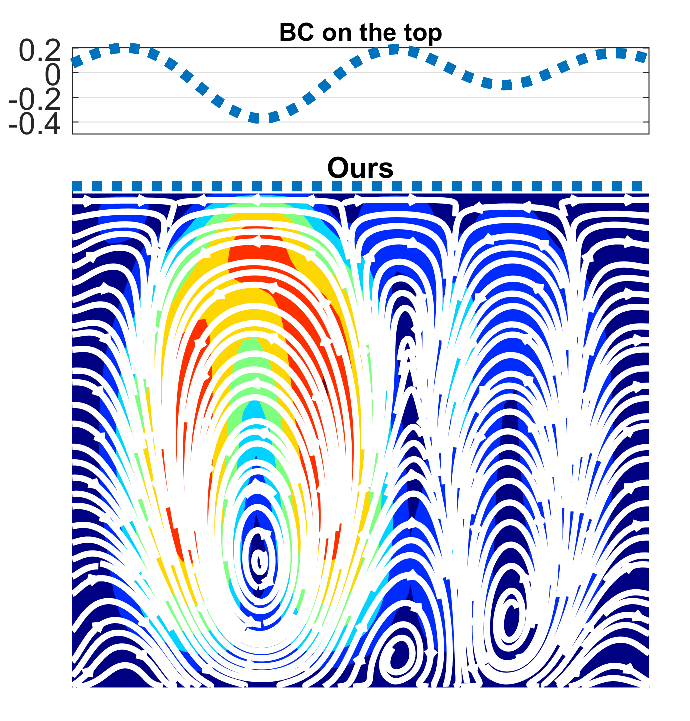}&\includegraphics[width=0.3\linewidth]{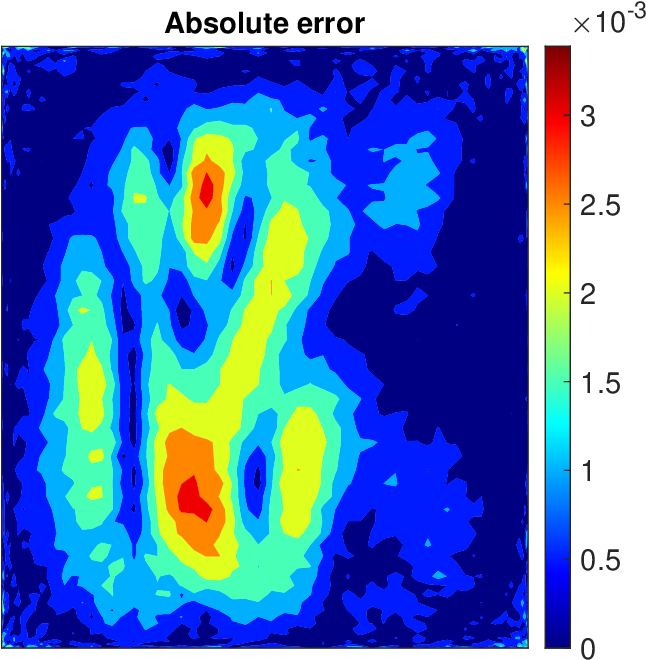}
      \end{tabular}
 \begin{tabular}{l} 
 \includegraphics[width=\linewidth]{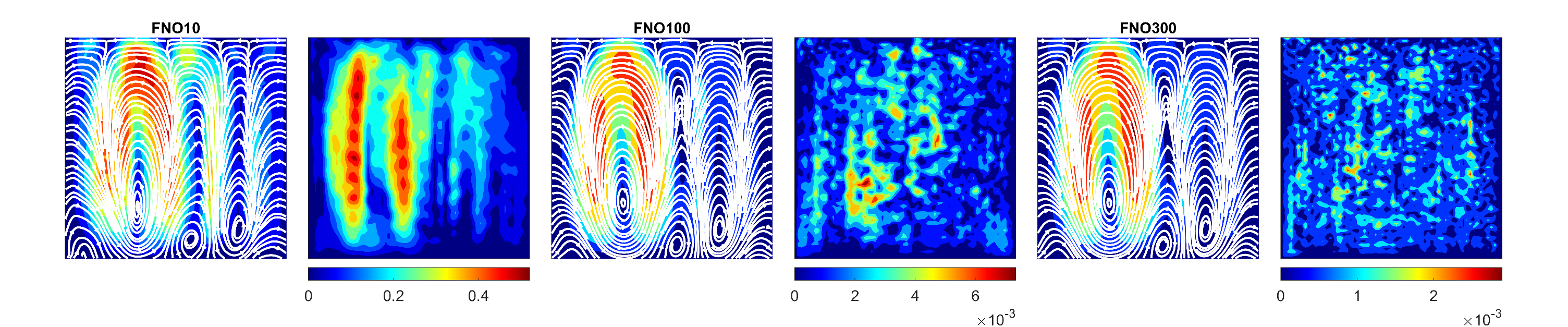}\\
 \includegraphics[width=\linewidth]{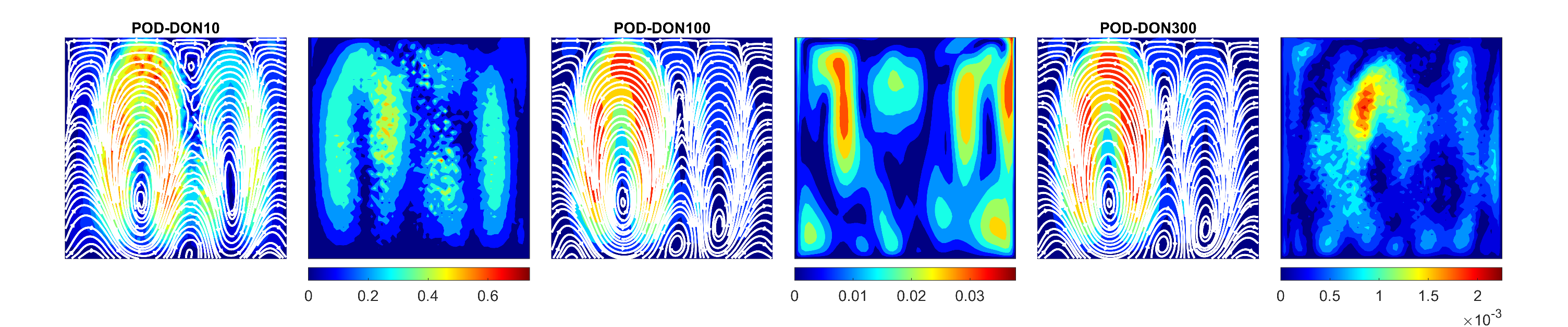}\\
     \includegraphics[width=\linewidth]{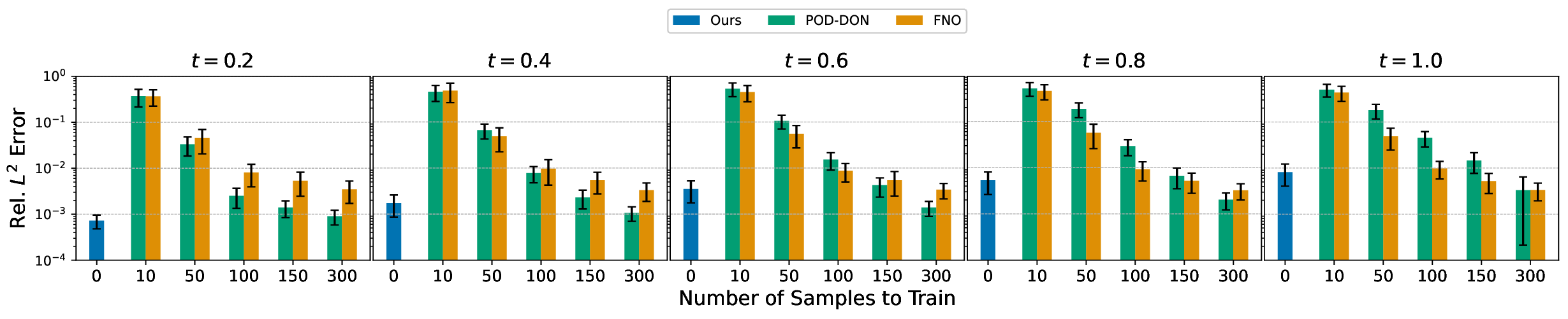}
      \end{tabular}
 \end{center}\caption{\textbf{Experiments for boundary conditions randomly generated by \eqref{generation:2d_boundary} from $N(0,5^2)$.} \textbf{(Top)} Magnitude of a reference solution with its stream line at $T=1$ in the left panel,  magnitude of the corresponding inference of SpecONet with its stream line in the middle panel, and magnitude of the pointwise error between them in the right panel.\textbf{(Middle)} The Inferences and errors of FNO and POD-DON against the same input. Note that FNO100 and POD-DON100 both used 100 reference solutions to train, and FNO300 and POD-DON300 both did 300 ones. \textbf{(Bottom)} The Rel.$L^2_x$ errors are displayed for different networks, varying numbers of reference solutions to train, and at each time point (see table\ref{tab:comparison_cavity_sigma5}).}\label{fig:comparison_cavity_sigma5}
 \end{figure}

\newpage
\subsubsection{\bf{Random boundary conditions on $\mathcal{N}(0,10^2)$}}This test data are sampled in a more complicated manner than the training data, which was sampled from $\mathcal{N}(0,5^2)$. In this case, the Rel.$L^2_x$ errors of SpecONet are comparable to FNO POD-DON without data reliance. Especially, SpecONet achieves better performance than the others except at 300 training samples. Only if at least 300 training samples are provided, the accuracy of the benchmarking networks are better than that of Speconet. Besides, we note that SpecONet exhibits a smaller error increase from the errors on $\mathcal{N}(0,5^2)$ compared to others, which implies that SpecONet is more robust for more complicated test samples than POD-DON and FNO. 
\begin{table}[h!]
\centering
\begin{tabular}{|c|c||c||c|c|}
\hline
Time&  \makecell{SpecONet \\ (ours)} & \makecell{The number of \\training samples}  & POD-DON & FNO \\
\hline\hline
\multicolumn{1}{|c|}{\multirow{5}{*}{0.2}}&\multicolumn{1}{c||}{\multirow{5}{*}{1.42e-03}} & 10&3.91e-01 & 4.23e-01\\
 && 50&6.07e-02 & 1.03e-01\\
 && 100&5.98e-03 & 2.66e-02\\
 && 150&3.26e-03 & 1.86e-02\\
 && 300&1.93e-03 & 1.21e-02\\
\hline
\multicolumn{1}{|c|}{\multirow{5}{*}{0.4}}&\multicolumn{1}{c||}{\multirow{5}{*}{3.73e-03}} & 10&4.86e-01 & 4.93e-01\\
 && 50&1.22e-01 & 1.14e-01\\
 && 100&2.29e-02 & 3.22e-02\\
 && 150&7.44e-03 & 1.91e-02\\
 && 300&3.34e-03 & 1.01e-02\\
\hline
\multicolumn{1}{|c|}{\multirow{5}{*}{0.6}}&\multicolumn{1}{c||}{\multirow{5}{*}{6.81e-03}} & 10&5.53e-01 & 4.96e-01\\
 && 50&1.84e-01 & 1.28e-01\\
 && 100&4.21e-02 & 2.97e-02\\
 && 150&1.43e-02 & 2.06e-02\\
 && 300&5.25e-03 & 1.04e-02\\
\hline
\multicolumn{1}{|c|}{\multirow{5}{*}{0.8}}&\multicolumn{1}{c||}{\multirow{5}{*}{1.10e-02}} & 10&5.35e-01 & 5.06e-01\\
 && 50&2.92e-01 & 1.35e-01\\
 && 100&7.06e-02 & 3.25e-02\\
 && 150&5.20e-02 & 1.84e-02\\
 && 300&8.56e-03 & 1.08e-02\\
\hline
\multicolumn{1}{|c|}{\multirow{5}{*}{1.0}}&\multicolumn{1}{c||}{\multirow{5}{*}{2.17e-02}} & 10&5.24e-01 & 5.14e-01\\
 && 50&2.74e-01 & 1.25e-01\\
 && 100&1.09e-01 & 3.51e-02\\
 && 150&2.75e-02 & 2.00e-02\\
 && 300&1.63e-02 & 1.24e-02\\
\hline
\end{tabular}
\caption{Comparison on errors of various networks over boundary conditions generated in $\mathcal{N}(0,10^2)$. The errors are averaged over the Rel.$L^2_x$ errors of inferences from 100 unseen data samples at each time step: 0.2, 0.4, 0.6, 0.8, and 1. Note that whereas our method did not employ reference solutions to train, POD-DON and FNO employed 10, 50, 100, 150, or 300 references to train. 
}\label{tab:comparison_cavity_sigma10}
\end{table}

\newpage
 \begin{figure}[th!]
 \begin{center}
 {\textbf{Numerical results for $\sigma=10$, $T=1$}}
 \begin{tabular}{cll} 
 \includegraphics[width=0.3\linewidth]{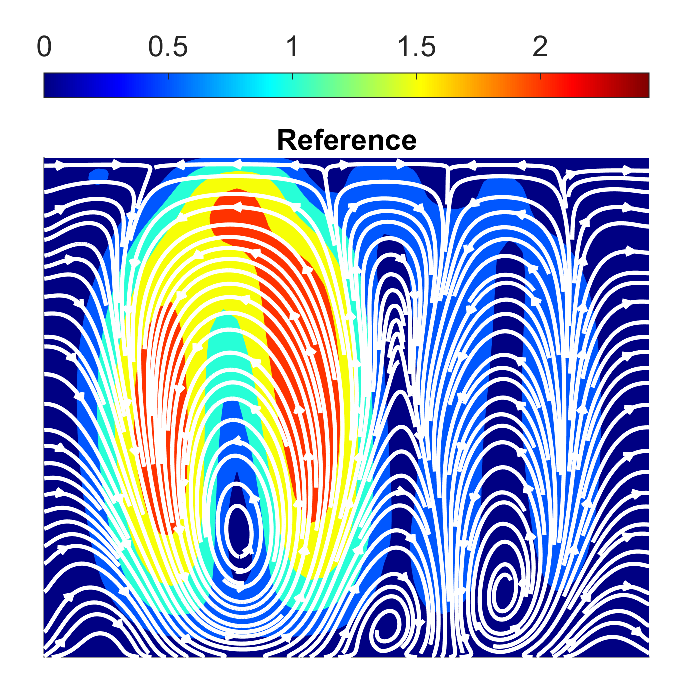}&\includegraphics[width=0.3\linewidth]{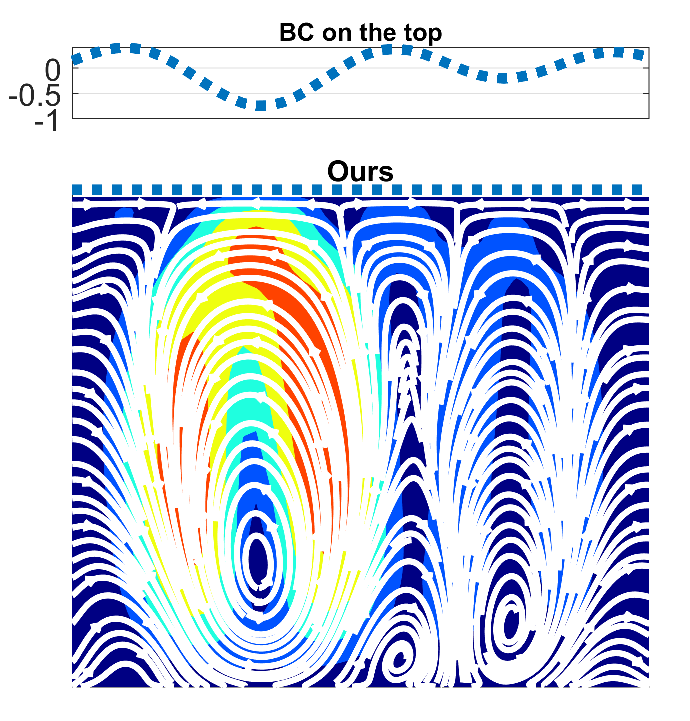}&\includegraphics[width=0.3\linewidth]{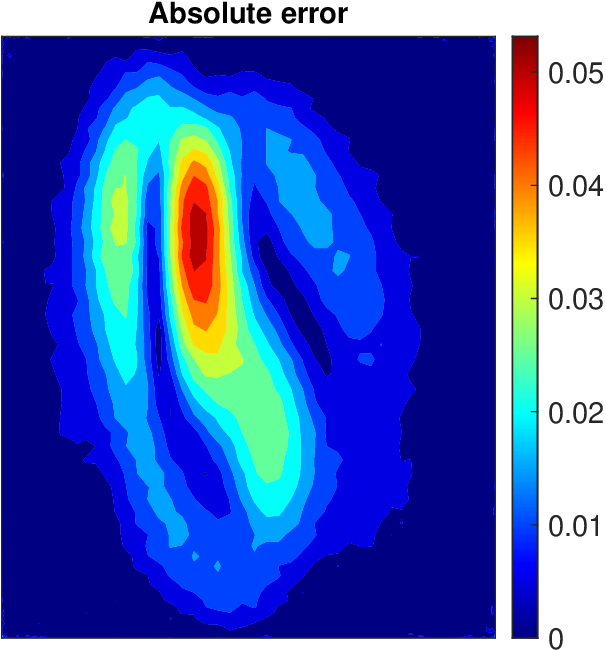}
      \end{tabular}
 \includegraphics[width=\linewidth]{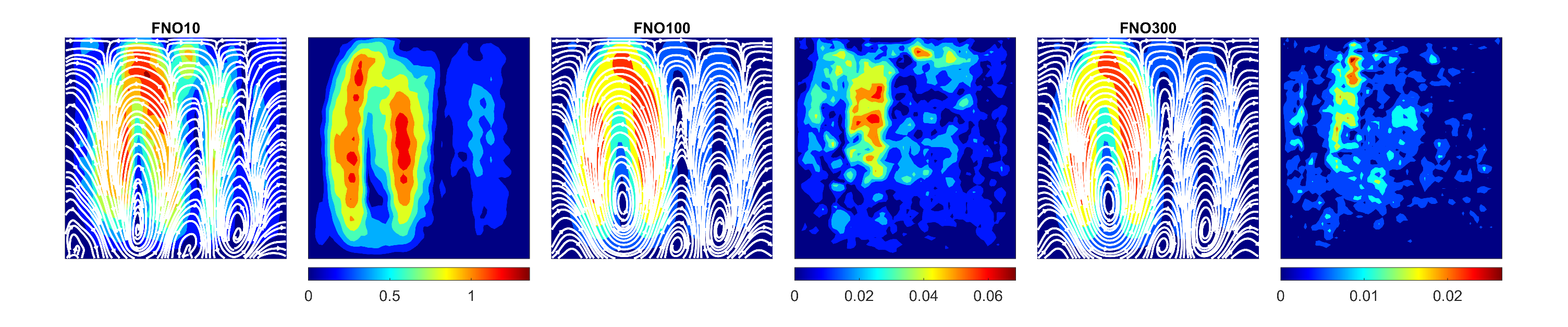}\\
 \includegraphics[width=\linewidth]{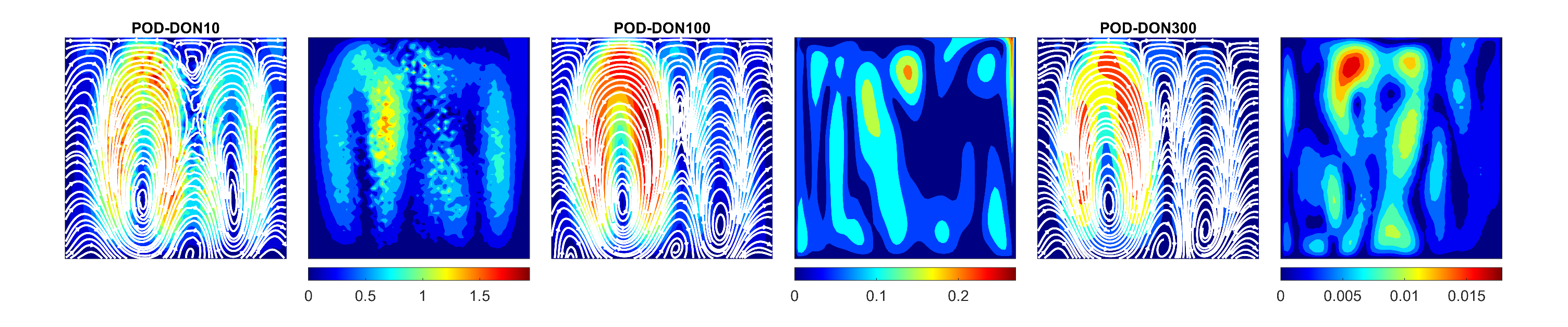}\\      
 \includegraphics[width=\linewidth]{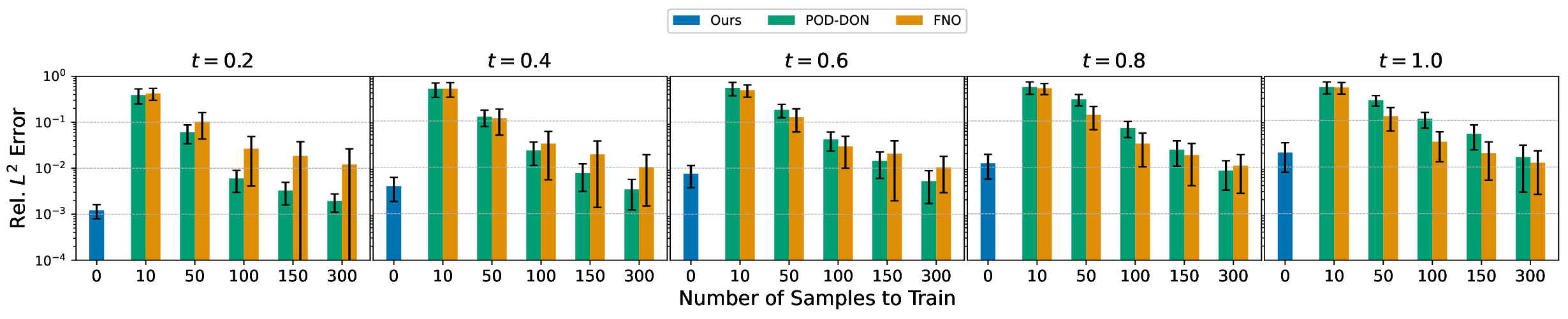}
    
\end{center}
 \caption{\textbf{Experiments for boundary conditions randomly generated by \eqref{generation:2d_boundary} from $N(0,10^2)$.} \textbf{(Top)} Magnitude of a reference solution with its stream line at $T=1$ in the left panel,  magnitude of the corresponding inference of SpecONet with its stream line in the middle panel, and magnitude of the pointwise error between them in the right panel.\textbf{(Middle)} The Inferences and errors of FNO and POD-DON against the same input. Note that FNO100 and POD-DON100 both used 100 reference solutions to train, and FNO300 and POD-DON300 both did 2000 ones. \textbf{(Bottom)} The Rel.$L^2_x$ errors are displayed for different networks, varying numbers of reference solutions to train, and at each time point (see table\ref{tab:comparison_cavity_sigma10}).}\label{fig:comparison_cavity_sigma10}
 \end{figure}

\newpage
\subsubsection{\bf{Random boundary conditions on $\mathcal{N}(0,20^2)$}}This test data are sampled in a far more complicated manner than the training data, which was sampled from $\mathcal{N}(0,5^2)$. In this case, the Rel.$L^2_x$ errors of SpecONet are smaller than FNO POD-DON unless the number of training samples is 300. Only if the number of training samples is more than 300, the errors of FNO and POD-DON falls below those of SpecONet. In addition, comparing to the errors on $\mathcal{N}(0,5^2)$, the errors of SpecOnet grow less rapidly than those of the others. This implies that SpecONet is more reliable for more complicated test samples than POD-DON and FNO.

\begin{table}[h!]
\centering
\begin{tabular}{|c|c||c||c|c|}
\hline
Time&  \makecell{SpecONet \\ (ours)} & \makecell{The number of \\training samples}  & POD-DON & FNO \\
\hline\hline
\multicolumn{1}{|c|}{\multirow{5}{*}{0.2}}&\multicolumn{1}{c||}{\multirow{5}{*}{3.35e-03}} & 10&4.82e-01 & 4.96e-01\\
 && 50&1.34e-01 & 1.90e-01\\
 && 100&1.98e-02 & 8.53e-02\\
 && 150&1.11e-03 & 6.57e-02\\
 && 300&6.72e-03 & 5.20e-02\\
\hline
\multicolumn{1}{|c|}{\multirow{5}{*}{0.4}}&\multicolumn{1}{c||}{\multirow{5}{*}{9.30e-03}} & 10&5.56e-01 & 5.34e-01\\
 && 50&2.34e-01 & 2.16e-01\\
 && 100&6.66e-02 & 9.93e-02\\
 && 150&2.72e-02 & 6.85e-02\\
 && 300&1.46e-02 & 4.1e-02\\
\hline
\multicolumn{1}{|c|}{\multirow{5}{*}{0.6}}&\multicolumn{1}{c||}{\multirow{5}{*}{2.01e-02}} & 10&6.02e-01 & 5.66e-01\\
 && 50&2.87e-01 & 2.42e-01\\
 && 100&9.76e-02 & 9.91e-02\\
 && 150&4.92e-02 & 7.90e-02\\
 && 300&2.39e-02 & 4.57e-02\\
\hline
\multicolumn{1}{|c|}{\multirow{5}{*}{0.8}}&\multicolumn{1}{c||}{\multirow{5}{*}{3.98e-02}} & 10&5.63e-01 & 5.70e-01\\
 && 50&4.04e-01 & 2.61e-01\\
 && 100&1.44e-01 & 1.15e-01\\
 && 150&7.58e-02 & 7.44e-02\\
 && 300&3.89e-02 & 5.16e-02\\
\hline
\multicolumn{1}{|c|}{\multirow{5}{*}{1.0}}&\multicolumn{1}{c||}{\multirow{5}{*}{8.77e-02}} & 10&5.64e-01 & 6.00e-01\\
 && 50&3.79e-01 & 2.66e-01\\
 && 100&2.25e-01 & 1.28e-01\\
 && 150&1.43e-01 & 5.16e-02\\
 && 300&7.04e-02 & 6.43e-02\\
\hline
\end{tabular}
\caption{Comparison on errors of various networks over boundary conditions generated in $\mathcal{N}(0,20^2)$. The errors are averaged over the Rel.$L^2_x$ errors of inferences from 100 unseen data samples at each time step: 0.2, 0.4, 0.6, 0.8, and 1. Note that whereas our method did not employ reference solutions to train, POD-DON and FNO employed 10, 50, 100, 150, or 300 references to train. 
}\label{tab:comparison_cavity_sigma20}
\end{table}

\newpage
  \begin{figure}[th!]
 \begin{center}
 {\textbf{Numerical results for $\sigma=20$, $T=1$}}
 \begin{tabular}{cll} 
 \includegraphics[width=0.3\linewidth]{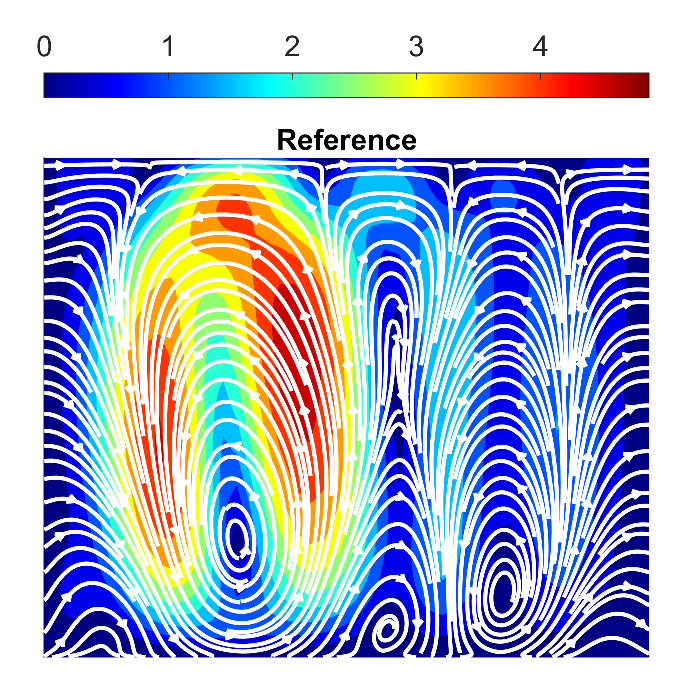}&\includegraphics[width=0.3\linewidth]{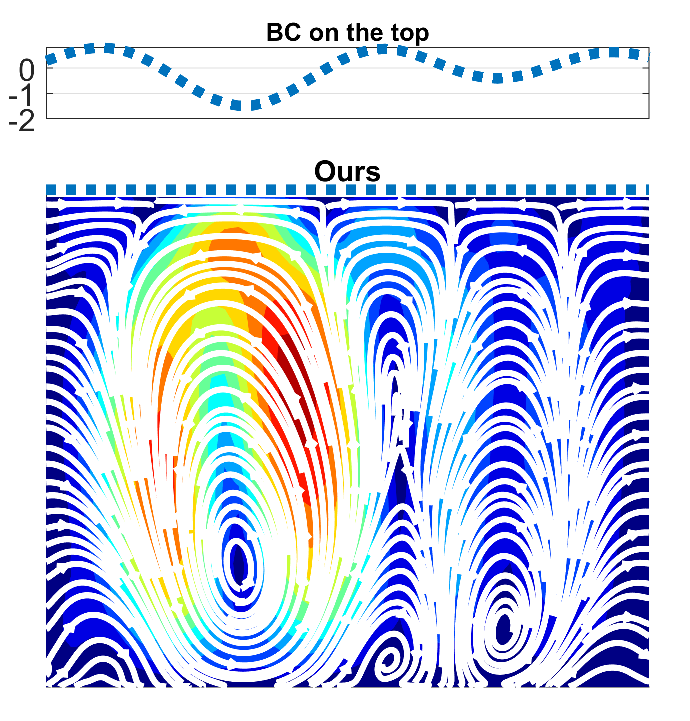}&\includegraphics[width=0.3\linewidth]{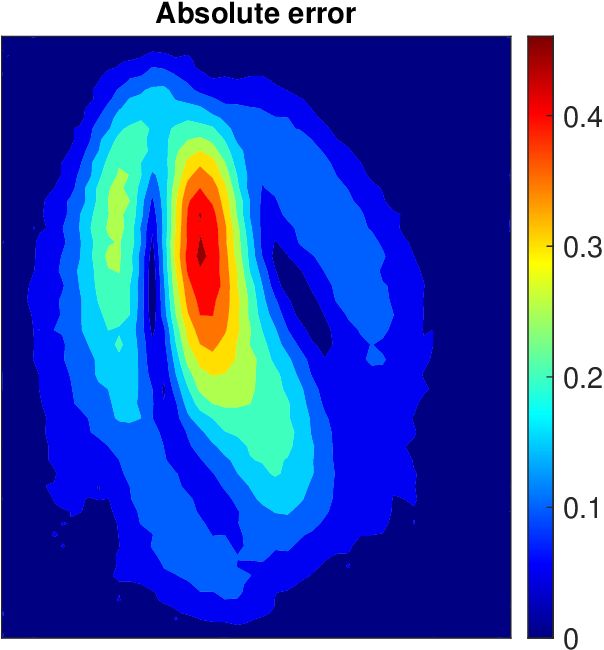}
      \end{tabular}
 \begin{tabular}{l} 
  \includegraphics[width=\linewidth]{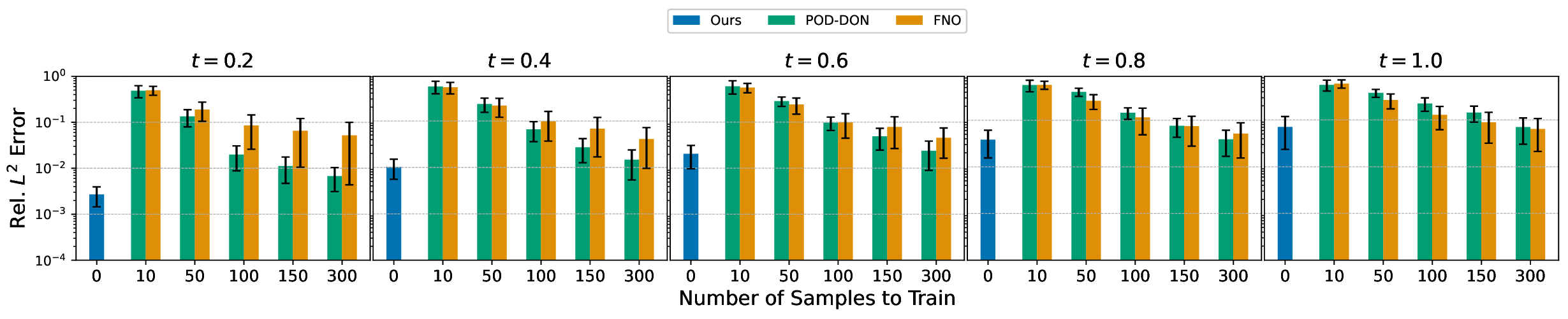}
      \end{tabular}
 \end{center}\caption{\textbf{Experiments for boundary conditions randomly generated by \eqref{generation:2d_boundary}  from $N(0,20^2)$.} \textbf{(Top)} Magnitude of a reference solution with its stream line at $T=1$ in the left panel,  magnitude of the corresponding inference of SpecONet with its stream line in the middle panel, and magnitude of the pointwise error between them in the right panel.\textbf{(Middle)} The Inferences and errors of FNO and POD-DON against the same input. Note that FNO100 and POD-DON100 both used 100 reference solutions to train, and FNO300 and POD-DON300 both did 300 ones. \textbf{(Bottom)} the Rel.$L^2_x$ errors are displayed for different networks, varying numbers of reference solutions to train, and at each time point (see table \ref{tab:comparison_cavity_sigma20}).}\label{fig:comparison_cavity_sigma20}
 \end{figure}

\newpage
\subsection{Three dimensional Beltrami flow with random initial conditions}\label{sub:3d_initial}
Beltrami flow is well known as flows in which the velocity is parallel to its vorticity. Thanks to do it, it has exact solution forms. Accordingly, it is usually utilized to test whether numerical schemes for NSE converge to its solutions. Based on the background, we tested our network to infer Beltrami flows when initial conditions were provided as input. In addition, a periodic boundary condition was imposed on a computational domain $[0,2\pi]^3$. Then, one can construct Beltrami flow solutions (cf. \cite{antuono2020tri}) as 

{\small
\begin{align}\label{Belt_sol}
 u&=A((a\cos(kx)+b\sin(kx))(-c\sin(ky)+d\cos(k y))(e\cos(kz)+f\sin(kz)) \\
 &-(-a\sin(kz)+b\cos(kz))(c\cos(kx)+d\sin(kx))(e\cos(ky)+f\sin(ky)))\exp(-3\nu k^2t)\nonumber\\
  v&=A((a\cos(ky)+b\sin(ky))(-c\sin(kz)+d\cos(kz))(e\cos(kx)+f\sin(kx)) \\
 &-(-a\sin(kx)+b\cos(kx))(c\cos(ky)+d\sin(ky))(e\cos(kz)+f\sin(kz)))\exp(-3\nu k^2t)\nonumber\\
  w&=A((a\cos(kz)+b\sin(kz))(-c\sin(kx)+d\cos(kx))(e\cos(ky)+f\sin(ky)) \\
 &-(-a\sin(ky)+b\cos(ky))(c\cos(kz)+d\sin(kz))(e\cos(kx)+f\sin(kx)))\exp(-3\nu k^2t)\nonumber\\
 p&=p_0-\frac{u^2+v^2+w^2}{2}
\end{align}}
where $r$, $C_4$, $C_5$, $C_6$, and $p_0$ are real constants and $k$ is an integer to compute 
\begin{align*}
C_1&=(\sqrt{3}-r)(\sqrt{3}r-1),~C_2=(\sqrt{3}+r)(\sqrt{3}r+1),~C_3=3r^2-1,\\
a&=C_1C_4,~b=C_3C_4,~c=C_2C_5,~d=C_3C_5,~e=C_3,~f=rC_3.
\end{align*}

In this experiment, we generated 600 sets of solutions for training samples as follows. First, we randomly drew $k$ from $\{1,2,3\}$. After that, we chose $C_4, C_5, C_6$ satisfying the following conditions: 1) they are random variables drawn from $\mathcal{N}(60,10^2)$; and 2) they are greater than 60 or less than $-60$. Note that the second condition prevents variance of $p$ from widening too much, which would cause the error of $p$ to get larger while training SpecONet. Accordingly, we denote the distribution satisfying these two conditions by $\mathcal{N}^{\ast}(60,10^2)$. Besides, we set $A=2\times 10^{-6}$, $\nu=0.1$, and $r,t=0$. The other information is as in table \ref{tab:3d_Beltrami}.
\begin{table}[h!]
    \centering    
    \begin{tabular}{|c|c||c|c|}
    \hline    
    Domain   & $[0,2\pi]^3$  & Boundary condition   & periodic   \\ \hline
    Bases type   & Fourier   & Forcing function   &$f_x=f_y=f_z=0$    \\ \hline
    $\Delta t$   & 0.01   & The number of time steps& 100       \\ \hline
    $\nu$   & 0.1   & $N$   & 24    \\ \hline
    \end{tabular}
    \caption{Information on the numerical schemes for 3D Beltarmi flows}
    \label{tab:3d_Beltrami}
\end{table}

For test samples, we made 100 forcing functions from four distribution, $\mathcal{N}(60,5^2)$, $\mathcal{N}^\ast(60,10^2)$, $\mathcal{N}(60,10^2)$, and $\mathcal{N}(60,20^2)$. Note that $\mathcal{N}^\ast(60,10^2)$ means the same distribution as the one used to produce the training samples. However, all the test samples were new unseen data from the training samples. Afterwards, velocity solutions to NSEs at $t\in[0,1]$ as in \eqref{Belt_sol} were computed to employ them as reference solutions. Then, four sets of 100 errors were computed in Rel.$L^2_{t,x}$ sense between inferences of each methods and the velocity reference solutions for $t\in[0,1]$. The comparison on the errors varying the distributions are displayed in table \ref{tab:error_all3dbelt}, and Fig.~\ref{fig:error3dbelt}).

\begin{table}[h!]
\centering
\begin{tabular}{|c|c|c|c|c|}
\hline
\makecell{Types of input}   & $\mathcal{N}(60,5^2)$&$\mathcal{N}^{\ast}(60,10^2)$&$\mathcal{N}(60,10^2)$&$\mathcal{N}(60,20^2)$ \\
\hline\hline
Rel.$L_{t,x}^2$ error of $u$ &9.89e-05 &6.82e-05 &1.16e-04 &5.61e-04  \\
\hline
Rel.$L_{t,x}^2$ error of $v$ &9.85e-05 &6.81e-05 &1.16e-04 &5.60e-04 \\
\hline
Rel.$L_{t,x}^2$ error of $w$ &9.90e-05& 6.82e-05 &1.16e-04 &5.63e-04 \\
\hline
Rel.$H_{t,x}^1$ error of $p$ &3.27e-01 &1.76e-02 &6.67e-01 &1.35e+01  \\
\hline
\end{tabular}
\caption{the Rel.$L^2_{t,x}$ error of 3D initial conditions.  }\label{tab:error_all3dbelt}
\end{table}

The errors for the four cases are exhibited in table and fig. Regarding the velocity filed, the errors are under 0.06\% in $L^2_{t,x}$ sense. However, the errors of pressure are relatively large in $H^1_{t,x}$. The large error in pressure is attributed to are three factors. The first factor relates to the input data, $\nabla\cdot \mathbf{\tilde{u}}$ of $\mathcal{G}_{\Phi}^\theta$, whose distribution is not as well-ordered as a normal distribution. The second factor involves the solution $p$. Because the pressure amplitude is of the order of $C_4$, $C_5$, $C_6$' square, the variance of $p$ is significantly wider than that of the velocities. These might cause the distribution of the corresponding  inferences of $\mathcal{G}_\Phi^\theta$ to be irregular. Lastly, Given these two factors, the global error of p accumulates as time progresses, since p is updated from its value at the previous time step. In contrast, The velocity field is not updated from previous values, so the error does not accumulate in the same way.

Fig.~\ref{fig:energy3dbelt} exhibits the average of the energy and enstrophy over 100 test samples for each distribution. Note that the energy is computed based on the exact solutions. Because the inferences are quite close to the corresponding exact solutions, their energy and the enstrophy evolutions also behave quite closely.

Fig.~\ref{fig:3dbelt_time} displays an example of solutions, comparing an inference to the exact solutions as time progresses.

Fig.~\ref{fig:3dbelt_sigma} describes an example of solutions, comparing an inference to the exact solutions as the normal distribution varies.

\newpage

 \begin{figure}[th!]
 \begin{center}
 \begin{tabular}{cccc} 
 {$u$}&{$v$}&{$w$}&{$\nabla p$}\\
 \includegraphics[width=0.2\linewidth]{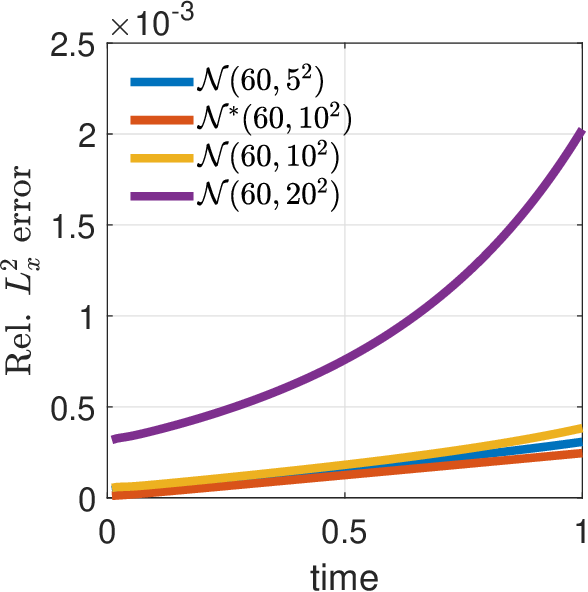}&\includegraphics[width=0.2\linewidth]{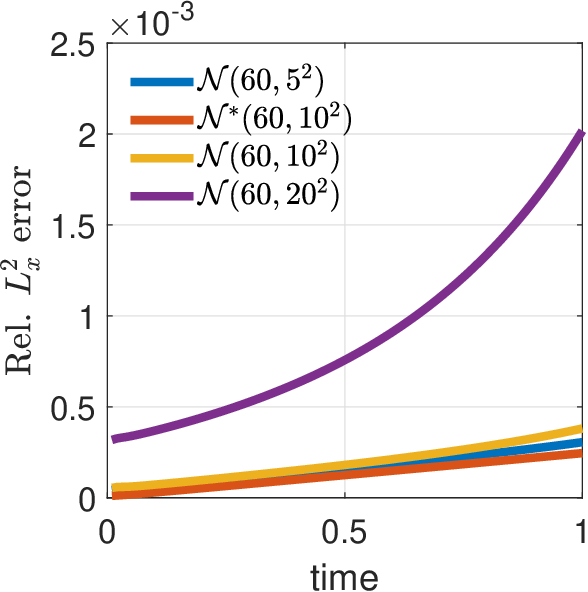}&\includegraphics[width=0.2\linewidth]{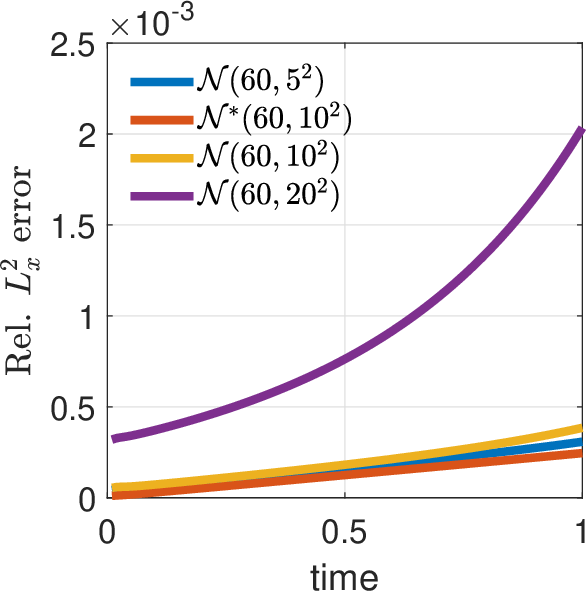}&\includegraphics[width=0.2\linewidth]{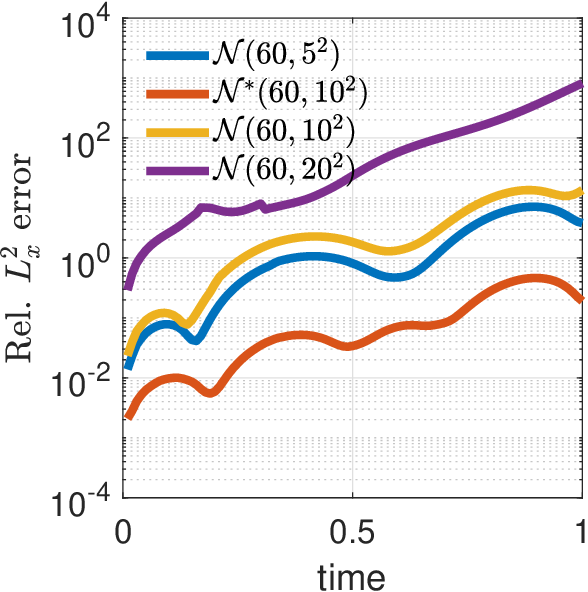}
      \end{tabular}
 \end{center}
 \caption{\textbf{Rel. $L_x^2$ error profile of 3D initial conditions over time}}\label{fig:error3dbelt}
 \end{figure}

\begin{figure}[hbp!]
 \begin{center}
 \begin{tabular}{cccc} 
 {$\mathcal{N}(60,5^2)$}&{$\mathcal{N}^{\ast}(60,10^2)$}&{$\mathcal{N}(60,10^2)$}&{$\mathcal{N}(60,20^2)$}\\
 \includegraphics[width=0.2\linewidth]{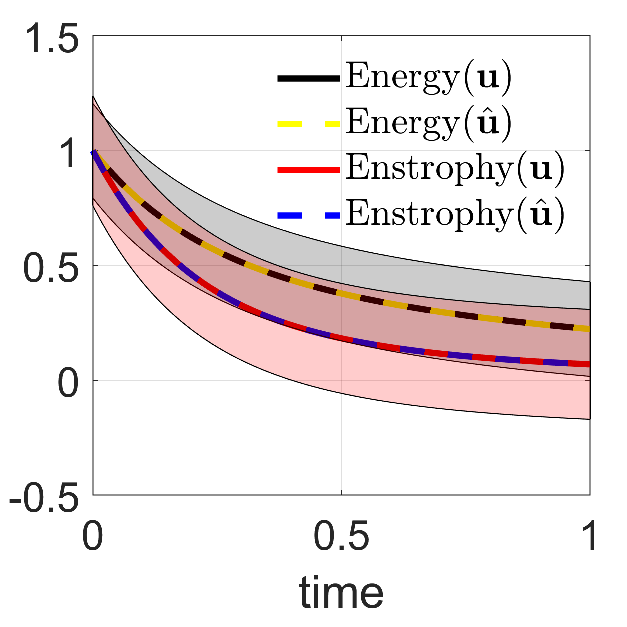}&\includegraphics[width=0.2\linewidth]{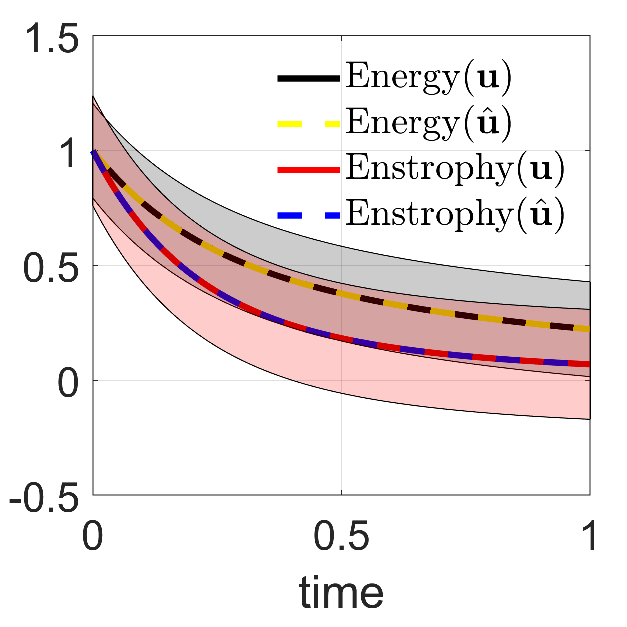}&\includegraphics[width=0.2\linewidth]{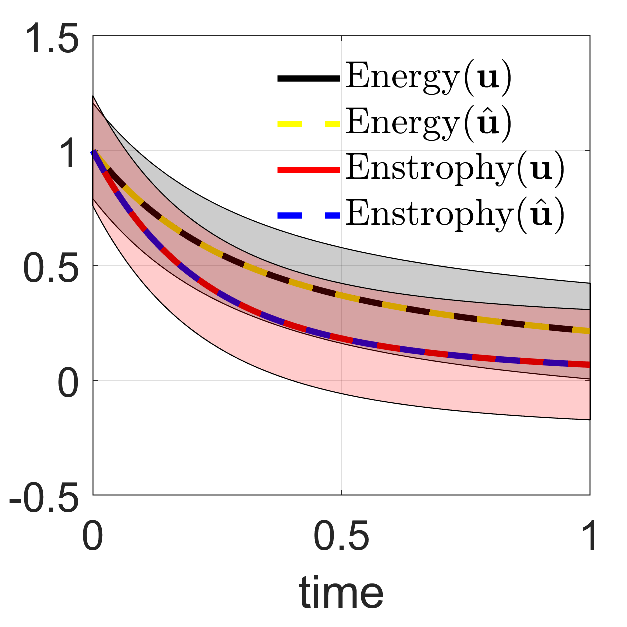}&\includegraphics[width=0.2\linewidth]{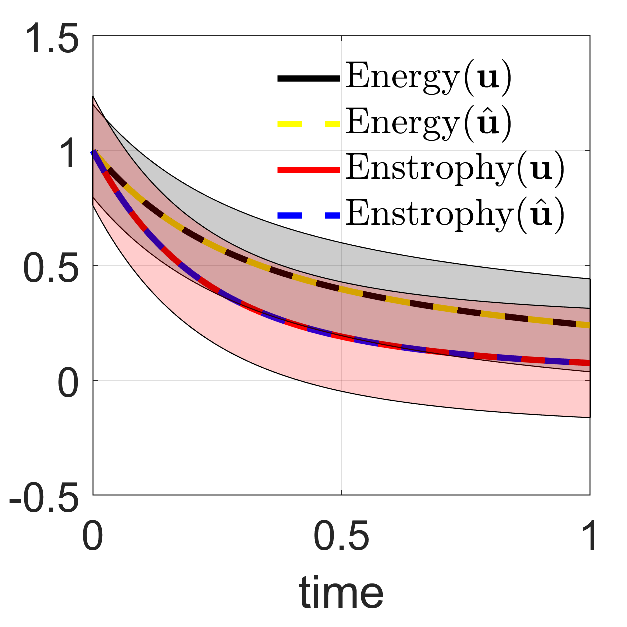}
      \end{tabular}
 \end{center}
 \caption{\textbf{Energy and Enstrophy evolution over time.} The solid and dashed lines indicate the average over the 100 quantities in the legend, while the shading shows the standard deviation range around them. }\label{fig:energy3dbelt}
 \end{figure}

\newpage
 \begin{figure}[th!]
 \begin{center}
 {\textbf{Numerical results for $\mathcal{N}(60,10^2)$}}\\{\textbf{}}\\
 \begin{tabular}{p{1em}cccc} 
 &{t=0.25}&{t=0.5}&{t=0.75}&{t=1}\\
  \begin{turn}{90}{\qquad\quad Inference}\end{turn}&
 \includegraphics[width=0.2\linewidth]{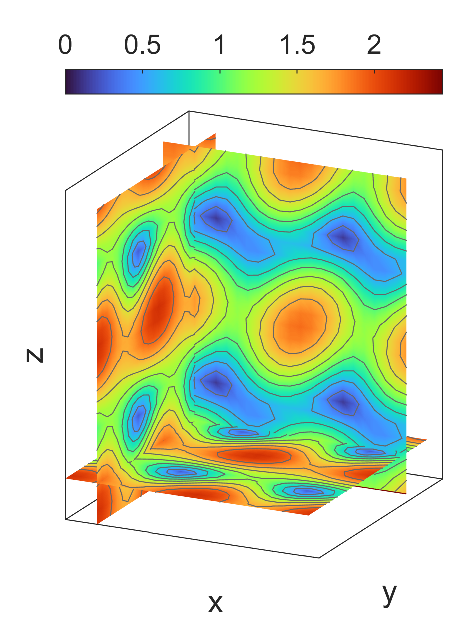}&\includegraphics[width=0.2\linewidth]{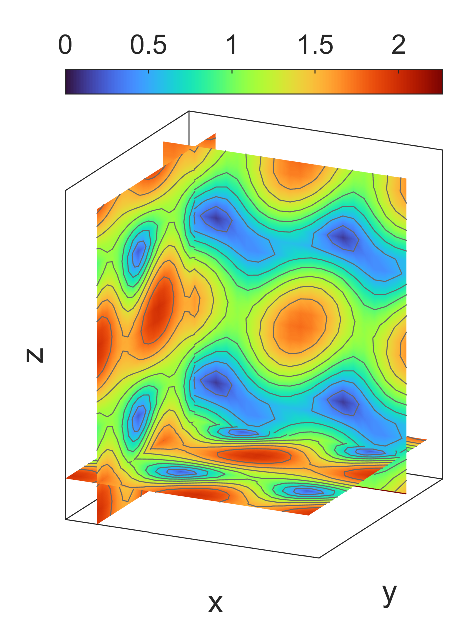}&\includegraphics[width=0.2\linewidth]{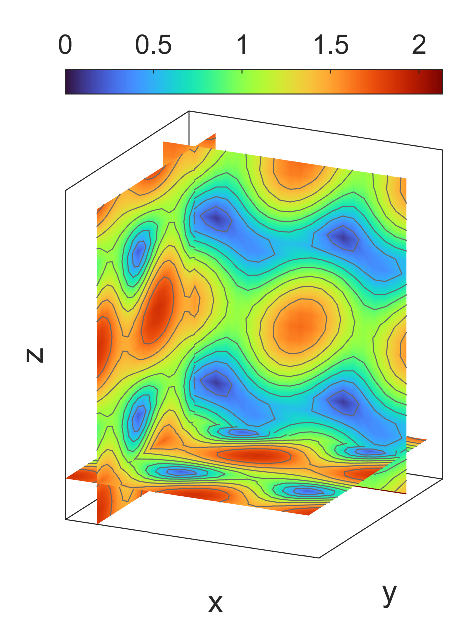}&\includegraphics[width=0.2\linewidth]{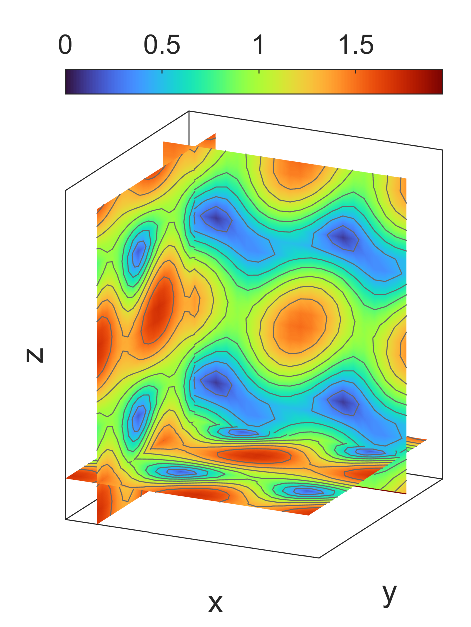}\\
 \begin{turn}{90}{\qquad\quad Pointwise error}\end{turn}&
 \includegraphics[width=0.2\linewidth]{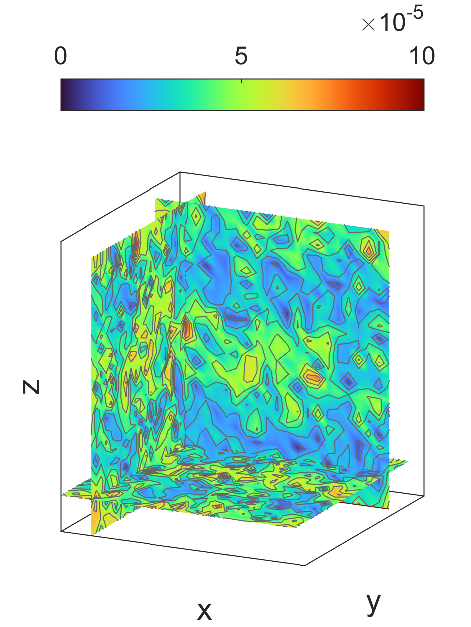}&\includegraphics[width=0.2\linewidth]{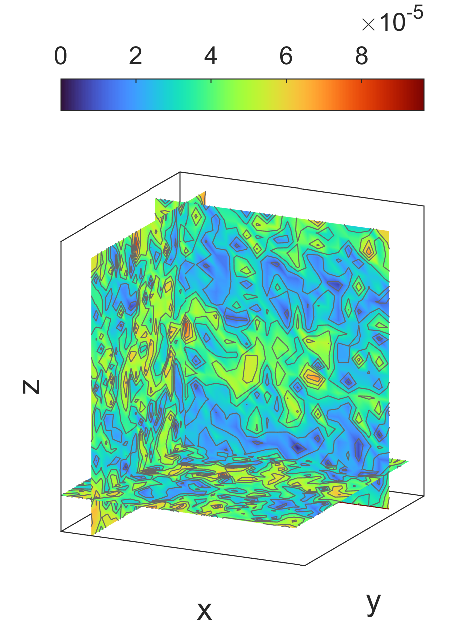}&\includegraphics[width=0.2\linewidth]{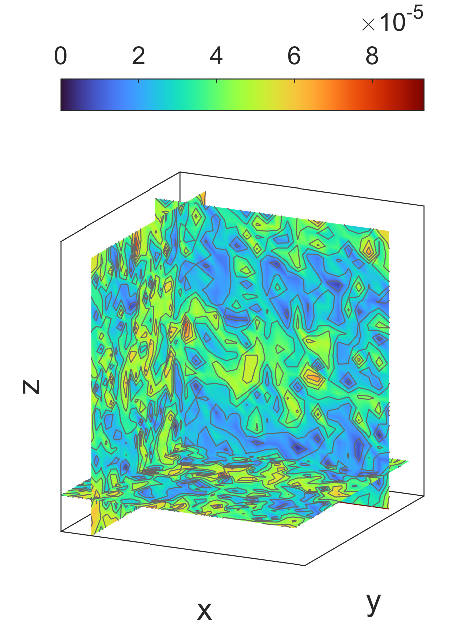}&\includegraphics[width=0.2\linewidth]{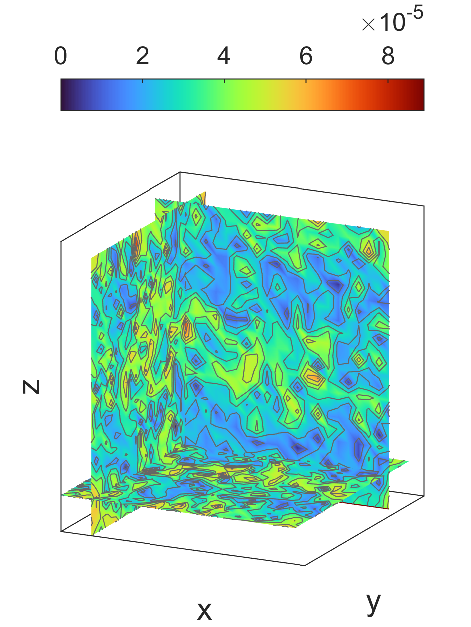}
      \end{tabular}  
 \end{center}
 \caption{\textbf{Experiments for Beltrami flow with 3D initial conditions randomly generated by \eqref{Belt_sol} on $N(0,10^2)$.}  }\label{fig:3dbelt_time}
 \end{figure}

\newpage
 \begin{figure}[th!]
 \begin{center}
 {\textbf{Numerical results at $T=1$}}\\{\textbf{}}\\
 \begin{tabular}{p{1em}cccc} 
 &{$\mathcal{N}(60,5^2)$}&{$\mathcal{N}^{\ast}(60,10^2)$}&{$\mathcal{N}(60,10^2)$}&{$\mathcal{N}(60,20^2)$}\\
 \begin{turn}{90}{\qquad\quad $\mathbf{u}_0$}\end{turn}&\includegraphics[width=0.2\linewidth]{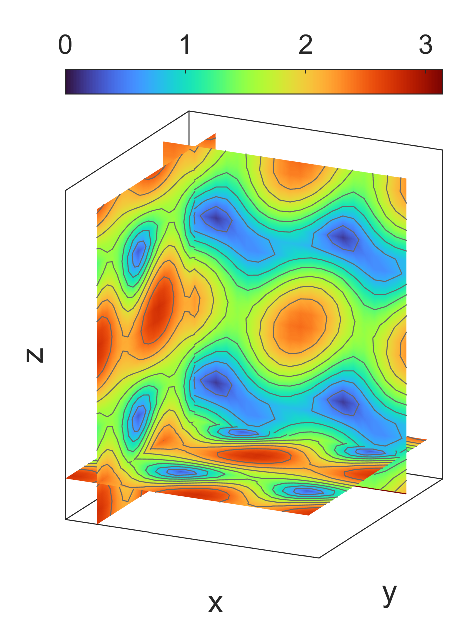}&\includegraphics[width=0.2\linewidth]{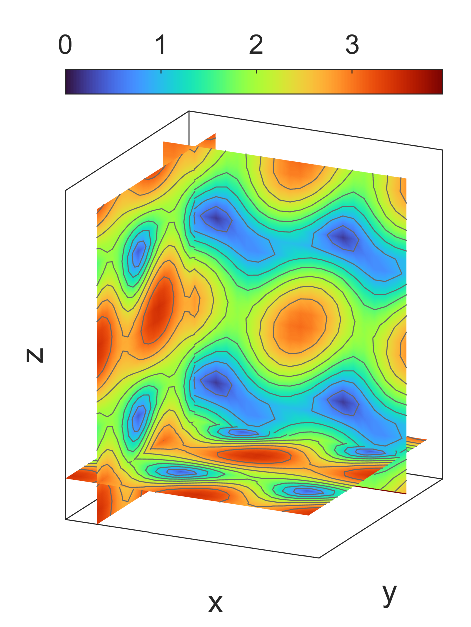}&\includegraphics[width=0.2\linewidth]{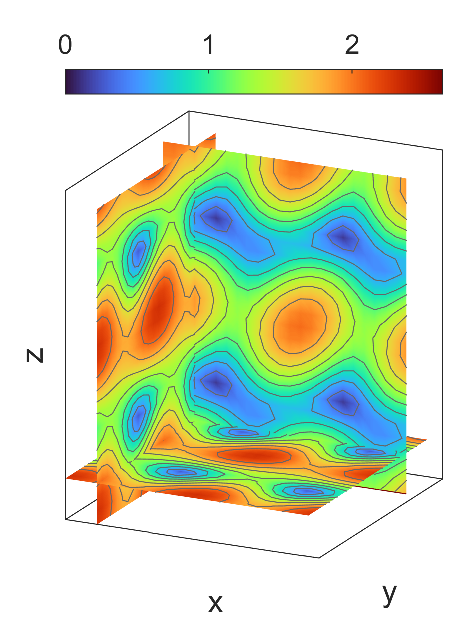}&\includegraphics[width=0.2\linewidth]{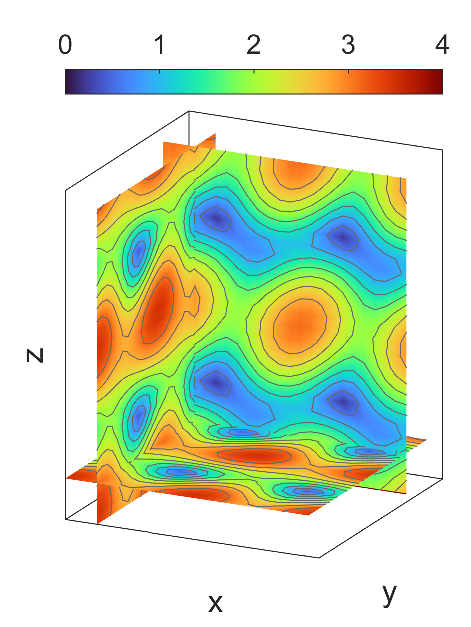}\\\begin{turn}{90}{\quad\quad Inference}\end{turn}&\includegraphics[width=0.2\linewidth]{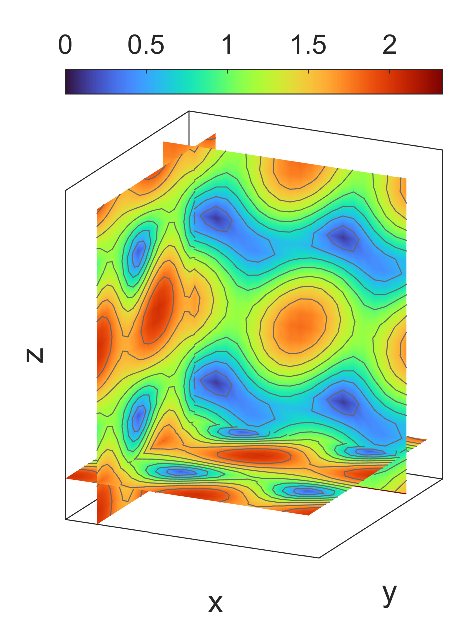}&\includegraphics[width=0.2\linewidth]{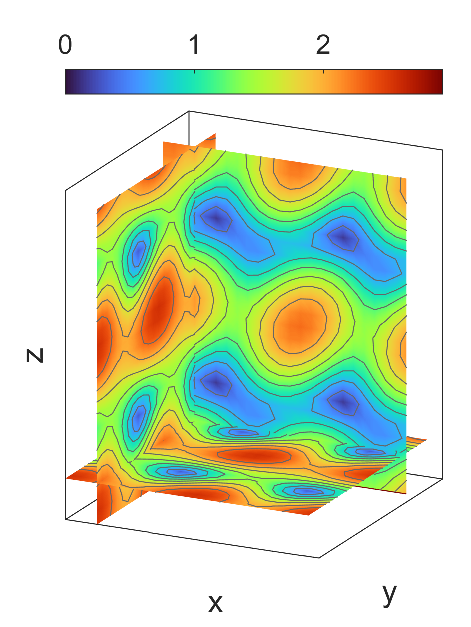}&\includegraphics[width=0.2\linewidth]{figure/3dbelt1sigma10bar.eps}&\includegraphics[width=0.2\linewidth]{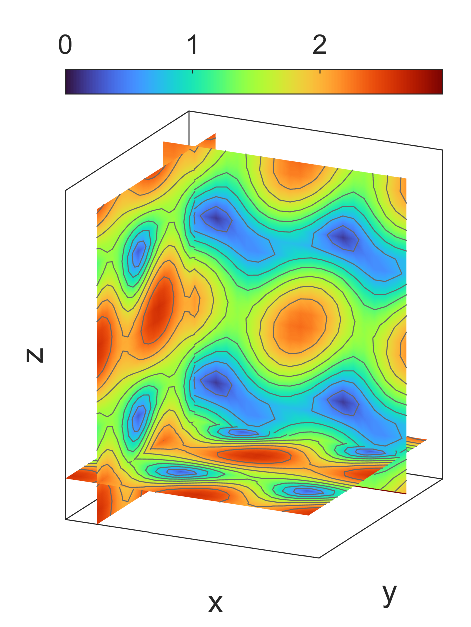}\\\begin{turn}{90}{\qquad\quad Pointwise error}\end{turn}&
 \includegraphics[width=0.2\linewidth]{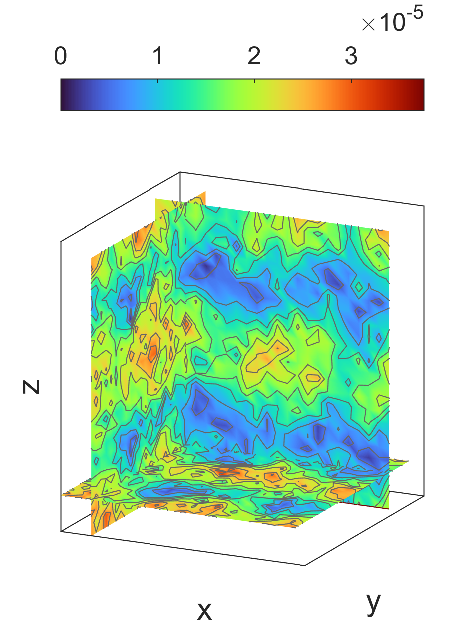}&\includegraphics[width=0.2\linewidth]{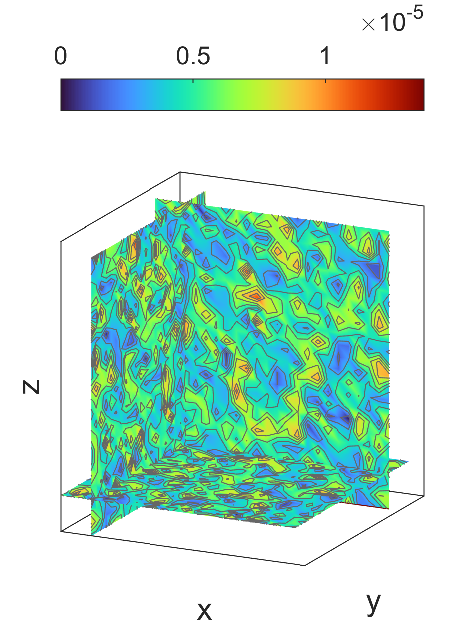}&\includegraphics[width=0.2\linewidth]{figure/3dbelt1sigma10err.eps}&\includegraphics[width=0.2\linewidth]{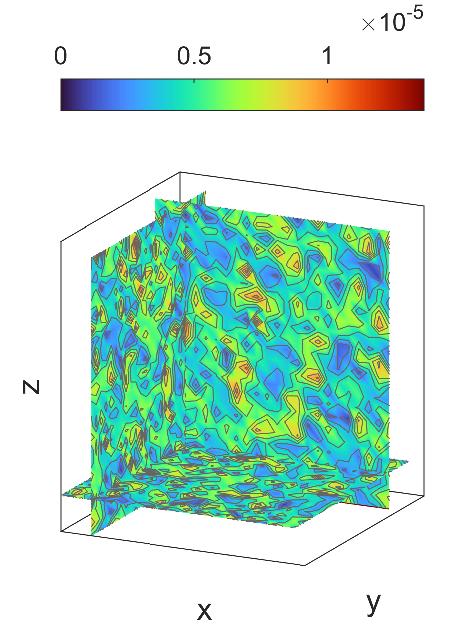}
      \end{tabular}
 \end{center}
 \caption{\textbf{Experiments for 3D initial conditions randomly generated by \eqref{Belt_sol} on $\mathcal{N}(60,5^2)$, $\mathcal{N}(60,10^2)$, $\mathcal{N}^\ast(60,10^2)$, and $\mathcal{N}(60,20^2)$.}}\label{fig:3dbelt_sigma}
 \end{figure}

\newpage
\subsection{Three dimensional NSE with random forcing functions}\label{sub:3d_force}
We performed experiments to measure the accuracy of inferences upon random forcing functions as input. The forcing functions, $f_x$, $f_y$, and $f_z$ were computed by the real part of
\begin{align}\label{generation:2d_force}
\frac{1}{2}\sin(t)\sum_{k_x,k_y,k_z=0}^{2}c_{k_xk_yk_z}\exp(i(k_xx+k_yy+k_zz))
\end{align}
where $c_{k_xk_yk_z}=a_{k_xk_yk_z}+ib_{k_xk_yk_z}$ after the random parameters $a_{k_xk_yk_z}$ and $b_{k_xk_yk_z}$ were drawn on $\mathcal{N}(0,5^2)$. Then, 600 forcing functions were produced for training samples. Information for the numerical methods is provided in \ref{tab:3d_forcing}.
\begin{table}[h!]    \centering   
    \begin{tabular}{|c|c||c|c|}
    \hline    
    Domain   & $[-1,1]^3$  & Boundary condition   & all zero (Dirichlet condition).   \\ \hline
    Bases type   & Legendre   & Initial condition   &$u_0=v_0=w_0=0$   \\ \hline
   $\Delta t$&0.01& The number of time steps& 100\\ \hline
    $\nu$&1& N& 18\\ \hline
    \end{tabular}
    \caption{Information on the numerical methods for 3D forcing functions.}
    \label{tab:3d_forcing}
\end{table}
 For test samples, we made 100 forcing functions from four cases, $\mathcal{N}(0,1^2)$, $\mathcal{N}(0,2^2)$, $\mathcal{N}(0,5^2)$, and $\mathcal{N}(0,10^2)$. Afterwards, solutions to NSEs for $t\in[0,1]$ were computed to employ them as reference solutions. Then, 4 sets of 100 errors in $L^2_{t,x}$, (for $p$, in $H^1_{t,x}$) were obtained between inferences of SpecONet and the reference solutions on $t\in[0,1]$. 
 
 Average of the errors over 100 samples is reported in table \ref{tab:error_force3d}. In particular, all the errors are within $4$\% implying that SpecONet is robust even though $\sigma$ increases from $1$ to $10$. In addition, error evolution over time is drown in  Fig.\ref{fig:3dforce_evolution}. The Rel.$L^2_x$ errors in Fig.\ref{fig:3dforce_evolution} are accumulated as time marched. That is because SpecONet emulates the rotational pressure correction method in temporal direction in order for unsupervised learning. 
  \begin{table}[h!]
\centering
\centering
\begin{tabular}{|c|c|c|c|c|}
\hline
\makecell{Types of input}   & $\sigma^2=1^2$&$\sigma^2=2^2$&$\sigma^2=5^2$&$\sigma^2=10^2$ \\
\hline\hline
Rel.$L^2_{t,x}$ error of $u$ &1.92E-02 &1.95e-02 &2.42E-02 &3.63E-02  \\
\hline
Rel.$L^2_{t,x}$ error of $v$ &1.95E-02 &2.00E-02 &2.47E-02 &3.69E-02  \\
\hline
Rel.$L^2_{t,x}$ error of $w$ &1.97E-02 &2.00E-02 &2.46E-02 &3.66E-02 \\
\hline
Rel.$H^1_{t,x}$ error of $p$ &2.61E-02 &1.87E-02 &1.83E-02 &2.47E-02  \\
\hline
\end{tabular}
\caption{\textbf{The relative error for $t\in(0,1]$ upon 3D force inputs.  }}\label{tab:error_force3d}
\end{table}

 \begin{figure}[h!]
 \centering
 \begin{tabular}{cccc} 
 {$u$}&{$v$}&{$w$}&{$\nabla p$}\\
 \includegraphics[width=0.2\linewidth]{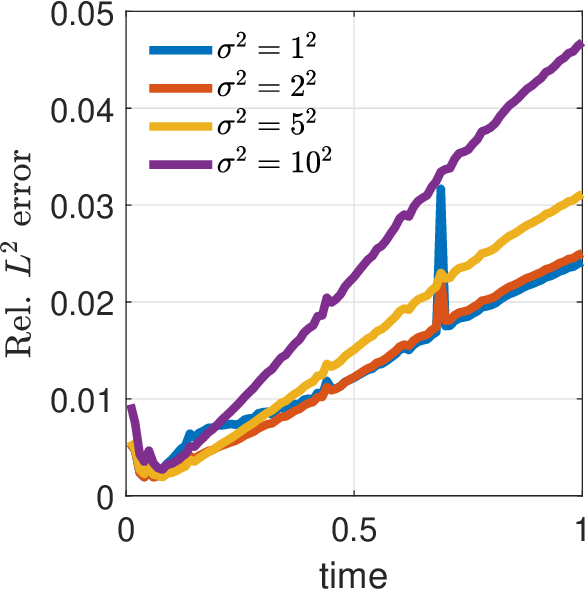}&\includegraphics[width=0.2\linewidth]{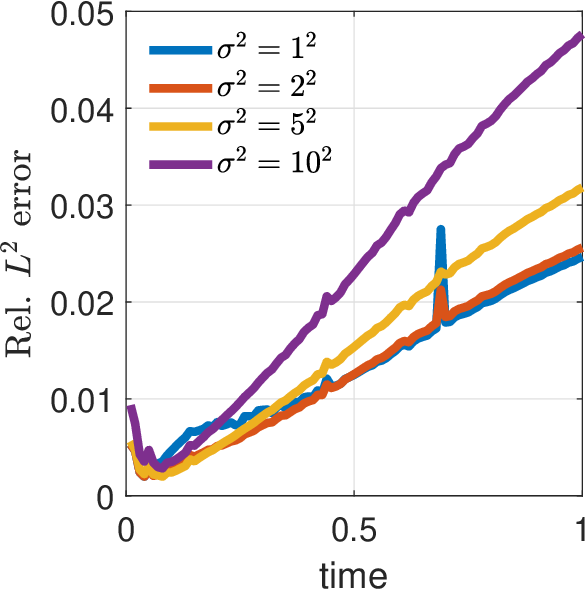}&\includegraphics[width=0.2\linewidth]{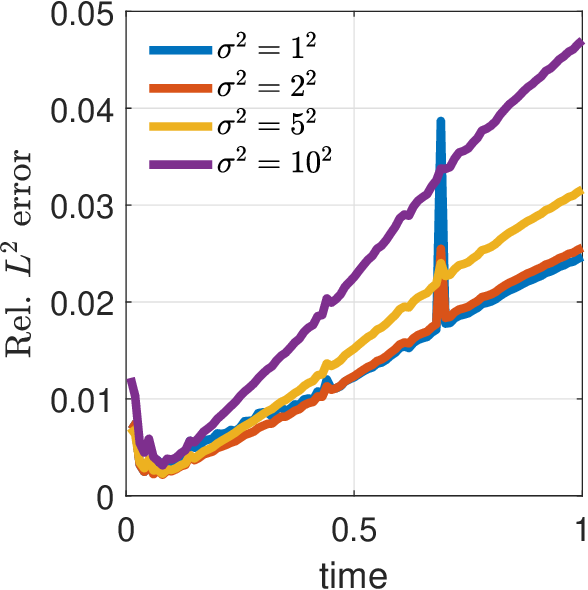}&\includegraphics[width=0.2\linewidth]{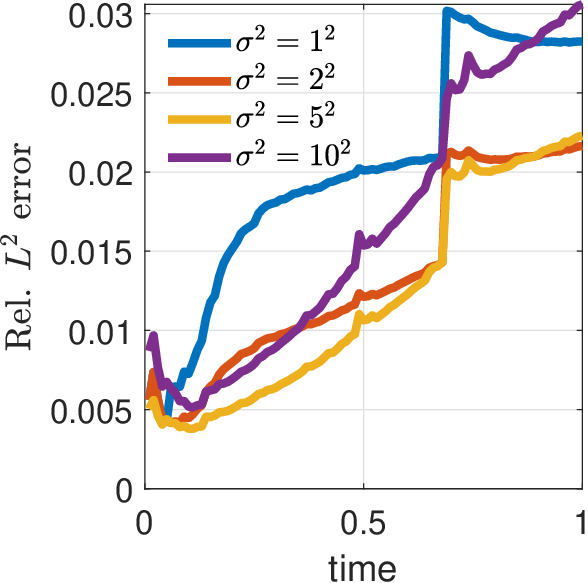}
      \end{tabular}
      \caption{\textbf{Rel. $L^2_x$ error profile over time}}\label{fig:3dforce_evolution}
 \end{figure}
  
 Figures of a reference solution and inferences depending on time are shown in Fig.\ref{fig:3dforce_time}. And also, figures of a reference solution and inferences varying $\sigma$ are displayed in Fig.\ref{fig:3dforce_sigma}.

\begin{figure}[hbp!]
\begin{center}
{\textbf{Numerical results for $\sigma=5$}}\\{\textbf{}}\\
\begin{tabular}{p{1em}cccc} 
&{t=0.25}&{t=0.5}&{t=0.75}&{t=1}\\
\begin{turn}{90}{\qquad\quad Inference}\end{turn}&
\includegraphics[width=0.2\linewidth]{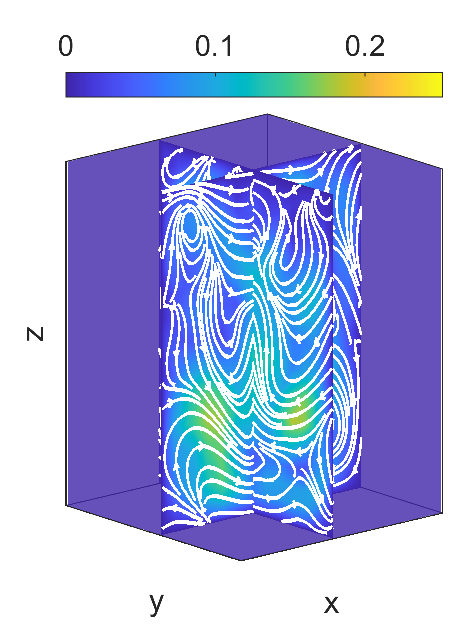}&\includegraphics[width=0.2\linewidth]{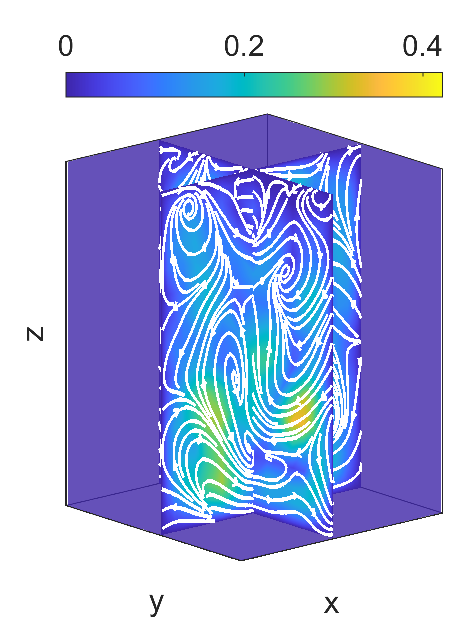}&\includegraphics[width=0.2\linewidth]{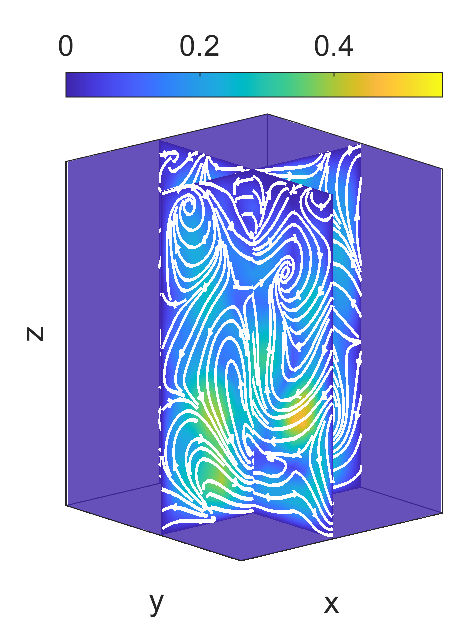}&\includegraphics[width=0.2\linewidth]{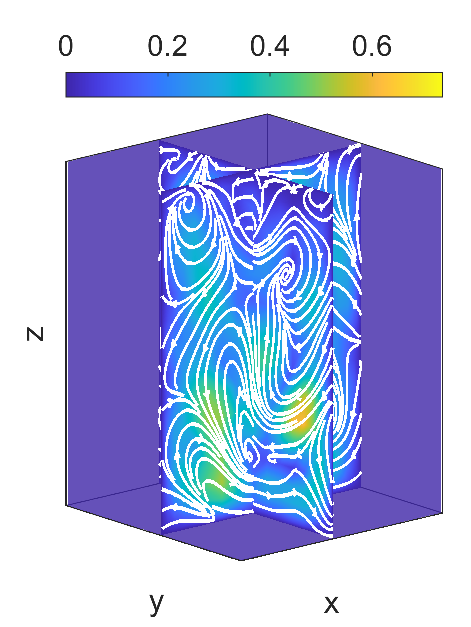}\\
\begin{turn}{90}{\quad\quad Pintwise error}\end{turn}&
\includegraphics[width=0.2\linewidth]{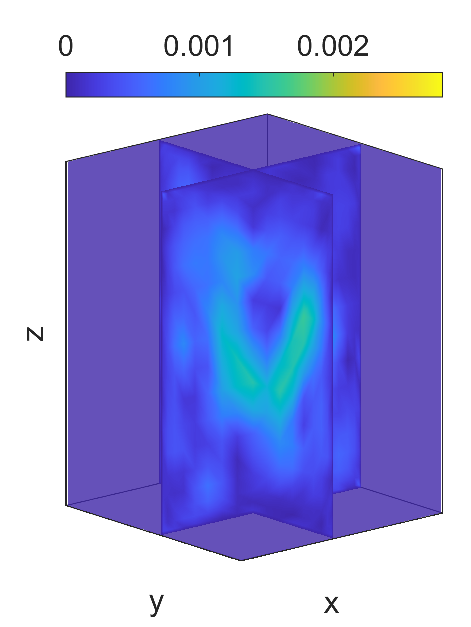}&\includegraphics[width=0.2\linewidth]{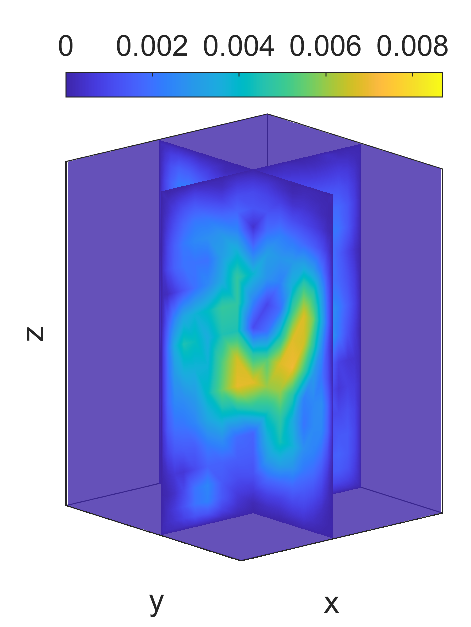}&\includegraphics[width=0.2\linewidth]{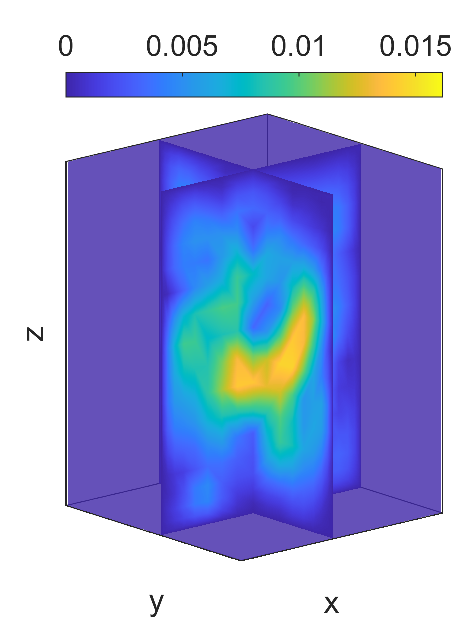}&\includegraphics[width=0.2\linewidth]{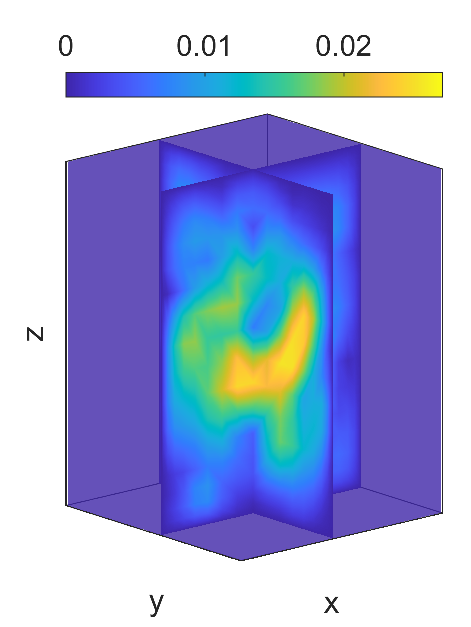}
     \end{tabular} 
\end{center}
\caption{\textbf{Experiments for 3D force functions randomly generated as in on $N(0,5^2)$.}  }\label{fig:3dforce_time}
\end{figure}

\begin{figure}[hbp!]
\begin{center}
{\textbf{Numerical results at $t=0.5$}}\\{\textbf{}}\\
\begin{tabular}{p{1em}cccc} 
&{$\sigma=1$}&{$\sigma=2$}&{$\sigma=5$}&{$\sigma=10$}\\
\begin{turn}{90}{\qquad\quad Force}\end{turn}&\includegraphics[width=0.2\linewidth]{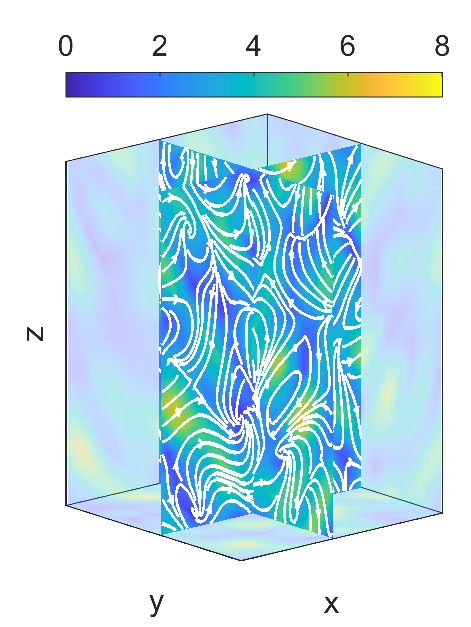}&\includegraphics[width=0.2\linewidth]{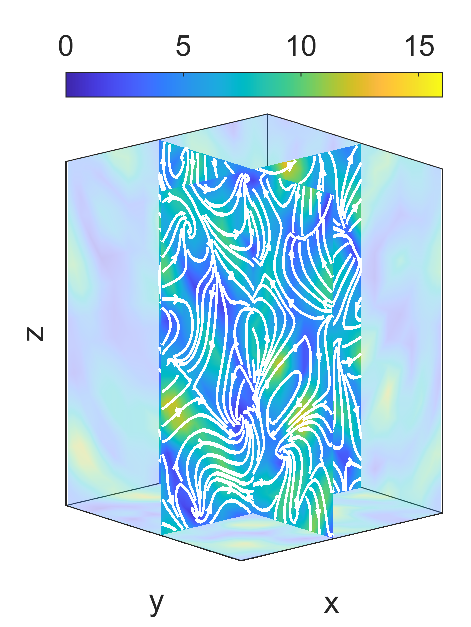}&\includegraphics[width=0.2\linewidth]{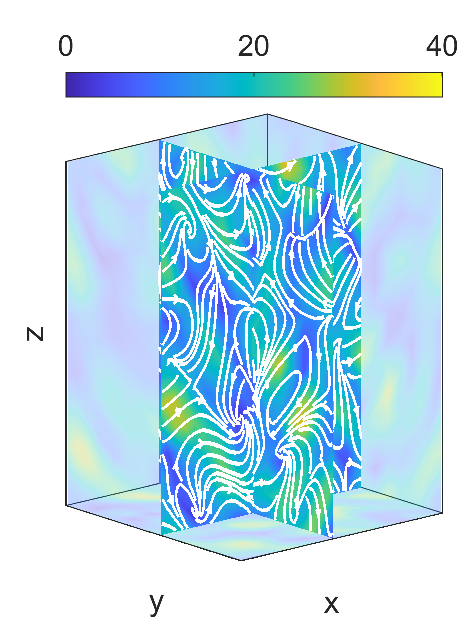}&\includegraphics[width=0.2\linewidth]{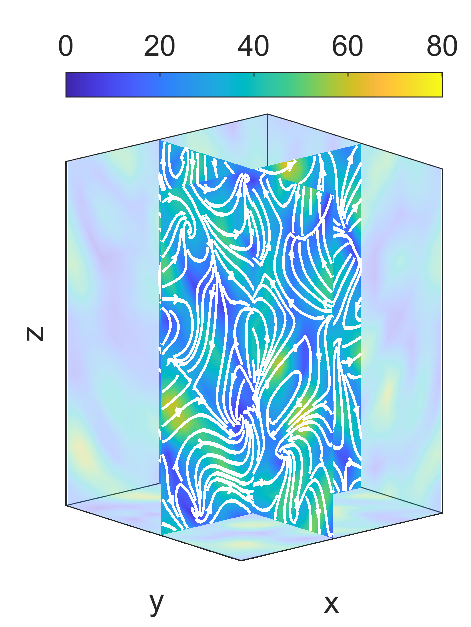}\\\begin{turn}{90}{\quad\qquad Inference}\end{turn}&\includegraphics[width=0.2\linewidth]{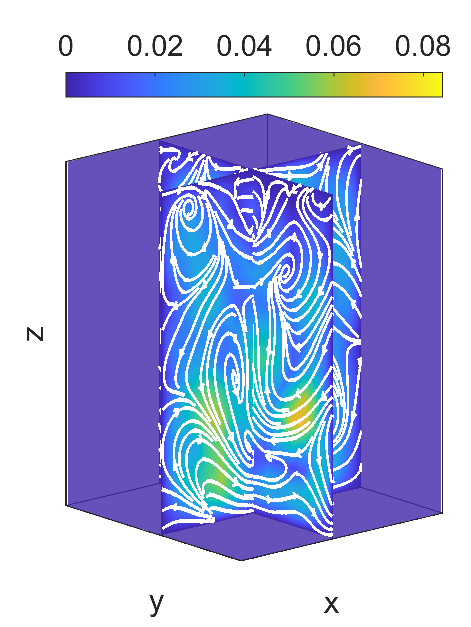}&\includegraphics[width=0.2\linewidth]{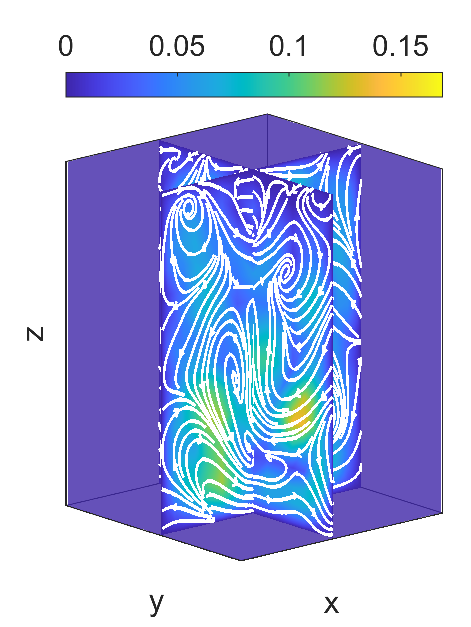}&\includegraphics[width=0.2\linewidth]{figure/3dforce50sigma5bar.eps}&\includegraphics[width=0.2\linewidth]{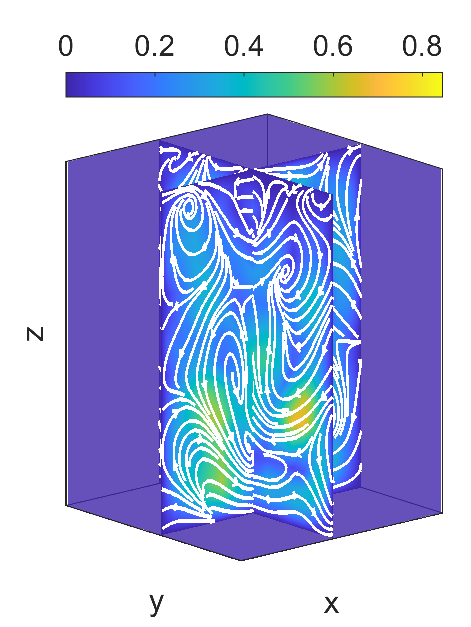}\\\begin{turn}{90}{\quad\quad Pointwise error}\end{turn}&
\includegraphics[width=0.2\linewidth]{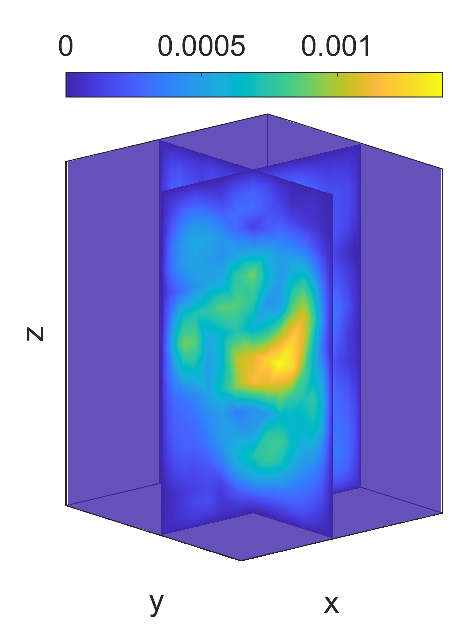}&\includegraphics[width=0.2\linewidth]{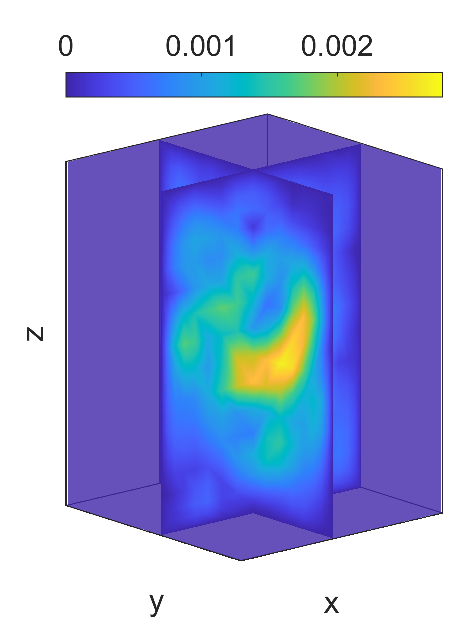}&\includegraphics[width=0.2\linewidth]{figure/3dforce50sigma5err.eps}&\includegraphics[width=0.2\linewidth]{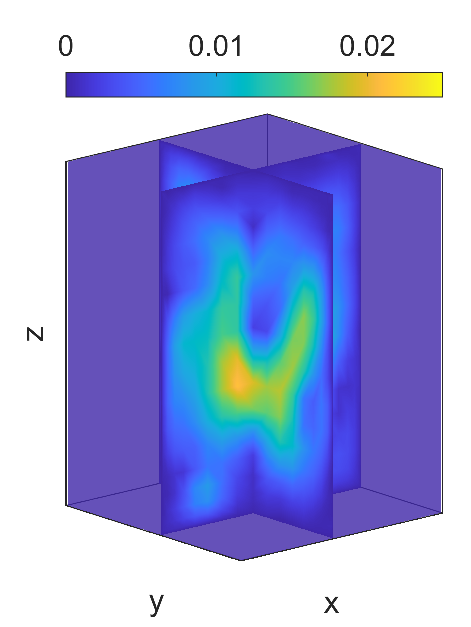}
     \end{tabular}
\end{center}
\caption{\textbf{Experiments for 3D force functions randomly generated varying $\sigma=1,2,5,$ and 10.}  }\label{fig:3dforce_sigma}
\end{figure}

\end{appendices}
\end{NoHyper}
\bibliography{ULGNet_ref}


\begin{thebibliography}{44}
\ifx \bisbn   \undefined \def \bisbn  #1{ISBN #1}\fi
\ifx \binits  \undefined \def \binits#1{#1}\fi
\ifx \bauthor  \undefined \def \bauthor#1{#1}\fi
\ifx \batitle  \undefined \def \batitle#1{#1}\fi
\ifx \bjtitle  \undefined \def \bjtitle#1{#1}\fi
\ifx \bvolume  \undefined \def \bvolume#1{\textbf{#1}}\fi
\ifx \byear  \undefined \def \byear#1{#1}\fi
\ifx \bissue  \undefined \def \bissue#1{#1}\fi
\ifx \bfpage  \undefined \def \bfpage#1{#1}\fi
\ifx \blpage  \undefined \def \blpage #1{#1}\fi
\ifx \burl  \undefined \def \burl#1{\textsf{#1}}\fi
\ifx \doiurl  \undefined \def \doiurl#1{\url{https://doi.org/#1}}\fi
\ifx \betal  \undefined \def \betal{\textit{et al.}}\fi
\ifx \binstitute  \undefined \def \binstitute#1{#1}\fi
\ifx \binstitutionaled  \undefined \def \binstitutionaled#1{#1}\fi
\ifx \bctitle  \undefined \def \bctitle#1{#1}\fi
\ifx \beditor  \undefined \def \beditor#1{#1}\fi
\ifx \bpublisher  \undefined \def \bpublisher#1{#1}\fi
\ifx \bbtitle  \undefined \def \bbtitle#1{#1}\fi
\ifx \bedition  \undefined \def \bedition#1{#1}\fi
\ifx \bseriesno  \undefined \def \bseriesno#1{#1}\fi
\ifx \blocation  \undefined \def \blocation#1{#1}\fi
\ifx \bsertitle  \undefined \def \bsertitle#1{#1}\fi
\ifx \bsnm \undefined \def \bsnm#1{#1}\fi
\ifx \bsuffix \undefined \def \bsuffix#1{#1}\fi
\ifx \bparticle \undefined \def \bparticle#1{#1}\fi
\ifx \barticle \undefined \def \barticle#1{#1}\fi
\bibcommenthead
\ifx \bconfdate \undefined \def \bconfdate #1{#1}\fi
\ifx \botherref \undefined \def \botherref #1{#1}\fi
\ifx \url \undefined \def \url#1{\textsf{#1}}\fi
\ifx \bchapter \undefined \def \bchapter#1{#1}\fi
\ifx \bbook \undefined \def \bbook#1{#1}\fi
\ifx \bcomment \undefined \def \bcomment#1{#1}\fi
\ifx \oauthor \undefined \def \oauthor#1{#1}\fi
\ifx \citeauthoryear \undefined \def \citeauthoryear#1{#1}\fi
\ifx \endbibitem  \undefined \def \endbibitem {}\fi
\ifx \bconflocation  \undefined \def \bconflocation#1{#1}\fi
\ifx \arxivurl  \undefined \def \arxivurl#1{\textsf{#1}}\fi
\csname PreBibitemsHook\endcsname

\bibitem[\protect\citeauthoryear{Anderson and
  Wendt}{1995}]{anderson1995computational}
\begin{bbook}
\bauthor{\bsnm{Anderson}, \binits{J.D.}},
\bauthor{\bsnm{Wendt}, \binits{J.}}:
\bbtitle{Computational Fluid Dynamics}
vol. \bseriesno{206}.
\bpublisher{Springer},
\blocation{-}
(\byear{1995})
\end{bbook}
\endbibitem

\bibitem[\protect\citeauthoryear{Wilcox et~al.}{1998}]{wilcox1998turbulence}
\begin{bbook}
\bauthor{\bsnm{Wilcox}, \binits{D.C.}}, \betal:
\bbtitle{Turbulence Modeling for CFD}
vol. \bseriesno{2}.
\bpublisher{DCW industries La Canada, CA},
\blocation{-}
(\byear{1998})
\end{bbook}
\endbibitem

\bibitem[\protect\citeauthoryear{Landau and Lifshitz}{2013}]{landau2013fluid}
\begin{bbook}
\bauthor{\bsnm{Landau}, \binits{L.D.}},
\bauthor{\bsnm{Lifshitz}, \binits{E.M.}}:
\bbtitle{Fluid Mechanics: Landau and Lifshitz: Course of Theoretical Physics,
  Volume 6}
vol. \bseriesno{6}.
\bpublisher{Elsevier},
\blocation{-}
(\byear{2013})
\end{bbook}
\endbibitem

\bibitem[\protect\citeauthoryear{Borden et~al.}{2012}]{borden2012phase}
\begin{barticle}
\bauthor{\bsnm{Borden}, \binits{M.J.}},
\bauthor{\bsnm{Verhoosel}, \binits{C.V.}},
\bauthor{\bsnm{Scott}, \binits{M.A.}},
\bauthor{\bsnm{Hughes}, \binits{T.J.}},
\bauthor{\bsnm{Landis}, \binits{C.M.}}:
\batitle{A phase-field description of dynamic brittle fracture}.
\bjtitle{Computer Methods in Applied Mechanics and Engineering}
\bvolume{217},
\bfpage{77}--\blpage{95}
(\byear{2012})
\end{barticle}
\endbibitem

\bibitem[\protect\citeauthoryear{Giustino}{2014}]{giustino2014materials}
\begin{bbook}
\bauthor{\bsnm{Giustino}, \binits{F.}}:
\bbtitle{Materials Modelling Using Density Functional Theory: Properties and
  Predictions}.
\bpublisher{Oxford University Press},
\blocation{-}
(\byear{2014})
\end{bbook}
\endbibitem

\bibitem[\protect\citeauthoryear{Saima et~al.}{2011}]{knj03}
\begin{bchapter}
\bauthor{\bsnm{Saima}, \binits{H.}},
\bauthor{\bsnm{Jaafar}, \binits{J.}},
\bauthor{\bsnm{Belhaouari}, \binits{S.}},
\bauthor{\bsnm{Jillani}, \binits{T.}}:
\bctitle{Intelligent methods for weather forecasting: A review}.
In: \bbtitle{2011 National Postgraduate Conference},
pp. \bfpage{1}--\blpage{6}
(\byear{2011}).
\bcomment{IEEE}
\end{bchapter}
\endbibitem

\bibitem[\protect\citeauthoryear{Bauer et~al.}{2015}]{bauer2015quiet}
\begin{barticle}
\bauthor{\bsnm{Bauer}, \binits{P.}},
\bauthor{\bsnm{Thorpe}, \binits{A.}},
\bauthor{\bsnm{Brunet}, \binits{G.}}:
\batitle{The quiet revolution of numerical weather prediction}.
\bjtitle{Nature}
\bvolume{525}(\bissue{7567}),
\bfpage{47}--\blpage{55}
(\byear{2015})
\end{barticle}
\endbibitem

\bibitem[\protect\citeauthoryear{Li et~al.}{2022}]{li2022machine}
\begin{barticle}
\bauthor{\bsnm{Li}, \binits{J.}},
\bauthor{\bsnm{Du}, \binits{X.}},
\bauthor{\bsnm{Martins}, \binits{J.R.}}:
\batitle{Machine learning in aerodynamic shape optimization}.
\bjtitle{Progress in Aerospace Sciences}
\bvolume{134},
\bfpage{100849}
(\byear{2022})
\end{barticle}
\endbibitem

\bibitem[\protect\citeauthoryear{Martins and
  Ning}{2021}]{martins2021engineering}
\begin{bbook}
\bauthor{\bsnm{Martins}, \binits{J.R.}},
\bauthor{\bsnm{Ning}, \binits{A.}}:
\bbtitle{Engineering Design Optimization}.
\bpublisher{Cambridge University Press},
\blocation{-}
(\byear{2021})
\end{bbook}
\endbibitem

\bibitem[\protect\citeauthoryear{Karniadakis et~al.}{2021}]{PINN007}
\begin{barticle}
\bauthor{\bsnm{Karniadakis}, \binits{G.E.}},
\bauthor{\bsnm{Kevrekidis}, \binits{I.G.}},
\bauthor{\bsnm{Lu}, \binits{L.}},
\bauthor{\bsnm{Perdikaris}, \binits{P.}},
\bauthor{\bsnm{Wang}, \binits{S.}},
\bauthor{\bsnm{Yang}, \binits{L.}}:
\batitle{Physics-informed machine learning}.
\bjtitle{Nature Reviews Physics}
\bvolume{3}(\bissue{6}),
\bfpage{422}--\blpage{440}
(\byear{2021})
\end{barticle}
\endbibitem

\bibitem[\protect\citeauthoryear{Vinuesa and
  Brunton}{2022}]{vinuesa2022enhancing}
\begin{barticle}
\bauthor{\bsnm{Vinuesa}, \binits{R.}},
\bauthor{\bsnm{Brunton}, \binits{S.L.}}:
\batitle{Enhancing computational fluid dynamics with machine learning}.
\bjtitle{Nature Computational Science}
\bvolume{2}(\bissue{6}),
\bfpage{358}--\blpage{366}
(\byear{2022})
\end{barticle}
\endbibitem

\bibitem[\protect\citeauthoryear{Price et~al.}{2025}]{price2025probabilistic}
\begin{barticle}
\bauthor{\bsnm{Price}, \binits{I.}},
\bauthor{\bsnm{Sanchez-Gonzalez}, \binits{A.}},
\bauthor{\bsnm{Alet}, \binits{F.}},
\bauthor{\bsnm{Andersson}, \binits{T.R.}},
\bauthor{\bsnm{El-Kadi}, \binits{A.}},
\bauthor{\bsnm{Masters}, \binits{D.}},
\bauthor{\bsnm{Ewalds}, \binits{T.}},
\bauthor{\bsnm{Stott}, \binits{J.}},
\bauthor{\bsnm{Mohamed}, \binits{S.}},
\bauthor{\bsnm{Battaglia}, \binits{P.}}, \betal:
\batitle{Probabilistic weather forecasting with machine learning}.
\bjtitle{Nature}
\bvolume{637}(\bissue{8044}),
\bfpage{84}--\blpage{90}
(\byear{2025})
\end{barticle}
\endbibitem

\bibitem[\protect\citeauthoryear{Lu et~al.}{2021}]{lu2021learning}
\begin{barticle}
\bauthor{\bsnm{Lu}, \binits{L.}},
\bauthor{\bsnm{Jin}, \binits{P.}},
\bauthor{\bsnm{Pang}, \binits{G.}},
\bauthor{\bsnm{Zhang}, \binits{Z.}},
\bauthor{\bsnm{Karniadakis}, \binits{G.E.}}:
\batitle{Learning nonlinear operators via {DeepONet} based on the universal
  approximation theorem of operators}.
\bjtitle{Nature Machine Intelligence}
\bvolume{3}(\bissue{3}),
\bfpage{218}--\blpage{229}
(\byear{2021})
\end{barticle}
\endbibitem

\bibitem[\protect\citeauthoryear{Lu et~al.}{2022}]{lu2022comprehensive}
\begin{barticle}
\bauthor{\bsnm{Lu}, \binits{L.}},
\bauthor{\bsnm{Meng}, \binits{X.}},
\bauthor{\bsnm{Cai}, \binits{S.}},
\bauthor{\bsnm{Mao}, \binits{Z.}},
\bauthor{\bsnm{Goswami}, \binits{S.}},
\bauthor{\bsnm{Zhang}, \binits{Z.}},
\bauthor{\bsnm{Karniadakis}, \binits{G.E.}}:
\batitle{A comprehensive and fair comparison of two neural operators (with
  practical extensions) based on fair data}.
\bjtitle{Computer Methods in Applied Mechanics and Engineering}
\bvolume{393},
\bfpage{114778}
(\byear{2022})
\end{barticle}
\endbibitem

\bibitem[\protect\citeauthoryear{Li et~al.}{2020}]{li2020fourier}
\begin{botherref}
\oauthor{\bsnm{Li}, \binits{Z.}},
\oauthor{\bsnm{Kovachki}, \binits{N.}},
\oauthor{\bsnm{Azizzadenesheli}, \binits{K.}},
\oauthor{\bsnm{Liu}, \binits{B.}},
\oauthor{\bsnm{Bhattacharya}, \binits{K.}},
\oauthor{\bsnm{Stuart}, \binits{A.}},
\oauthor{\bsnm{Anandkumar}, \binits{A.}}:
Fourier neural operator for parametric partial differential equations.
arXiv preprint arXiv:2010.08895
(2020)
\end{botherref}
\endbibitem

\bibitem[\protect\citeauthoryear{Li et~al.}{2023}]{li2023geometry}
\begin{barticle}
\bauthor{\bsnm{Li}, \binits{Z.}},
\bauthor{\bsnm{Kovachki}, \binits{N.}},
\bauthor{\bsnm{Choy}, \binits{C.}},
\bauthor{\bsnm{Li}, \binits{B.}},
\bauthor{\bsnm{Kossaifi}, \binits{J.}},
\bauthor{\bsnm{Otta}, \binits{S.}},
\bauthor{\bsnm{Nabian}, \binits{M.A.}},
\bauthor{\bsnm{Stadler}, \binits{M.}},
\bauthor{\bsnm{Hundt}, \binits{C.}},
\bauthor{\bsnm{Azizzadenesheli}, \binits{K.}}, \betal:
\batitle{Geometry-informed neural operator for large-scale 3d pdes}.
\bjtitle{Advances in Neural Information Processing Systems}
\bvolume{36},
\bfpage{35836}--\blpage{35854}
(\byear{2023})
\end{barticle}
\endbibitem

\bibitem[\protect\citeauthoryear{Yin et~al.}{2024}]{yin2024scalable}
\begin{barticle}
\bauthor{\bsnm{Yin}, \binits{M.}},
\bauthor{\bsnm{Charon}, \binits{N.}},
\bauthor{\bsnm{Brody}, \binits{R.}},
\bauthor{\bsnm{Lu}, \binits{L.}},
\bauthor{\bsnm{Trayanova}, \binits{N.}},
\bauthor{\bsnm{Maggioni}, \binits{M.}}:
\batitle{A scalable framework for learning the geometry-dependent solution
  operators of partial differential equations}.
\bjtitle{Nature computational science}
\bvolume{4}(\bissue{12}),
\bfpage{928}--\blpage{940}
(\byear{2024})
\end{barticle}
\endbibitem

\bibitem[\protect\citeauthoryear{Kadeethum
  et~al.}{2021}]{kadeethum2021framework}
\begin{barticle}
\bauthor{\bsnm{Kadeethum}, \binits{T.}},
\bauthor{\bsnm{O’Malley}, \binits{D.}},
\bauthor{\bsnm{Fuhg}, \binits{J.N.}},
\bauthor{\bsnm{Choi}, \binits{Y.}},
\bauthor{\bsnm{Lee}, \binits{J.}},
\bauthor{\bsnm{Viswanathan}, \binits{H.S.}},
\bauthor{\bsnm{Bouklas}, \binits{N.}}:
\batitle{A framework for data-driven solution and parameter estimation of pdes
  using conditional generative adversarial networks}.
\bjtitle{Nature Computational Science}
\bvolume{1}(\bissue{12}),
\bfpage{819}--\blpage{829}
(\byear{2021})
\end{barticle}
\endbibitem

\bibitem[\protect\citeauthoryear{Jin et~al.}{2021}]{jin2021nsfnets}
\begin{barticle}
\bauthor{\bsnm{Jin}, \binits{X.}},
\bauthor{\bsnm{Cai}, \binits{S.}},
\bauthor{\bsnm{Li}, \binits{H.}},
\bauthor{\bsnm{Karniadakis}, \binits{G.E.}}:
\batitle{Nsfnets (navier-stokes flow nets): Physics-informed neural networks
  for the incompressible navier-stokes equations}.
\bjtitle{Journal of Computational Physics}
\bvolume{426},
\bfpage{109951}
(\byear{2021})
\end{barticle}
\endbibitem

\bibitem[\protect\citeauthoryear{Cho et~al.}{2023}]{cho2023separable}
\begin{barticle}
\bauthor{\bsnm{Cho}, \binits{J.}},
\bauthor{\bsnm{Nam}, \binits{S.}},
\bauthor{\bsnm{Yang}, \binits{H.}},
\bauthor{\bsnm{Yun}, \binits{S.-B.}},
\bauthor{\bsnm{Hong}, \binits{Y.}},
\bauthor{\bsnm{Park}, \binits{E.}}:
\batitle{Separable physics-informed neural networks}.
\bjtitle{Advances in Neural Information Processing Systems}
\bvolume{36},
\bfpage{23761}--\blpage{23788}
(\byear{2023})
\end{barticle}
\endbibitem

\bibitem[\protect\citeauthoryear{Wang et~al.}{2024}]{wang2024respecting}
\begin{barticle}
\bauthor{\bsnm{Wang}, \binits{S.}},
\bauthor{\bsnm{Sankaran}, \binits{S.}},
\bauthor{\bsnm{Perdikaris}, \binits{P.}}:
\batitle{Respecting causality for training physics-informed neural networks}.
\bjtitle{Computer Methods in Applied Mechanics and Engineering}
\bvolume{421},
\bfpage{116813}
(\byear{2024})
\end{barticle}
\endbibitem

\bibitem[\protect\citeauthoryear{Ovadia et~al.}{2024}]{ovadia2024vito}
\begin{barticle}
\bauthor{\bsnm{Ovadia}, \binits{O.}},
\bauthor{\bsnm{Kahana}, \binits{A.}},
\bauthor{\bsnm{Stinis}, \binits{P.}},
\bauthor{\bsnm{Turkel}, \binits{E.}},
\bauthor{\bsnm{Givoli}, \binits{D.}},
\bauthor{\bsnm{Karniadakis}, \binits{G.E.}}:
\batitle{Vito: Vision transformer-operator}.
\bjtitle{Computer Methods in Applied Mechanics and Engineering}
\bvolume{428},
\bfpage{117109}
(\byear{2024})
\end{barticle}
\endbibitem

\bibitem[\protect\citeauthoryear{Raissi et~al.}{2019}]{PINN009}
\begin{barticle}
\bauthor{\bsnm{Raissi}, \binits{M.}},
\bauthor{\bsnm{Perdikaris}, \binits{P.}},
\bauthor{\bsnm{Karniadakis}, \binits{G.E.}}:
\batitle{Physics-informed neural networks: A deep learning framework for
  solving forward and inverse problems involving nonlinear partial differential
  equations}.
\bjtitle{Journal of Computational physics}
\bvolume{378},
\bfpage{686}--\blpage{707}
(\byear{2019})
\end{barticle}
\endbibitem

\bibitem[\protect\citeauthoryear{Cuomo et~al.}{2022}]{cuomo2022scientific}
\begin{barticle}
\bauthor{\bsnm{Cuomo}, \binits{S.}},
\bauthor{\bsnm{Di~Cola}, \binits{V.S.}},
\bauthor{\bsnm{Giampaolo}, \binits{F.}},
\bauthor{\bsnm{Rozza}, \binits{G.}},
\bauthor{\bsnm{Raissi}, \binits{M.}},
\bauthor{\bsnm{Piccialli}, \binits{F.}}:
\batitle{Scientific machine learning through physics--informed neural networks:
  where we are and what’s next}.
\bjtitle{Journal of Scientific Computing}
\bvolume{92}(\bissue{3}),
\bfpage{88}
(\byear{2022})
\end{barticle}
\endbibitem

\bibitem[\protect\citeauthoryear{Shukla et~al.}{2024}]{shukla2024comprehensive}
\begin{barticle}
\bauthor{\bsnm{Shukla}, \binits{K.}},
\bauthor{\bsnm{Toscano}, \binits{J.D.}},
\bauthor{\bsnm{Wang}, \binits{Z.}},
\bauthor{\bsnm{Zou}, \binits{Z.}},
\bauthor{\bsnm{Karniadakis}, \binits{G.E.}}:
\batitle{A comprehensive and fair comparison between mlp and kan
  representations for differential equations and operator networks}.
\bjtitle{Computer Methods in Applied Mechanics and Engineering}
\bvolume{431},
\bfpage{117290}
(\byear{2024})
\end{barticle}
\endbibitem

\bibitem[\protect\citeauthoryear{Guermond and Shen}{2004}]{guermond2004error}
\begin{barticle}
\bauthor{\bsnm{Guermond}, \binits{J.}},
\bauthor{\bsnm{Shen}, \binits{J.}}:
\batitle{On the error estimates for the rotational pressure-correction
  projection methods}.
\bjtitle{Mathematics of Computation}
\bvolume{73}(\bissue{248}),
\bfpage{1719}--\blpage{1737}
(\byear{2004})
\end{barticle}
\endbibitem

\bibitem[\protect\citeauthoryear{Shen et~al.}{2011}]{shen2011spectral}
\begin{bbook}
\bauthor{\bsnm{Shen}, \binits{J.}},
\bauthor{\bsnm{Tang}, \binits{T.}},
\bauthor{\bsnm{Wang}, \binits{L.-L.}}:
\bbtitle{Spectral Methods: Algorithms, Analysis and Applications}
vol. \bseriesno{41}.
\bpublisher{Springer},
\blocation{-}
(\byear{2011})
\end{bbook}
\endbibitem

\bibitem[\protect\citeauthoryear{Kalnay}{2003}]{kalnay2003atmospheric}
\begin{bbook}
\bauthor{\bsnm{Kalnay}, \binits{E.}}:
\bbtitle{Atmospheric Modeling, Data Assimilation and Predictability}.
\bpublisher{Cambridge university press},
\blocation{-}
(\byear{2003})
\end{bbook}
\endbibitem

\bibitem[\protect\citeauthoryear{Cao et~al.}{2020}]{cao2020ensemble}
\begin{barticle}
\bauthor{\bsnm{Cao}, \binits{Y.}},
\bauthor{\bsnm{Geddes}, \binits{T.A.}},
\bauthor{\bsnm{Yang}, \binits{J.Y.H.}},
\bauthor{\bsnm{Yang}, \binits{P.}}:
\batitle{Ensemble deep learning in bioinformatics}.
\bjtitle{Nature Machine Intelligence}
\bvolume{2}(\bissue{9}),
\bfpage{500}--\blpage{508}
(\byear{2020})
\end{barticle}
\endbibitem

\bibitem[\protect\citeauthoryear{Nti et~al.}{2020}]{nti2020comprehensive}
\begin{barticle}
\bauthor{\bsnm{Nti}, \binits{I.K.}},
\bauthor{\bsnm{Adekoya}, \binits{A.F.}},
\bauthor{\bsnm{Weyori}, \binits{B.A.}}:
\batitle{A comprehensive evaluation of ensemble learning for stock-market
  prediction}.
\bjtitle{Journal of Big Data}
\bvolume{7}(\bissue{1}),
\bfpage{20}
(\byear{2020})
\end{barticle}
\endbibitem

\bibitem[\protect\citeauthoryear{Luo and Wang}{2018}]{luo2018ensemble}
\begin{barticle}
\bauthor{\bsnm{Luo}, \binits{Y.}},
\bauthor{\bsnm{Wang}, \binits{Z.}}:
\batitle{An ensemble algorithm for numerical solutions to deterministic and
  random parabolic pdes}.
\bjtitle{SIAM Journal on Numerical Analysis}
\bvolume{56}(\bissue{2}),
\bfpage{859}--\blpage{876}
(\byear{2018})
\end{barticle}
\endbibitem

\bibitem[\protect\citeauthoryear{Kovachki et~al.}{2023}]{kovachki2023neural}
\begin{barticle}
\bauthor{\bsnm{Kovachki}, \binits{N.}},
\bauthor{\bsnm{Li}, \binits{Z.}},
\bauthor{\bsnm{Liu}, \binits{B.}},
\bauthor{\bsnm{Azizzadenesheli}, \binits{K.}},
\bauthor{\bsnm{Bhattacharya}, \binits{K.}},
\bauthor{\bsnm{Stuart}, \binits{A.}},
\bauthor{\bsnm{Anandkumar}, \binits{A.}}:
\batitle{Neural operator: Learning maps between function spaces with
  applications to pdes}.
\bjtitle{Journal of Machine Learning Research}
\bvolume{24}(\bissue{89}),
\bfpage{1}--\blpage{97}
(\byear{2023})
\end{barticle}
\endbibitem

\bibitem[\protect\citeauthoryear{Tran et~al.}{2021}]{tran2021factorized}
\begin{botherref}
\oauthor{\bsnm{Tran}, \binits{A.}},
\oauthor{\bsnm{Mathews}, \binits{A.}},
\oauthor{\bsnm{Xie}, \binits{L.}},
\oauthor{\bsnm{Ong}, \binits{C.S.}}:
Factorized fourier neural operators.
arXiv preprint arXiv:2111.13802
(2021)
\end{botherref}
\endbibitem

\bibitem[\protect\citeauthoryear{Kovachki et~al.}{2021}]{kovachki2021universal}
\begin{barticle}
\bauthor{\bsnm{Kovachki}, \binits{N.}},
\bauthor{\bsnm{Lanthaler}, \binits{S.}},
\bauthor{\bsnm{Mishra}, \binits{S.}}:
\batitle{On universal approximation and error bounds for fourier neural
  operators}.
\bjtitle{Journal of Machine Learning Research}
\bvolume{22}(\bissue{290}),
\bfpage{1}--\blpage{76}
(\byear{2021})
\end{barticle}
\endbibitem

\bibitem[\protect\citeauthoryear{Lee et~al.}{2025}]{lee2025finite}
\begin{barticle}
\bauthor{\bsnm{Lee}, \binits{J.Y.}},
\bauthor{\bsnm{Ko}, \binits{S.}},
\bauthor{\bsnm{Hong}, \binits{Y.}}:
\batitle{Finite element operator network for solving elliptic-type parametric
  pdes}.
\bjtitle{SIAM Journal on Scientific Computing}
\bvolume{47}(\bissue{2}),
\bfpage{501}--\blpage{528}
(\byear{2025})
\end{barticle}
\endbibitem

\bibitem[\protect\citeauthoryear{Lynch}{2008}]{lynch2008origins}
\begin{barticle}
\bauthor{\bsnm{Lynch}, \binits{P.}}:
\batitle{The origins of computer weather prediction and climate modeling}.
\bjtitle{Journal of computational physics}
\bvolume{227}(\bissue{7}),
\bfpage{3431}--\blpage{3444}
(\byear{2008})
\end{barticle}
\endbibitem

\bibitem[\protect\citeauthoryear{Stocker}{2014}]{stocker2014climate}
\begin{bbook}
\bauthor{\bsnm{Stocker}, \binits{T.}}:
\bbtitle{Climate Change 2013: the Physical Science Basis: Working Group I
  Contribution to the Fifth Assessment Report of the Intergovernmental Panel on
  Climate Change}.
\bpublisher{Cambridge university press},
\blocation{-}
(\byear{2014})
\end{bbook}
\endbibitem

\bibitem[\protect\citeauthoryear{Evans}{2022}]{evans2022partial}
\begin{bbook}
\bauthor{\bsnm{Evans}, \binits{L.C.}}:
\bbtitle{Partial Differential Equations}
vol. \bseriesno{19}.
\bpublisher{American mathematical society},
\blocation{-}
(\byear{2022})
\end{bbook}
\endbibitem

\bibitem[\protect\citeauthoryear{Trefethen}{2000}]{trefethen2000spectral}
\begin{bbook}
\bauthor{\bsnm{Trefethen}, \binits{L.N.}}:
\bbtitle{Spectral Methods in MATLAB}.
\bpublisher{SIAM},
\blocation{-}
(\byear{2000})
\end{bbook}
\endbibitem

\bibitem[\protect\citeauthoryear{Shen}{1994}]{shen1994efficient}
\begin{barticle}
\bauthor{\bsnm{Shen}, \binits{J.}}:
\batitle{Efficient spectral-galerkin method i. direct solvers of second-and
  fourth-order equations using legendre polynomials}.
\bjtitle{SIAM Journal on Scientific Computing}
\bvolume{15}(\bissue{6}),
\bfpage{1489}--\blpage{1505}
(\byear{1994})
\end{barticle}
\endbibitem

\bibitem[\protect\citeauthoryear{Choi et~al.}{2024}]{choi2024spectral}
\begin{barticle}
\bauthor{\bsnm{Choi}, \binits{J.}},
\bauthor{\bsnm{Yun}, \binits{T.}},
\bauthor{\bsnm{Kim}, \binits{N.}},
\bauthor{\bsnm{Hong}, \binits{Y.}}:
\batitle{Spectral operator learning for parametric pdes without data reliance}.
\bjtitle{Computer Methods in Applied Mechanics and Engineering}
\bvolume{420},
\bfpage{116678}
(\byear{2024})
\end{barticle}
\endbibitem

\bibitem[\protect\citeauthoryear{Yu et~al.}{2022}]{yu2022gradient}
\begin{barticle}
\bauthor{\bsnm{Yu}, \binits{J.}},
\bauthor{\bsnm{Lu}, \binits{L.}},
\bauthor{\bsnm{Meng}, \binits{X.}},
\bauthor{\bsnm{Karniadakis}, \binits{G.E.}}:
\batitle{Gradient-enhanced physics-informed neural networks for forward and
  inverse pde problems}.
\bjtitle{Computer Methods in Applied Mechanics and Engineering}
\bvolume{393},
\bfpage{114823}
(\byear{2022})
\end{barticle}
\endbibitem

\bibitem[\protect\citeauthoryear{Shin et~al.}{2020}]{shin2020convergence}
\begin{botherref}
\oauthor{\bsnm{Shin}, \binits{Y.}},
\oauthor{\bsnm{Darbon}, \binits{J.}},
\oauthor{\bsnm{Karniadakis}, \binits{G.E.}}:
On the convergence of physics informed neural networks for linear second-order
  elliptic and parabolic type pdes.
arXiv preprint arXiv:2004.01806
(2020)
\end{botherref}
\endbibitem

\bibitem[\protect\citeauthoryear{Antuono}{2020}]{antuono2020tri}
\begin{barticle}
\bauthor{\bsnm{Antuono}, \binits{M.}}:
\batitle{Tri-periodic fully three-dimensional analytic solutions for the
  navier--stokes equations}.
\bjtitle{Journal of Fluid Mechanics}
\bvolume{890},
\bfpage{23}
(\byear{2020})
\end{barticle}
\endbibitem

\end{thebibliography}

\end{document}